\documentclass{article}

\usepackage[preprint]{neurips_2026}

\usepackage[utf8]{inputenc}
\usepackage[T1]{fontenc}
\usepackage{hyperref}
\usepackage{url}
\usepackage{booktabs}
\usepackage{amsfonts}
\usepackage{amsmath}
\usepackage{amssymb}
\usepackage{nicefrac}
\usepackage{microtype}
\usepackage{graphicx}
\usepackage{xcolor}
\usepackage{subcaption}
\usepackage{multirow}
\usepackage{algorithm}
\usepackage{algorithmic}

\usepackage{xspace}

\usepackage{tcolorbox}  %
\usepackage{colortbl}   %

\newcommand{\muP}{\mu\text{P}}

\newcommand{\Ocal}{\mathcal{O}}
\newcommand{\Lcal}{\mathcal{L}}
\newcommand{\dmodel}{d_{\text{model}}}
\newcommand{\dinner}{d_{\text{inner}}}

\newcommand{\MM}{\texttt{MM}\xspace}
\newcommand{\TT}{\texttt{TT}\xspace}
\newcommand{\MTMT}{\texttt{MTMT}\xspace}
\newcommand{\MHMH}{\texttt{MHMH}\xspace}
\newcommand{\HHHH}{\texttt{HHHH}\xspace}
\newcommand{\HTHT}{\texttt{HTHT}\xspace}

\newcommand{\HMHM}{\texttt{HMHM}\xspace}

\newcommand{\TMTM}{\texttt{TMTM}\xspace}
\newcommand{\THTH}{\texttt{THTH}\xspace}

\newcommand{\naive}{{na\"ive}\xspace}
\newcommand{\Naive}{{Na\"ive}\xspace}

\newcommand{\nConfigs}{4{,}500+\xspace}
\newcommand{\nGpuHrs}{${\sim}$5{,}500\xspace}

\newcommand{\snu}{Seoul National University}
\newcommand{\hodoo}{Hodoo AI}
\newcommand{\lgai}{LG AI Research}
\newcommand{\sbi}{SB Intuitions}

\title{Learning Rate Transfer \\ for Hybrid Transformer-SSM Architectures}

\author{%
  Jimin Seo\textsuperscript{1} \quad
  Gyubok Lee\textsuperscript{2}\thanks{Work done prior to joining SB Intuitions.} \quad
  Yeonsik Jo\textsuperscript{3} \quad
  Kiwoong Yoo\textsuperscript{3} \quad
  Yeongoon Kim\textsuperscript{1} \\
  \textbf{Minhae Oh\textsuperscript{1} \quad
  Jin Woo Koo\textsuperscript{1} \quad
  Suhwan Kim\textsuperscript{1} \quad
  Nakyung Lee\textsuperscript{1} \quad
  Minsik Seol\textsuperscript{1}} \\
  \textbf{Idris Nechnech\textsuperscript{1} \quad
  Jaehyeon Kim\textsuperscript{4} \quad
  Giho Lee\textsuperscript{4} \quad
  Jungwoo Lee\textsuperscript{1,4}\thanks{Corresponding author.}} \\[6pt]
  \textsuperscript{1}\snu \quad
  \textsuperscript{2}\sbi \quad
  \textsuperscript{3}\lgai \quad
  \textsuperscript{4}\hodoo \\[4pt]
  {\footnotesize\texttt{\{jimin.seo, ygoonkim, minhae.oh, jinukoo, curtis05, leena, seolminsik\}@cml.snu.ac.kr}} \\
  {\footnotesize\texttt{\{inechnech, junglee\}@snu.ac.kr, gyubok.lee@sbintuitions.co.jp}} \\
  {\footnotesize\texttt{\{yeonsik.jo, kiwoong.yoo\}@lgresearch.ai, \{jaehyeon.kim, giho.lee\}@hodooai.com}}
}

\begin{document}

\maketitle

\begin{abstract}
We study learning rate (LR) scaling for hybrid architectures combining Transformer and State-Space Model (SSM) blocks, a class adopted by several recent production language models.
In particular, we focus on the gap between the theoretical scaling rules derived for SSMs under zero-order-hold (ZOH) discretization at infinite width with growing state size, and the field-standard practical implementations using simplified-ZOH Mamba at fixed state size.
Surprisingly, in this practical regime hybrid architectures achieve a \textbf{near-zero LR transfer gap} across widths $256$--$2048$ and depths $4$--$32$ up to billion-parameter scale using only the original $\muP$ prescription, even though SSM operations fall outside its Tensor Programs representability conditions and every parameterization we test fails the standard coordinate-check diagnostic of $\muP$ correctness.
We attribute this to a two-condition decomposition of LR transfer in hybrid architectures: a \emph{global} update-to-weight invariance, enforced by $\muP$'s initialization and LR scaling; and a \emph{local} per-component balance, provided by AdamW's per-parameter normalization.
Our observations show that the optimal LR is invariant to width up to $8\times$, that this width invariance holds across depth, sequence length, batch size, and Transformer-to-SSM ratio, and that it transfers to Nemotron-H, a production hybrid outside our custom architecture set.
We hope these findings fill the gap between theoretical scaling rules and practical hybrid implementations, and stimulate further research toward bridging it.\footnote{Code is available at \url{https://github.com/argyroneta-aquatica/hybrid-mup}.}

\end{abstract}

\section{Introduction}
\label{sec:introduction}

\begin{figure}[h]
  \centering
  \includegraphics[width=1.0\textwidth]{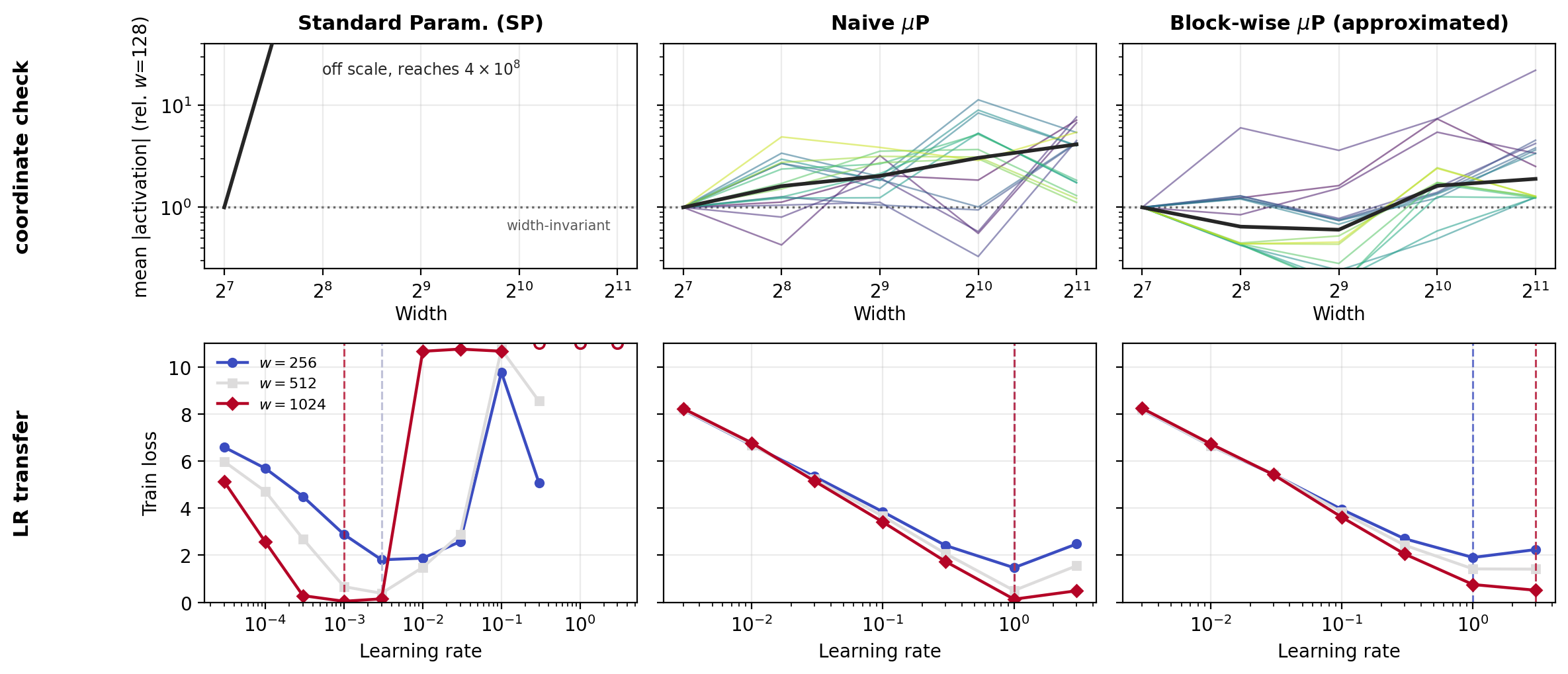}
  \caption{The coordinate check and LR transfer disagree on hybrids (\MHMH, $d{=}16$).
  \textbf{Top:} mean absolute activation per block after 10 steps on \textsc{WikiText-103}, relative to $w{=}128$; dotted line is width invariance, bold line the median over blocks.
  \textbf{Bottom:} training loss against learning rate on \textsc{WikiText-2} (2k steps), each width's optimum dashed and diverged runs open.
  The diagnostic separates SP from both $\muP$ variants, while transfer separates \naive $\muP$, which holds $\eta^*{=}1.0$ at every width, from the other two.}
  \label{fig:overall_central_obs}
  \label{fig:overall_desc}
\vspace{-1.0em}
\end{figure}

Hybrid architectures combining Transformer attention with State-Space Model (SSM) blocks have emerged as a promising direction for large language models, pairing attention's quality on long-range dependencies with SSMs' subquadratic inference cost.
Controlled comparisons at 8B parameters show Mamba-2-Hybrid outperforming pure Transformers across benchmarks while running up to $8\times$ faster~\citep{waleffe2024}, and similar trends appear in recent production hybrids~\citep{nvidia2025nemotron,lieber2024jamba,zuo2025falcon}; a fundamental practical problem remains, however: \emph{training these models requires expensive per-scale hyperparameter tuning.}

The learning rate (LR) must be retuned at each model scale because the optimum shifts with architectural choices like width and depth; at production scale this sweep can substantially exceed the single-run training budget~\citep{yang2022tensor}.
Maximal Update Parameterization ($\muP$)~\citep{yang2022tensor} solves this for pure Transformers by scaling initialization and LR so the optimum LR becomes width-independent.
\citet{vankadara2024feature} showed that SSM operations fall outside the Tensor Programs framework and derived component-specific rules ($\muP$-SSM), validated on pure SSMs.
To our knowledge, no prior work addresses learning rate transfer for \textbf{hybrid architectures} at \textbf{practical finite widths}.

Without any SSM-specific extension, the original $\muP$ prescription (which we refer to as \emph{naïve} $\muP$) under AdamW~\citep{Loshchilov2017DecoupledWD} achieves \textbf{near-zero learning rate transfer gap} across all tested hybrid architectures, up to $8\times$ width (256 to 2048) and $8\times$ depth (4 to 32).%
\footnote{Measured on the 7-point grid $\eta \in \{0.003, 0.01, 0.03, 0.1, 0.3, 1.0, 3.0\}$ where $\eta^*{=}1$ at every cell; a finer sweep (Appendix~\ref{sec:appendix:width_transfer}) adds intermediate points $\{0.5, 0.7, 1.5, 2.0\}$ and shows the basin between $\eta{=}0.5$ and $\eta{=}1.5$.}
Strikingly, the standard coordinate-check diagnostic does not predict LR transfer in this regime (Figure~\ref{fig:overall_central_obs}): every parameterization we test fails the coordinate check on hybrids, yet \naive $\muP$ transfers cleanly while $\muP$-SSM~\citep{vankadara2024feature} (designed for pure SSMs at infinite width under SGD with true-ZOH and growing $N_x$) does not. Specifically, $\muP$-SSM's $\eta^*$ drifts with width on every SSM-bearing architecture (Table~\ref{tab:width_transfer}), and at fixed $\eta$ it produces a $5$--$8\%$ higher convergence loss than \naive $\muP$ on \textsc{WikiText-103}.
Section~\ref{sec:analysis} interprets this observation in a two-condition decomposition: \naive $\muP$ delivers a global update-to-weight invariance, and AdamW's per-parameter normalization $g/\sqrt{v}$ delivers the local per-component balance directly across the SSM boundary, while $\muP$-SSM's explicit per-component LR rules do not deliver this balance in our regime (an optimizer ablation at a fixed parameterization supports this attribution).

Our contributions are threefold:
\begin{enumerate}
    \item \textbf{We empirically establish} that \naive $\muP$ with AdamW achieves near-zero LR transfer gap across hybrid architectures, widths 256--2048 ($8\times$), and depths 4--32 up to billion-parameter scale (Section~\ref{sec:results}; Appendix~\ref{sec:appendix:scale}).
    \item \textbf{We propose a two-condition decomposition} of LR transfer (global update-to-weight invariance, from $\muP$; local per-component balance, which Adam satisfies automatically), grounded by a per-component scaling test showing Vankadara's prescription works at initialization for $B$ and $C$ but fails for $A_{\log}$ under the field-standard simplified-ZOH Mamba implementation (Section~\ref{sec:analysis}, Appendix~\ref{sec:appendix:gradient}).
    \item \textbf{We validate generalization} to a production hybrid model~\citep{nvidia2025nemotron}, which transfers at 0.0\% gap without architecture-specific tuning (Section~\ref{sec:results:nemotron}).
\end{enumerate}

\section{Background and Related Work}
\label{sec:background}

\subsection{Maximal Update Parameterization ($\muP$)}

Under standard parameterization (SP), weight matrices are initialized as $W \!\sim\! \mathcal{N}(0, 1/\sqrt{d})$ per element with a single learning rate $\eta$ shared across all parameters; as fan-in width $d$ grows, the update-to-weight ratio $\|\Delta W\|_F / \|W\|_F$ shifts with $d$ and the optimal LR has to be re-tuned at each target scale.
\citet{yang2022tensor} introduced Maximal Update Parameterization ($\muP$) within the Tensor Programs framework, which scales hidden-layer initialization to $W \!\sim\! \mathcal{N}(0, 1/d)$ and learning rates to $\eta_{\text{base}}/d$ so this ratio stays $\mathcal{O}(1)$ as $d \!\to\! \infty$, allowing the LR found at a small \emph{proxy} width to transfer directly to the target.
The Tensor Programs framework gives a rigorous infinite-width limit for architectures composed of \emph{representable} operations, and it was validated for pure Transformers up to 6.7B parameters~\citep{yang2022tensor} and studied empirically up to 10B~\citep{lingle2024empirical}.
\citet{yang2024tp_ivb} extended the theory to Adam~\citep{kingma2015adam} and other entrywise optimizers, characterizing $\muP$ as the unique maximal-update (feature-learning) limit for these optimizers.
Subsequent works extend $\muP$ to other architectures and optimizer families~\citep{dey2024sparse,zheng2025scaling,malasnicki2025moe,qiu2025hyperparameter}; \citet{hayou2025proof} gives the first formal width-transfer proof for linear MLP, and \citet{noci2024why} provide a complementary spectral analysis showing $\muP$ induces a \emph{super consistency} of the loss-landscape sharpness across widths.

\subsection{$\muP$-SSM: Scaling Rules for State-Space Models}
\label{sec:background:mup_ssm}

State space models are the family of architectures built on continuous-time linear dynamical systems \citep{gu2020hippo,gu2022s4}, represented by Mamba~\citep{gu2023mamba}.
\citet{vankadara2024feature} showed that three operations in the Mamba block violate the representability conditions of the Tensor Programs framework, so standard $\muP$ does not guarantee width-independent dynamics for SSMs.
They derived SSM-specific scaling rules ($\muP$-SSM) such that each component's contribution to the output remains constant with width; their analysis establishes that SSMs fail the feature learning condition under \emph{both} SGD and Adam in the infinite-width limit, with empirical validation under SGD on pure S4 and Mamba.

Three gaps motivate extending $\muP$ to hybrid architectures.
\textbf{(1)~Pure-SSM only}: \citet{vankadara2024feature}'s Assumption~3.1 takes all non-SSM components to already follow $\muP$, focusing the analysis on the SSM layer alone; the interaction between well-parameterized Transformer blocks and SSM blocks in a hybrid was not tested.
\textbf{(2)~Infinite-width asymptotic}: their claims hold as $N_u, N_x \!\to\! \infty$ (where $N_u$ is the SSM operating width and $N_x$ the state dimension), and the asymptotic failure may not manifest at the practical finite widths (256--2048) we test, particularly when AdamW's per-parameter normalization ($g/\sqrt{v}$) compensates.
\textbf{(3)~True-ZOH with growing $N_x$}: the derivation assumes the true zero-order-hold discretization with state size $N_x \!\to\! \infty$, while the field-standard Mamba implementation~\citep{gu2023mamba} uses a simplified-ZOH at fixed $N_x{=}16$ regardless of width.

\subsection{Adam and Hyperparameter Transfer}
\label{sec:background:adam}

\citet{tomihari2025gradient} prove the formal mechanism on pure Transformers: Adam's coordinate-wise normalization makes updates depend on gradient signs, conferring robustness to cross-block heterogeneity.
We show this mechanism extends across the Transformer-SSM boundary in our setting, where gradient magnitudes differ widely across SSM and Transformer components (Appendix~\ref{sec:appendix:gradient}).
\citet{kosson2025weight} identify weight decay as the primary stabilizer for pure Transformers at 20B-token scale. In our hybrid ablation, removing weight decay changes the loss by at most $0.4\%$ while $g/\sqrt{v}$ is essential.\footnote{The ablation spans 2k steps, which is shorter than the weight-decay timescale $1/(\eta\lambda)$ at our target widths. It therefore does not cover the longer training horizons examined by \citet{kosson2025weight}.}
\citet{mlodozeniec2025completed} extend HP transfer to other axes (depth, batch size, duration, per-module HP); our batch-size variation gap of $\leq\!1.1\%$ is consistent with this broader picture.
\citet{everett2024scaling} show that per-layer LR scaling under SP can match $\muP$ on Transformers, indicating that multiple paths to width-invariant LR exist; we focus on the $\muP$-based path under AdamW for hybrid architectures.

\subsection{Hybrid Architectures and Production Training}
\label{sec:background:gap}

Hybrid Transformer-SSM models are increasingly used in production language models~\citep{nvidia2025nemotron,lieber2024jamba,dong2024hymba,zuo2025falcon} and studied as competitive alternatives to pure Transformers in controlled comparisons~\citep{waleffe2024}; production hybrid training, however, generally relies on per-scale hyperparameter tuning rather than $\muP$-style transfer.
Falcon-H1~\citep{zuo2025falcon} is the closest concurrent work that tunes hybrids at scale: they jointly optimize 35 per-component multipliers (LRs, weight decay, and forward-pass activation scales) for hybrid architectures up to 34B parameters.
Their stated motivation is production training stability (e.g., scaling the $\Delta$-projection to prevent its explosion) and per-component performance tuning, not $\muP$-style LR transfer; their work and ours therefore solve different problems, with our work isolating the minimal sufficient ingredients for the LR-transfer subproblem under AdamW.
Nemotron-H~\citep{nvidia2025nemotron}, which we revisit in Section~\ref{sec:results:nemotron}, similarly relies on per-scale hyperparameter tuning despite using a different hybrid composition (Mamba-2~\citep{dao2024mamba2} SSD blocks with $\sim$8\% attention layers and squared-ReLU FFN).

\section{Experimental Setup}
\label{sec:setup}

We empirically validate $\muP$ on hybrid Transformer-SSM architectures, focusing on LR transfer across widths while maintaining training stability.
Production hybrid architectures use a wide range of mixing patterns without consensus on a canonical design (Section~\ref{sec:background:gap}), and training dynamics depend on block composition.
Therefore we first probe LR transferability at heterogeneous block interfaces, where training dynamics are most likely to shift; then across depth and width; and finally in larger training runs with downstream evaluation.

\subsection{Hybrid Transformer-SSM Architectures}

We study models represented by a \texttt{layer\_pattern} string where each character selects a block type:
\textbf{Transformer Block (T)} is from the original Transformer architecture~\citep{vaswani2017attention} with grouped-query attention~\citep{ainslie2023gqa}, rotary position embeddings~\citep{su2021roformer}, and a SwiGLU FFN~\citep{shazeer2020glu}. This follows the LLaMA-style design~\citep{touvron2023llama}, which is the most widely used Transformer architecture in practice. 
\textbf{Mamba Block (M)} is a selective state-space model (S6) with causal conv1d from the original Mamba architecture~\citep{gu2023mamba}\footnote{We used the original version of \citet{gu2023mamba}, but an additional experiment empirically showed that it is also applicable to Mamba-2 blocks~\citep{dao2024mamba2} (Appendix~\ref{sec:appendix:arch_variants}).}.
\textbf{Hybrid Block (H)} is a parallel fusion of Transformer and Mamba sub-blocks with additive merge. This block is an example of the parallel fusion of attention and SSM operations with minimal configuration, but still it represents the architectural family of \citet{zuo2025falcon, dong2024hymba} that contains parallel branches.
Our architecture set pairs pure baselines (\MM, \TT) with a serial hybrid (\MTMT), parallel hybrids at half and full H-block density (\MHMH, \HMHM, \HHHH), and a hybrid-Transformer mixture (\HTHT); the depth sweep adds the reversed orders \TMTM and \THTH.
\MTMT and \MHMH place M first, consistent with \citet{cooper2026expressivity}; the reversed orders are included for transfer coverage, not to compare orderings.
We conduct additional experiments varying the serial/parallel Transformer-to-SSM block ratio to check whether different ratios affect the LR-transfer behavior; and no effect was found. See Appendix~\ref{sec:appendix:jamba_imbalanced},~\ref{sec:appendix:tm_ratio} for details.

\subsection{Parameterization Schemes}
\label{sec:setup:schemes}

We compare four schemes: Standard Parameterization (SP), \Naive $\muP$, Block-wise $\muP$, and Block-wise $\muP$ (approximated).

\textbf{Standard Parameterization (SP)}: weight matrices are initialized as $W \sim \mathcal{N}(0, 1/\sqrt{\dmodel})$ per element, with a single shared learning rate $\eta$ across all parameters, where $\dmodel$ is the model hidden dimension.
\textbf{\Naive $\muP$}: embeddings $\mathcal{N}(0,1)$, hidden layers $\mathcal{N}(0, 1/\dmodel)$, output head zeros; hidden-layer LR scaled by $\eta_{\text{base}} / \dmodel$.
\textbf{Block-wise $\muP$}: \naive $\muP$ on Transformer components plus \citet{vankadara2024feature}'s Table~1 SSM prescription (per-component init scales $\sigma_B, \sigma_C$ and learning rates $\eta_A, \eta_B, \eta_C$ in $N_u$ (SSM operating width) and $N_x$ (state dim); explicit formulas in Appendix~\ref{sec:appendix:a_scaling}).

We also use the \textbf{Block-wise $\muP$ (approximated)} variant, which keeps the per-component initialization and the $\eta_C$ scaling of \citet{vankadara2024feature} but scales $\eta_A$ and $\eta_B$ with the \naive-$\muP$ exponent (Appendix~\ref{sec:appendix:a_scaling}; motivation in Section~\ref{sec:analysis:redundant_normalization}).
Both block-wise variants use the SSM discretization from the official Mamba library~\citep{gu2023mamba}, which differs from the true-ZOH form underlying \citet{vankadara2024feature}'s Table~1 derivation; we discuss the resulting regime gaps in Appendix~\ref{sec:appendix:vankadara_faithful}.

\subsection{Training Details}

All experiments use AdamW~\citep{Loshchilov2017DecoupledWD} ($\beta_1{=}0.9$, $\beta_2{=}0.95$), cosine warmup, gradient clipping at 1.0, and a fixed seed.
We use \textsc{WikiText-2}/\textsc{WikiText-103}~\citep{merity2017pointer} (2.4M/120M tokens) for short-horizon and convergence runs, and \textsc{FineWeb-Edu}~\citep{penedo2024fineweb} (500M tokens) for larger-corpus validation; per-experiment hyperparameters appear in Section~\ref{sec:results}.
\textsc{WikiText-103} carries the controlled convergence sweeps and \textsc{FineWeb-Edu} tests the recipe on more realistic web-scale data; since $\muP$'s transfer is a statement at fixed data, we make no cross-dataset LR-optimality claim.
Each LR sweep uses a single tokenizer: GPT-2 BPE on \textsc{WikiText-2} and tiktoken \texttt{cl100k\_base} on \textsc{WikiText-103} and \textsc{FineWeb-Edu} (Appendix~\ref{sec:appendix:dataset_coverage}); the baseline-matched comparison of Table~\ref{tab:downstream} additionally uses the GPT-NeoX tokenizer, and its Pythia and Mamba baselines share GPT-NeoX's initialization and differ in optimizer settings.
Unless a sweep varies them, our LR sweeps use sequence length 128 and batch size 64 for 2k or 20k steps (16M or 164M tokens), extended to 60k steps (492M tokens) at 1.24B parameters (Appendix~\ref{sec:appendix:scale}); the 5.0B-token run (Section~\ref{sec:results:width}) and the 24.9B-token run (Appendix~\ref{sec:appendix:scale}) use the transferred LR at sequence length 2048 and are convergence runs, not re-sweeps.
Robustness experiments are in Appendix~\ref{sec:appendix:robustness} and total project compute (\nGpuHrs\ GPU-hours, with the per-architecture tuning-cost saving) is in Appendix~\ref{sec:appendix:compute_budget}.

\subsection{Evaluation Metrics}

Our main metric is the \textbf{learning rate transfer gap}: the relative loss difference\footnote{Following \citet{yang2022tensor}, we measure transfer on training loss, which excludes the effect of regularization; appendix results evaluated on a held-out split report validation loss, as indicated in each case. Across all sweeps on \textsc{WikiText-103} and \textsc{FineWeb-Edu} that were checked on both training and validation loss, the two select the same $\eta^*$ in $209$ of $218$ cells. Where they differ, the validation loss at the training-loss optimum lies within $0.6\%$ of the validation minimum, except in the 60k-step runs of Appendix~\ref{sec:appendix:scale} (up to $1.7\%$). This difference compares the optima of the two losses within a single configuration and is not a transfer gap; the validation-loss transfer gap in those runs is $0.0\%$. On \textsc{WikiText-2}, which a 2k-step run traverses about seven times, validation loss prefers a lower LR than training loss ($0.1$ against $1.0$ for \naive $\muP$, where the grid resolves the optimum).} between training the target with the proxy's optimum LR versus its own optimum,
\begin{equation}
\text{Transfer Gap} = \frac{\bigl|\mathcal{L}(\eta^*(c_{\text{proxy}}); c_{\text{target}}) - \mathcal{L}(\eta^*(c_{\text{target}}); c_{\text{target}})\bigr|}{\bigl|\mathcal{L}(\eta^*(c_{\text{target}}); c_{\text{target}})\bigr|},
\label{eq:lr_transfer_gap}
\end{equation}
where $\mathcal{L}(\eta; c)$ is the end-of-training loss, $\eta^*(c) = \arg\min_\eta \mathcal{L}(\eta; c)$, and $c$ is the configuration (architecture, width, depth, etc.).
If the transfer gap is 0\%, it indicates the proxy's optimal LR is also optimal at the target within the tested LR resolution.

\section{Results}
\label{sec:results}

We validate \naive $\muP$ LR transfer through stages of increasing complexity: unit-level interface checks and depth scaling in Section~\ref{sec:results:unit}, width transfer across architectures and scaling to a larger dataset in Section~\ref{sec:results:width}, and generalization to a production architecture in Section~\ref{sec:results:nemotron}.
Full per-experiment details are in Appendices~\ref{sec:appendix:unit_interface},~\ref{sec:appendix:width_transfer}, and~\ref{sec:appendix:depth_width}; this section summarizes the main findings.

\subsection{Unit Interface and Depth Validation}
\label{sec:results:unit}

Hybrid architectures compose heterogeneous block types, creating interfaces (e.g., Mamba$\to$Transformer, Hybrid$\to$Mamba) from distinct signal propagation dynamics per block.
We enumerate all 9 unit interfaces constructible from T, M, and H blocks and measure training stability across widths 128--2048 for each parameterization scheme.

The stability window of \naive $\muP$ extends beyond the highest LR tested in this sweep ($\eta{=}1.0$) across all widths and interfaces.\footnote{Extended sweeps (Table~\ref{tab:full_sgd_vs_adamw}) confirm \naive $\muP$ trains stably at $\eta{=}3$ on \MHMH and \HHHH, and at $\eta{=}10$ on \MHMH.}
The stability window of SP shrinks with width, dropping from $0.30$--$1.0$ at $w{=}128$ to $0.03$--$0.10$ at $w{=}2048$.
Block-wise $\muP$ (approximated) with $\eta_A$ instead set to $\eta_{\text{base}} N_u$ collapses to $\eta{\leq}0.01$ at $w{\geq}256$ for serial SSM interfaces (\MM, \texttt{MT}); Section~\ref{sec:analysis:redundant_normalization} interprets this as a redundant per-component correction stacking with Adam's $g/\sqrt{v}$.
Full stability tables are in Appendix~\ref{sec:appendix:unit_interface}.

Across depths $4$--$32$ on 8 architectures (\MM, \TT, \MTMT, \MHMH, \HHHH, \HTHT, \TMTM, \THTH; spanning pure SSM/Transformer, serial hybrids, and parallel/reversed compositions) at $w{=}512$, \naive $\muP$ keeps $\eta^*{=}1.0$ in all 32 cells and the approximated block-wise variant in 23, moving to $\eta^*{=}3.0$ on \MM{} at every depth and in five hybrid cells at $d{\leq}8$; full block-wise drifts to $\eta^*{=}3.0$ in 25 of the 28 SSM-bearing cells (consistent with $\eta_A \!\propto\! N_u$) and destabilizes at $\eta \!\geq\! 10$ in all 28 (Appendix~\ref{sec:appendix:depth_scale}).
The approximated variant's 20k-step loss on \textsc{WikiText-103} stays within $0.5\%$ of \naive at $w{=}1024$ (Appendix~\ref{sec:appendix:vankadara_faithful}; interpreted in Section~\ref{sec:analysis:redundant_normalization}).

\subsection{Width Transfer and Scale Validation}
\label{sec:results:width}

We train 7 architectures (\MM, \TT, \MTMT, \MHMH, \HMHM, \HHHH, \HTHT; spanning pure SSM/Transformer baselines, serial and parallel hybrids of varying H-block density) at widths 256, 512, 1024 across all four schemes (Table~\ref{tab:width_transfer}; 20k-step \textsc{WikiText-103} convergence in Appendix~\ref{sec:appendix:vankadara_faithful}).

\Naive $\muP$ is the unique scheme that transfers cleanly: $\eta^*{=}1.0$ at all widths for all 7 architectures, while the other three drift with width.\footnote{The discretization used in the official Mamba library~\citep{gu2023mamba} differs from the ZOH or Euler form underlying \citet{vankadara2024feature}'s Table~1 derivation; see Section~\ref{sec:setup:schemes}.}
We therefore test the joint depth-width grid only under \naive $\muP$, building on its zero transfer gap on width (Table~\ref{tab:width_transfer}) and depth (Section~\ref{sec:results:unit}) individually.  %

\definecolor{lr3}{gray}{1.00}      %
\definecolor{lr1}{gray}{0.95}      %
\definecolor{lr01}{gray}{0.85}     %
\definecolor{lr003}{gray}{0.80}    %
\definecolor{lr001}{gray}{0.70}    %
\definecolor{lr0003}{gray}{0.55}   %
\definecolor{lr00003}{gray}{0.40}  %

\begin{table}[t]
\centering
\small
\caption{Optimal LR at each (scheme, architecture, width); uniform shading within each row means LR transfer to the corresponding widths. Values from 2k-step training on \textsc{WikiText-2}. The SP grid for \MM and \MTMT starts at $\eta{=}0.003$.}
\label{tab:width_transfer}
\resizebox{\columnwidth}{!}{%
\begin{tabular}{l ccc ccc ccc ccc}
\toprule
 & \multicolumn{3}{c}{\Naive $\muP$} & \multicolumn{3}{c}{Block-wise $\muP$ (approximated)} & \multicolumn{3}{c}{Block-wise $\muP$} & \multicolumn{3}{c}{SP} \\
\cmidrule(lr){2-4} \cmidrule(lr){5-7} \cmidrule(lr){8-10} \cmidrule(lr){11-13}
Arch & $w{=}256$ & $w{=}512$ & $w{=}1024$ & $w{=}256$ & $w{=}512$ & $w{=}1024$ & $w{=}256$ & $w{=}512$ & $w{=}1024$ & $w{=}256$ & $w{=}512$ & $w{=}1024$ \\
\midrule
\MM   & \cellcolor{lr1}\textbf{1.0} & \cellcolor{lr1}\textbf{1.0} & \cellcolor{lr1}\textbf{1.0} & \cellcolor{lr1}1.0 & \cellcolor{lr3}3.0 & \cellcolor{lr3}3.0 & \cellcolor{lr3}3.0 & \cellcolor{lr3}3.0 & \cellcolor{lr1}1.0 & \cellcolor{lr01}0.010 & \cellcolor{lr003}0.003 & \cellcolor{lr003}0.003 \\
\TT   & \cellcolor{lr1}\textbf{1.0} & \cellcolor{lr1}\textbf{1.0} & \cellcolor{lr1}\textbf{1.0} & \cellcolor{lr1}1.0 & \cellcolor{lr1}1.0 & \cellcolor{lr1}1.0 & \cellcolor{lr1}1.0 & \cellcolor{lr1}1.0 & \cellcolor{lr1}1.0 & \cellcolor{lr003}0.003 & \cellcolor{lr001}0.001 & \cellcolor{lr0003}0.0003 \\
\MTMT & \cellcolor{lr1}\textbf{1.0} & \cellcolor{lr1}\textbf{1.0} & \cellcolor{lr1}\textbf{1.0} & \cellcolor{lr1}1.0 & \cellcolor{lr1}1.0 & \cellcolor{lr1}1.0 & \cellcolor{lr1}1.0 & \cellcolor{lr3}3.0 & \cellcolor{lr1}1.0 & \cellcolor{lr01}0.010 & \cellcolor{lr003}0.003 & \cellcolor{lr003}0.003 \\
\MHMH & \cellcolor{lr1}\textbf{1.0} & \cellcolor{lr1}\textbf{1.0} & \cellcolor{lr1}\textbf{1.0} & \cellcolor{lr1}1.0 & \cellcolor{lr3}3.0 & \cellcolor{lr3}3.0 & \cellcolor{lr3}3.0 & \cellcolor{lr3}3.0 & \cellcolor{lr1}1.0 & \cellcolor{lr003}0.003 & \cellcolor{lr003}0.003 & \cellcolor{lr001}0.001 \\
\HMHM & \cellcolor{lr1}\textbf{1.0} & \cellcolor{lr1}\textbf{1.0} & \cellcolor{lr1}\textbf{1.0} & \cellcolor{lr1}1.0 & \cellcolor{lr1}1.0 & \cellcolor{lr3}3.0 & \cellcolor{lr3}3.0 & \cellcolor{lr3}3.0 & \cellcolor{lr1}1.0 & \cellcolor{lr003}0.003 & \cellcolor{lr003}0.003 & \cellcolor{lr001}0.001 \\
\HHHH & \cellcolor{lr1}\textbf{1.0} & \cellcolor{lr1}\textbf{1.0} & \cellcolor{lr1}\textbf{1.0} & \cellcolor{lr1}1.0 & \cellcolor{lr1}1.0 & \cellcolor{lr3}3.0 & \cellcolor{lr3}3.0 & \cellcolor{lr3}3.0 & \cellcolor{lr1}1.0 & \cellcolor{lr003}0.003 & \cellcolor{lr003}0.003 & \cellcolor{lr001}0.001 \\
\HTHT & \cellcolor{lr1}\textbf{1.0} & \cellcolor{lr1}\textbf{1.0} & \cellcolor{lr1}\textbf{1.0} & \cellcolor{lr1}1.0 & \cellcolor{lr1}1.0 & \cellcolor{lr1}1.0 & \cellcolor{lr3}3.0 & \cellcolor{lr1}1.0 & \cellcolor{lr1}1.0 & \cellcolor{lr003}0.003 & \cellcolor{lr003}0.003 & \cellcolor{lr001}0.001 \\
\midrule
\multicolumn{1}{l}{Transfers?} & \multicolumn{3}{c}{\textbf{Yes} (0/7 shift)} & \multicolumn{3}{c}{Partially (4/7 shift)} & \multicolumn{3}{c}{Drifts (6/6 SSM-bearing shift)} & \multicolumn{3}{c}{No (7/7 shift)} \\
\bottomrule
\end{tabular}%
}
\vspace{-1.0em}
\end{table}

From the joint depth-width experiment, \naive $\muP$ shows $\eta{=}1.0$ transfer from a proxy ($w{=}256$, $d{=}8$, ${\sim}58$M params for \MHMH; ${\sim}61$M for \HHHH) to the full target ($w{=}1024$, $d{=}32$, ${\sim}624$M params for \MHMH; ${\sim}830$M for \HHHH), an approximate $11\times$ (\MHMH) to $14\times$ (\HHHH) parameter increase with 0.0\% gap for both architectures (Appendix~\ref{sec:appendix:depth_width}).
We further validate on \textsc{FineWeb-Edu} (500M tokens, 20k steps) to test whether LR transfer holds on a curated web-scale corpus. %
Across the 3 tested architectures, \naive $\muP$ achieves 0.0\% transfer gap, while SP's optimum shifts $3\times$ smaller ($0.003 \!\to\! 0.001$) and transferring the proxy LR without retuning incurs $1.1$--$8.7\%$ penalty (Appendix~\ref{sec:appendix:fineweb_detail}).

\paragraph{Downstream evaluation.}
We train \MHMH on \textsc{FineWeb-Edu} for 38k steps (5.0B tokens, ${\sim}0.6\times$ Chinchilla-optimal~\citep{hoffmann2022chinchilla}) using $\eta{=}1.0$ transferred from the proxy ($w{=}256$, $d{=}16$), reporting \MHMH-415M (tiktoken \texttt{cl100k\_base}) and \MHMH-312M (GPT-NeoX, matching Pythia~\citep{biderman2023pythia}); we retrain Pythia-410M and Mamba-370M~\citep{gu2023mamba} on the same data for comparison.
The goal is not to claim architectural superiority but to show that \textbf{the transferred LR produces a competitive model without retuning} (Table~\ref{tab:downstream}). Widths up to $w{=}2048$ are also LR-transfer validated at the 2k-step horizon (Appendix~\ref{sec:results:upto_w2048}).
Following the standard zero-shot suite for sub-billion-parameter language models~\citep{biderman2023pythia,gu2023mamba}, we evaluate on HellaSwag, PIQA, WinoGrande, ARC, and LAMBADA~\citep{zellers2019hellaswag,bisk2020piqa,sakaguchi2021winogrande,clark2018think,paperno2016lambada}.

\begin{table}[t]
  \centering
  \small
  \caption{Zero-shot downstream evaluation.
  Top: published baselines (300B tokens, The Pile).
  Bottom: all models retrained on \textsc{FineWeb-Edu} 5B tokens.
  \MHMH variants: $\dagger$ \MHMH-312M uses the GPT-NeoX tokenizer (50304 vocab, tokenizer-matched to Pythia); $\ddagger$ \MHMH-415M uses tiktoken \texttt{cl100k\_base} (100277 vocab). } 
  \label{tab:downstream}
  \begin{tabular}{lrccccccc}
    \toprule
    Model & Params & HellaSwag & PIQA & WinoGr & ARC-E & ARC-C & LAMBADA & Avg \\
    & & \scriptsize{acc\_n} & \scriptsize{acc} & \scriptsize{acc} & \scriptsize{acc} & \scriptsize{acc\_n} & \scriptsize{acc} & \\
    \midrule
    \multicolumn{9}{l}{\textit{Published baselines (The Pile, 300B tokens)}} \\
    Mamba-370M & 370M & 46.5 & 69.5 & 55.3 & 55.1 & 28.0 & 55.6 & 51.7 \\
    Pythia-410M & 410M & 40.6 & 66.9 & 53.8 & 52.1 & 24.6 & 51.4 & 48.2 \\
    \midrule
    \multicolumn{9}{l}{\textit{Retrained on FineWeb-Edu (5B tokens)}} \\
    Mamba-370M & 370M & 34.6 & 61.6 & \textbf{53.8} & 55.0 & 26.2 & 19.9 & 41.9 \\
    Pythia-410M & 410M & 33.1 & 63.2 & 50.4 & 54.3 & 26.3 & 27.2 & 42.4 \\
    \MHMH-\textbf{312M}$^{\dagger}$ & \textbf{312M} & \textbf{35.1} & 64.1 & 52.8 & \textbf{57.2} & 26.1 & \textbf{29.2} & \textbf{44.1} \\
    \MHMH-\textbf{415M}$^{\ddagger}$  & \textbf{415M} & 34.3 & \textbf{64.3} & 51.9 & 55.9 & \textbf{26.5} & 27.7 & 43.4 \\
    \bottomrule
  \end{tabular}
\end{table}

\subsection{Generalization to Nemotron-H}
\label{sec:results:nemotron}

To test whether \naive $\muP$ generalizes beyond our simple custom architectures, we validate on Nemotron-H~\citep{nvidia2025nemotron}, a production hybrid using Mamba-2~\citep{dao2024mamba2}, squared-ReLU FFN, and grouped-query attention~\citep{ainslie2023gqa} without rotary embeddings~\citep{su2021roformer}.
We follow the Nemotron-H architecture: Mamba-2 (N), squared-ReLU FFN (F), and attention-only blocks (A) with GQA at ratio~4 and no RoPE (Appendix~\ref{sec:appendix:nemotron_blocks}). This architecture is designed with different block types than our set.

\begin{table}[t]
\centering
\small
\caption{Nemotron-H width transfer (\textsc{WikiText-103}, 20k steps, warmup=1000; $\eta \in \{1, 3, 5\}$ for \naive $\muP$, $\{0.001, 0.003, 0.01\}$ for SP).
\Naive $\muP$ achieves $\eta^*{=}1.0$ at all widths (matching our other architectures); SP optimum shifts from $\eta{=}0.003$ at $w{=}256$ to $\eta{=}0.001$ at $w{=}1024$.}
\label{tab:nemotron}
\begin{tabular}{llccccc}
\toprule
Depth & Scheme & Best LR & $w{=}256$ & $w{=}512$ & $w{=}1024$ & Gap \\
\midrule
16 & \Naive $\muP$ & \textbf{1.0} & 3.760 & 3.344 & \textbf{3.369} & \textbf{0.0\%} \\
16 & SP            & 0.001--0.003 & 3.775 & 3.382 & 3.356 & 1.9\% \\
32 & \Naive $\muP$ & \textbf{1.0} & 3.527 & 3.313 & \textbf{3.138} & \textbf{0.0\%} \\
32 & SP            & 0.001--0.003 & 3.542 & 3.350 & 3.152 & 1.9\% \\
\bottomrule
\end{tabular}
\vspace{-0.5em}
\end{table}

\Naive $\muP$ achieves $\eta^*{=}1.0$ across all (depth, width) combinations on Nemotron-H, matching the optimum found across our custom architectures. The transfer gap is $0.0\%$ at every cell, within a broad basin ($\eta{\in}\{1, 3, 5\}$ all within ${\sim}3\%$ of optimum) that makes transfer robust to LR misspecification.
The optimum also stays at $\eta{=}1.0$ from width 512 to 2048 (980M parameters) when the MLP ratio, SSM state, Mamba-2 group count and attention head dimension are those of the released Nemotron-H-8B configuration, and two runs at about 20 tokens per parameter (57M and 299M parameters) train without divergence (Appendix~\ref{sec:appendix:nemotron_blocks}).
SP's optimum shifts $3\times$ across width ($0.003 \!\to\! 0.001$), incurring a $1.9\%$ transfer penalty at both depths if the proxy LR is reused.

\section{Analysis}
\label{sec:analysis}

\subsection{\Naive $\muP$ Violates Coordinate Checks but Learning Rate Transfers}
\label{sec:analysis:coord_surprise}

$\muP$'s LR transfer is theoretically expected to hold only when the coordinate check is satisfied~\citep{yang2022tensor}, i.e., when pre-activation magnitudes remain width-invariant. However, on hybrid architectures, all four parameterizations we test fail the coordinate check (Figure~\ref{fig:coord_check_appendix}): activations vary with width across SP, \naive $\muP$, and both block-wise variants, more than in a pure Transformer (Table~\ref{tab:coord_check_spread}).
Despite this universal coordinate check failure, \naive $\muP$ achieves $0.0\%$ transfer gap across every architecture in Table~\ref{tab:width_transfer}, while block-wise $\muP$'s optimum LR drifts with width on every SSM-bearing architecture.
Coordinate checks therefore do not predict LR transfer on hybrids; Section~\ref{sec:analysis:two_conditions} instead proposes a two-condition decomposition as an explanatory diagnostic of where transfer holds and where it breaks.

\subsection{Two Conditions for Scalable Transfer in Hybrid Architectures}
\label{sec:analysis:two_conditions}

We propose a two-condition decomposition of LR transfer, drawing on existing theory and supported by an optimizer ablation.

\paragraph{Condition 1: Global update-to-weight ratio invariance.}

For a parameter matrix $W$, define $\rho(W) \;=\; \frac{\|\Delta W\|_F}{\|W\|_F}$; update-to-weight ratio invariance requires $\rho(W) = \Ocal(1)$ as $\dmodel \to \infty$ for every parameter group.
This is the design principle of $\muP$: \citet{yang2022tensor,yang2024tp_ivb} establish width-independent $\rho$, for SGD and Adam-style optimizers.
Under SP, $\rho = \Ocal(\eta\sqrt{d})$ grows with width, causing the optimal learning rate to shift with model width; under \naive $\muP$, $\rho = \Ocal(\eta)$ depends on the learning rate but is width-independent.
Section~\ref{sec:analysis:coord_surprise} showed that LR transfer holds in our hybrid setting despite a width-invariance violation at the activation level, motivating the second condition below.

\paragraph{Condition 2: Local per-component update balance.}
For any pair of parameter groups $W_X, W_Y$ at configuration $c$, let $\beta_{X/Y}(\eta;c) = \|\Delta W_X\|_F / \|\Delta W_Y\|_F$ denote the ratio of their per-step update magnitudes; a width-shift in $\beta$ means the optimization rebalances across components as the model scales, breaking the proxy's local optimality at the target.
Condition~2 requires that the proxy's optimum LR $\eta^*$ preserves this ratio at the target, with deviation
\begin{equation}
\Delta\beta_{X/Y}(c_{\text{proxy}} \!\to\! c_{\text{target}}) \;=\; \frac{\bigl|\beta_{X/Y}(\eta^*; c_{\text{target}}) - \beta_{X/Y}(\eta^*; c_{\text{proxy}})\bigr|}{\bigl|\beta_{X/Y}(\eta^*; c_{\text{proxy}})\bigr|},
\label{eq:cond2_gap}
\end{equation}
the update-balance analogue of Equation~\ref{eq:lr_transfer_gap}: Condition~2 holds when the per-component update ratio $\beta_{X/Y}$ is preserved between proxy and target widths.

\subsection{Optimizer Choice Under the Two Conditions}
\label{sec:analysis:optimizer_choice}

For Condition~1, \naive $\muP$ delivers $\rho = \Ocal(\eta)$ across widths under AdamW~\citep{yang2022tensor,yang2024tp_ivb}, so we treat Condition~1 as held under AdamW throughout this subsection.
For Condition~2, in contrast, the choice of optimizer determines which of two normalization mechanisms is active: per-component LR scaling, or per-parameter normalization.

\paragraph{Under SGD: per-component LR scaling.}
Under SGD, $\beta_{M/T} \propto (\eta_M/\eta_T)\,\gamma_{M/T}$, where $\gamma_{M/T} = \mathrm{RMS}(\nabla_{W_M} \Lcal) / \mathrm{RMS}(\nabla_{W_T} \Lcal)$ is the gradient-RMS ratio for a Mamba/Transformer parameter pair.
Width-dependence in $\gamma$ must therefore be canceled by per-component LR scaling, which is the role of \citet{vankadara2024feature}'s SSM rules in their derivation regime.

\paragraph{Under AdamW: per-parameter normalization.}
Under AdamW, $\eta_{\mathrm{eff}}(W) = \eta/\sqrt{v_W}$ normalizes each parameter's gradient by its own running RMS magnitude, so the update direction is dominated by gradient \emph{sign} rather than magnitude (\citealp{tomihari2025gradient}, Thm.~4.7--4.9; \citealp{balles2018dissecting,zhang2024transformers_adam}).
The ratio $\beta_{M/T}$ then decouples from $\gamma$, and \naive $\muP$ achieves $\Delta\beta \to 0$ without any per-component LR scaling (interpretive across the SSM boundary).
Table~\ref{tab:condition2_gap} confirms this and exposes a stacking failure: the (SGD, \naive $\muP$) row gives $\Delta\beta_{A_{\log}/\text{FFN}}{=}235$--$342\%$; AdamW closes the same pair to $20$--$21\%$ (\naive $\muP$ row); adding Vankadara's per-component LR scaling on top inflates it to $166$--$436\%$ (block-wise $\muP$ row), as the explicit $\eta_A \!\propto\! N_u$ correction stacks with the normalization AdamW already supplies.

\begin{table}[t]
\centering
\small
\caption{Condition~2 gap $\Delta\beta_{M/T}(c_{\text{proxy}}\!\to\!c_{\text{target}})$ in \%, with proxy width $128$.
The $A_{\log}/$FFN pair is the most width-divergent under \citet{vankadara2024feature}'s $\eta_A \!\in\! \Theta(N_u)$ rule; in\_proj$/$FFN is a near-isotropic Mamba projection.
\MHMH $d{=}16$, ssm\_expand$=$2, $N_x{=}16$, $\eta_{\text{base}}{=}1.0$; mean over steps $50$--$200$ and $3$ seeds.
Same-seed reruns of the AdamW rows leave a run-to-run floor of about $20\%$; in those rows, the \naive entries and the in\_proj$/$FFN entries lie within the floor, and the block-wise $A_{\log}/$FFN entries lie well above it.}
\label{tab:condition2_gap}
\begin{tabular}{llcccc}
\toprule
 & & \multicolumn{2}{c}{$A_{\log}/$FFN} & \multicolumn{2}{c}{in\_proj$/$FFN} \\
\cmidrule(lr){3-4} \cmidrule(lr){5-6}
Optimizer & Scheme & $128{\to}256$ & $128{\to}512$ & $128{\to}256$ & $128{\to}512$ \\
\midrule
AdamW & \Naive $\muP$ & 20\% & 21\% & 4\% & 5\% \\
AdamW & Block-wise $\muP$ & 166\% & 436\% & 15\% & 20\% \\
SGD   & \Naive $\muP$ & 235\% & 342\% & 0\% & 0\% \\
SGD   & Block-wise $\muP$ & 233\% & 835\% & 23\% & 31\% \\
\bottomrule
\end{tabular}
\vspace{-0.5em}
\end{table}

\begin{table}[t]
\centering
\small
\caption{Optimizer ablation at a fixed parameterization (\HHHH at $d{=}16$, \naive-$\muP$ parameter groups for every optimizer, 2k steps on \textsc{WikiText-103}, $\eta \in \{0.001, \dots, 1.0\}$; per-LR sweeps in Appendix~\ref{sec:appendix:optimizer}). Every best loss falls at $\eta{=}1.0$, the top of this grid. The last column counts the LRs at which $w{=}1024$ reaches a lower loss than $w{=}256$.}
\begin{tabular}{lccccc}
\toprule
Optimizer & Adaptive LR & Momentum & Best loss $w{=}256$ & $w{=}1024$ & $w{=}1024$ lower \\
\midrule
AdamW & Yes & Yes ($\beta_1{=}0.9$) & 4.392 & 4.099 & 7/7 \\
Adam no-momentum & Yes & No ($\beta_1{=}0$) & 4.470 & 4.465 & 7/7 \\
SGD + momentum & No & Yes (0.9) & 7.732 & 7.955 & 1/7 \\
SGD no-momentum & No & No & 9.398 & 10.803 & 2/7 \\
\bottomrule
\end{tabular}
\vspace{-0.5em}
\label{tab:optimizer_ablation}
\end{table}

\paragraph{The block-wise with SGD anomaly and its diagnosis.}
Symmetrically, applying Vankadara's prescription under SGD, the configuration the prescription was designed for, also produces $\Delta\beta$ above $100\%$ on the $A_{\log}/$FFN pair (Table~\ref{tab:condition2_gap}, last row).
We diagnose this at the parameter granularity Vankadara's derivation assumes ($B$ and $C$ as separate parameter groups; Appendix~\ref{sec:appendix:gradient}, Table~\ref{tab:vankadara_step1_split}).
At step 1 under SGD, the faithful prescription delivers the predicted width-invariant effective update for $B$ ($\alpha(N_u){=}{+}0.006$) and $C$ ($\alpha{=}{+}0.029$): $\eta_B$ and $\eta_C$ act as designed.
For $A_{\log}$, however, the faithful update has $\alpha{=}{+}1.00$: the $\eta_A \!\propto\! N_u$ rule does not cancel an underlying $1/N_u$ gradient decay (no such decay is present in our gradients) but instead \emph{amplifies} the existing $N_u$-dependence already at step~1.
This reflects the regime gap with Vankadara's derivation: we use SGD and their per-component granularity, but use the field-standard simplified ZOH at fixed $N_x{=}16$ on a hybrid (Appendix~\ref{sec:appendix:vankadara_faithful}).
Even where the prescription does work at init (for $B$ and $C$), the invariance is destroyed during training: $\Delta\alpha$ from step 1 to the stable window reaches $+2.4$ ($A_{\log}$), $+1.34$ ($B$), $+0.68$ ($C$), all larger than \naive's drift ($\Delta\alpha \!\leq\! +0.7$).

\begin{table}[t]
\centering
\small
\caption{Effective-update $N_u$ scaling at step 1 vs the stable window (steps 100--199) under SGD, base LR $\eta\!=\!1$. Vankadara's framework predicts $\alpha(N_u)\!\approx\!0$ at initialization for the faithful prescription (width-invariant update). The faithful values for $B$ and $C$ at step~1 are bolded; $A_{\log}$ is bolded to flag the \emph{predicted-but-not-observed} cancellation.}
\label{tab:vankadara_step1_split}
\begin{tabular}{l l ccc}
\toprule
Component & Scheme & $\alpha(N_u)$ step 1 & $\alpha(N_u)$ stable & $\Delta\alpha$ \\
\midrule
\multirow{2}{*}{$A_{\log}$} & \naive   & $+0.86$ & $+1.55$ & $+0.68$ \\
                            & faithful & $\mathbf{+1.00}$ & $+3.42$ & $+2.42$ \\
\multirow{2}{*}{$B$}        & \naive   & $+0.99$ & $+1.33$ & $+0.34$ \\
                            & faithful & $\mathbf{+0.006}$ & $+1.35$ & $+1.34$ \\
\multirow{2}{*}{$C$}        & \naive   & $+1.00$ & $+1.34$ & $+0.33$ \\
                            & faithful & $\mathbf{+0.029}$ & $+0.71$ & $+0.68$ \\
\bottomrule
\end{tabular}
\vspace{-0.5em}
\end{table}

\paragraph{AdamW supplies Condition~2 in our regime; Vankadara's rules do not.}
\label{sec:analysis:redundant_normalization}
The diagnosis above identifies the ZOH regime gap as the reason Vankadara's prescription does not deliver Condition~2: under SGD, $\eta_A \!\propto\! N_u$ does not cancel its target gradient decay (no such decay is present in our gradients) and the init-level $B/C$ invariance does not survive training; under AdamW, the rules stack on top of the normalization $g/\sqrt{v}$ already supplies, producing the over-correction in Table~\ref{tab:condition2_gap}. This motivates our approximated variant (Section~\ref{sec:setup}, Appendix~\ref{sec:appendix:a_scaling}), which keeps the $\muP$-SSM initialization but drops the $\eta_A \!\in\! \Theta(N_u)$ width-amplification.
Under SGD with \naive $\muP$, the best LR sits at the top of our 20k-step grid ($\eta{=}100$, Table~\ref{tab:full_sgd_vs_adamw}) at both widths, so the grid cannot establish transfer. Condition~2 can flag such a sweep: $\Delta\beta_{A_{\log}/\text{FFN}}$ is $235$--$342\%$ under SGD against $20$--$21\%$ under AdamW (Table~\ref{tab:condition2_gap}). An extended \HHHH{} grid (Appendix~\ref{sec:appendix:sgd_extended}) places the optimum above $\eta{=}100$ at both widths, and $\eta{=}3000$ trains $w{=}1024$ but diverges at $w{=}256$.

\subsection{Summary} 

The decomposition isolates two distinct failure modes of LR transfer in hybrid architectures: Condition~1 violations shift the optimum LR with width (the canonical $\muP$ failure mode); Condition~2 violations rebalance components across widths (a finite-width hybrid-specific failure mode).
In our regime (simplified-ZOH Mamba at fixed $N_x{=}16$), the $\muP$-SSM scaling rules do not show evidence of delivering Condition~2 under either optimizer; under SGD the rule's intended cancellation does not apply, and under AdamW the rules add scaling on top of the normalization $g/\sqrt{v}$ already supplies. AdamW's $g/\sqrt{v}$ alone delivers Condition~2 across the SSM boundary. Whether either picture changes at infinite width or under true-ZOH SSMs with growing $N_x$ remains open.

In this section, we extracted two conditions from existing theory, and neither covers our hybrid setting under our regime. We use the two conditions as explanatory diagnostics and support this empirically (Tables~\ref{tab:condition2_gap}--\ref{tab:vankadara_step1_split}, Section~\ref{sec:analysis:coord_surprise}; Appendices~\ref{sec:appendix:coord_check}--\ref{sec:appendix:optimizer}).

\section{Conclusion}
\label{sec:conclusion}

We demonstrate that \naive $\muP$ with AdamW yields LR transfer across widths (up to billion-parameter scale; Appendix~\ref{sec:appendix:scale}) for hybrid Transformer-SSM architectures with near-zero transfer gap in our experimental setting, and propose a two-condition decomposition of the joint action of $\muP$'s update-to-weight invariance and the optimizer's per-component normalization.
Transfer is robust to batch size, architecture pattern, training duration, sequence length, and T:M ratio, each varied individually, and generalizes to Nemotron-H, an independently designed production hybrid.

\paragraph{Limitations.}
Our results characterize a specific operating regime; three scope boundaries are worth flagging.
First, billion scale, the full training horizon, and multiple seeds are validated individually rather than jointly: the LR-transfer sweep reaches 1.24B parameters (three seeds, 20k steps), and one 1.24B model trains cleanly to a Chinchilla-optimal horizon (24.9B tokens) at the transferred LR with a single LR and seed (Appendix~\ref{sec:appendix:scale}); whether the near-zero gap persists when all three hold at once remains open. Larger architectures, including Mixture-of-Experts, remain untested, and Mamba-3~\citep{gu2026mamba3} is left to future work: its complex-valued state lies outside the real-valued $\muP$ rules, and we know of no complex-state $\muP$ formulation; a hybrid built on a real-valued non-Mamba recurrent mixer, Gated Linear Attention (GLA)~\citep{yang2024gla}, does transfer (Appendix~\ref{sec:appendix:gla}).
Second, we additionally ran Sophia~\citep{liu2024sophia}, which also normalizes per element and likewise transfers ($0.00\%$ gap on \HHHH{} and \MHMH{} at 20k steps, three seeds), whereas Muon~\citep{jordan2024muon}, under its own recommended scaling, does not (Appendix~\ref{sec:appendix:sophia_muon}); the two differ in prescription as well as normalizer, so this does not isolate per-element normalization.
Finally, two formal-theory questions remain open: whether the two-condition decomposition is jointly sufficient for transfer at infinite width, and a faithful test of $\muP$-SSM under ZOH SSMs with growing $N_x$, the regime in which \citet{vankadara2024feature}'s prescription was derived (our comparisons apply it to the field-standard simplified-ZOH Mamba~\citep{gu2023mamba} at fixed $N_x{=}16$; Appendix~\ref{sec:appendix:vankadara_faithful}).

\begin{ack}
This work is in part supported by the National Research Foundation of Korea (NRF, RS-2024-00451435(20\%), RS-2024-00413957(20\%)), Institute of Information \& communications Technology Planning \& Evaluation (IITP, RS-2025-02305453(15\%), RS-2025-02273157(15\%), RS-2025-25442149(15\%), RS-2021-II211343(15\%)) grant funded by the Ministry of Science and ICT (MSIT), Institute of New Media and Communications (INMAC), and the BK21 FOUR program of the Education, Artificial Intelligence Graduate School Program (Seoul National University), and Research Program for Future ICT Pioneers, Seoul National University in 2026. The authors declare no competing interests.
\end{ack}

\bibliographystyle{plainnat}
\bibliography{references}

\pagebreak

\appendix
\section*{Experiment Inventory}
\label{sec:appendix:inventory}

\noindent
The appendix collects all per-experiment details. Table~\ref{tab:appendix:inventory} indexes each experiment to its section, reports the design size (number of configurations), and the headline finding. Total compute and per-config wall times are reported in Section~\ref{sec:appendix:compute_budget}.

\begin{table}[H]
\centering
\small
\caption{Experiment inventory across the appendix. ``\#'' is the number of configurations in the design; ``naïve / approx.~/ faithful'' denote \naive $\muP$, approximated block-wise $\muP$, and the literal Vankadara prescription, respectively.}
\label{tab:appendix:inventory}
\begin{tabular}{l p{6.2cm} r p{4.5cm}}
\toprule
§ & Experiment & \# & Key finding \\
\midrule
\ref{sec:appendix:vankadara_faithful} & Faithful $\muP$-SSM: pure-Mamba SGD + hybrid AdamW & 33 & Lower loss in Vankadara regime; $5$--$8\%$ higher loss on hybrids \\
\ref{sec:appendix:max_stable_lr} & Max stable LR across 9 unit interfaces $\times$ widths $\times$ schemes & 195 & Naïve $\muP$ stable to $\eta{=}1$ on all; SP shrinks with width \\
\ref{sec:appendix:depth_scale} & Depth $4$--$32$ across 8 architectures ($w{=}512$) & 128 & Naïve $\muP$ keeps $\eta^*{=}1$ at all depths \\
\ref{sec:appendix:depth_width} & Joint depth$\times$width grid (proxy $\to$ target) & 54 & $\eta{=}1$ transfers $d{=}8\, w{=}256 \to d{=}32\, w{=}1024$ \\
\ref{sec:appendix:fineweb_detail} & FineWeb-Edu width transfer (500M tokens) & 132 & $0.0\%$ gap; SP $3\times$ shift \\
\ref{sec:results:upto_w2048} & Width 2048 ($8\times$ scaling; \MHMH{} 1.24B, \HHHH{} 1.66B params) & 20 & $0.0\%$ gap at $8\times$ \\
\ref{sec:appendix:gla} & GLA--Transformer hybrid (\texttt{GTGT}) width transfer & 9 & $\eta^*{=}1$ at every width \\
\ref{sec:appendix:scale} & 1.24B: three seeds at 20k/60k steps, a 24.9B-token run, depth to 1.88B & 53 & $\eta^*{=}1$ at every seed; converges at 24.9B tokens \\
\ref{sec:results:finegrained_LR} & Fine-grained LR ($\eta\in\{0.5, 0.7, 1.5, 2.0\}$) & 24 & Curves superimpose; basin $\eta\in[0.5,1.5]$ \\
\ref{sec:appendix:nemotron_blocks} & Nemotron-H: MHA vs GQA, released-configuration dimensions, two runs at ${\sim}20$ tokens per parameter & 77 & $\eta^*$ width-independent in each variant; no divergence \\
\ref{sec:appendix:convergence} & Convergence at 20k steps (\textsc{WikiText-103}) & 156 & $0.0\%$ gap holds; SP divergence rate $100\%$ \\
\ref{sec:appendix:multiseed} & Multi-seed validation (3 seeds) & 36 & $0.0\%$ at every seed; CV $\leq 1.7\%$ \\
\ref{sec:appendix:batchsize} & Batch size $\{8, 16, 32, 64\}$ & 112 & Gap $\leq\!1.1\%$ across batch sizes \\
\ref{sec:appendix:seqlen} & Sequence length $\{128, 256, 512, 1024, 2048\}$ & 90 & $\eta^*{=}1$ at every seq length \\
\ref{sec:appendix:tm_ratio} & Parallel T:M ratio (in-block imbalance) & 252 & $0.0\%$ gap, all 6 ratios \\
\ref{sec:appendix:jamba_imbalanced} & Serial T:M sweep (incl.\ Jamba layout) & 504 & Naïve $\muP$ $0.0\%$; SP and faithful drift \\
\ref{sec:appendix:grad_clip} & Gradient clipping $\{$off$, 0.5, 1.0, 5.0\}$ & 24 & Optimal LR unchanged \\
\ref{sec:appendix:coord_check} & Coordinate check (4 schemes $\times$ 5 widths $\times$ 5 seeds) & 100 & All four parameterizations fail coord check on hybrids; transfer not predicted by it \\
\ref{sec:appendix:gradient} & Gradient + step-1 effective update test & 108 & $\gamma$ width-dep; faithful $\alpha(N_u){=}{+}1.00$ for $A_{\log}$ \\
\ref{sec:appendix:optimizer} & Optimizer ablations (SGD vs AdamW, momentum, WD, $\epsilon$; Sophia, Muon) & 415 & Only $g/\sqrt{v}$ is load-bearing \\
\bottomrule
\end{tabular}
\end{table}

\section{Setup and Parameterization Details}
\label{sec:appendix:parameterization}

\subsection{Compute Budget and Tuning-Cost Savings}
\label{sec:appendix:compute_budget}

Total project compute: \nGpuHrs\ GPU-hours across \nConfigs\ completed configurations on NVIDIA A6000 (48\,GB), A5000 (24\,GB), and A100 (80\,GB) on an internal SLURM cluster, counting every job of the project from its Slurm accounting, including preliminary, failed, and rerun jobs; runs on standalone workstations are not counted.

\paragraph{Per-config wall time.}
We estimate single-config wall time on a single A6000 from Slurm accounting (HHHH at $d{=}16$, batch $= 64$, $\text{seq\_len} = 128$); other architectures and GPUs are within $\pm 20\%$ of these values.
Table~\ref{tab:per_config_time} reports the per-config means we use to convert sweep budgets to GPU-hours.

\begin{table}[H]
\centering
\small
\caption{Estimated mean wall time per config on a single A6000 (HHHH, $d{=}16$). Times rise super-linearly because the SSM scan and attention costs are non-quadratic in width; the proxy-to-target ratio is $5.5\times$ at $2$k steps and ${\sim}12\times$ at $20$k.}
\label{tab:per_config_time}
\begin{tabular}{lcc}
\toprule
Width & Params (HHHH) & 2k steps / 20k steps \\
\midrule
$w{=}256$  & 71M  & 3.1 min / ${\sim}16$ min  \\
$w{=}512$  & 181M & 7.1 min / ${\sim}55$ min  \\
$w{=}1024$ & 518M & 17.1 min / ${\sim}191$ min \\
\bottomrule
\end{tabular}
\end{table}

\paragraph{Per-architecture LR tuning cost.}
The headline practical implication of \naive $\muP$ + AdamW is that practitioners can replace a per-width LR sweep at the target with the same sweep at a small proxy plus one target-width verification run.
Using the wall times above, on a target width of $w{=}1024$:
\begin{itemize}
    \setlength{\itemsep}{0pt}
    \item \emph{SP baseline} (no transfer, sweep $7$ LRs at $w{=}1024$): $7 \times 17.1 \text{ min} \approx \mathbf{2.0}$ GPU-hours per architecture at $2$k steps; $7 \times 191 \text{ min} \approx \mathbf{22.3}$ GPU-hours at $20$k steps.
    \item \emph{\Naive $\muP$ transfer} (the same $7$-LR sweep at proxy $w{=}256$, $1$ verification run at $w{=}1024$): $7 \times 3.1 + 1 \times 17.1 \approx 38.8 \text{ min} \approx \mathbf{0.65}$ GPU-hours at $2$k; $7 \times 16 + 1 \times 191 \approx 303 \text{ min} \approx \mathbf{5.1}$ GPU-hours at $20$k.
\end{itemize}
Per-architecture saving: $\mathbf{68\%}$ reduction at the $2$k-step transfer horizon ($\mathbf{3.1\times}$ speedup); $\mathbf{77\%}$ at convergence ($\mathbf{4.4\times}$).
The savings grow with target width because the proxy cost stays constant while the target cost scales superlinearly: a width-$2048$ extrapolation using the same recipe pushes the saving to ${\sim}82\%$ ($5.7\times$).

\subsection{Dataset Sizes and Training Coverage}
\label{sec:appendix:dataset_coverage}

We use three text corpora across the experiments. Table~\ref{tab:dataset_coverage} reports the total training-split token counts, the tokenizer used, and how many optimizer steps make one epoch under our canonical configuration ($\text{seq\_len}{=}128$, $\text{batch\_size}{=}64$, so $128 \times 64 = 8{,}192$ tokens per step).

\begin{table}[H]
\centering
\small
\caption{Dataset sizes and training coverage at the canonical $\text{seq\_len}{=}128$, $\text{batch\_size}{=}64$ setup ($8{,}192$ tokens/step). The right two columns express the step budget as a fraction (or multiple) of one full epoch.}
\label{tab:dataset_coverage}
\resizebox{\columnwidth}{!}{%
\begin{tabular}{lrlrcc}
\toprule
Dataset & Train tokens & Tokenizer (vocab) & Steps/epoch & 2k steps & 20k steps \\
\midrule
\textsc{WikiText-2}    & $2.39$M  & GPT-2 BPE ($50{,}257$)              & $292$       & $6.9$ epochs  & $68$ epochs   \\
\textsc{WikiText-103}  & ${\sim}120$M & tiktoken cl100k\_base ($100{,}277$) & $14{,}648$ & $14\%$ epoch  & $1.37$ epochs \\
\textsc{FineWeb-Edu} 500M & $500$M & tiktoken cl100k\_base ($100{,}277$) & $61{,}035$ & $3.3\%$ epoch & $33\%$ epoch  \\
\bottomrule
\end{tabular}
}
\end{table}

\paragraph{Implications for the headline claims.}
The width-transfer grid (2k steps on \textsc{WikiText-2}; main-text Table~\ref{tab:width_transfer}) covers about $7$ passes through that small corpus, so the $0.0\%$ transfer gap there is measured under data-replay rather than a fresh-data regime.
The convergence-validation runs (20k steps on \textsc{WikiText-103}; Section~\ref{sec:appendix:vankadara_faithful}, Section~\ref{sec:appendix:convergence}) cover one full epoch with a small overflow.
The \textsc{FineWeb-Edu} sweeps (Section~\ref{sec:appendix:fineweb_detail}) at 20k steps consume ${\sim}1/3$ of the 500M-token corpus, well within a single epoch.
For the 5B-token run (\MHMH-415M on \textsc{FineWeb-Edu}, $5$B tokens, $38$k steps), the canonical setup uses a larger effective batch; under the same $8{,}192$-token-per-step accounting, $5$B tokens would correspond to ${\sim}610$k steps.

\subsection{Approximated Block-wise $\muP$: Formula}
\label{sec:appendix:a_scaling}

Block-wise $\muP$~\citep{vankadara2024feature} prescribes $\eta_A \in \Theta(N_u)$, $\eta_B \in \Theta(1/\sqrt{N_u})$, and $\eta_C \in \Theta(1/(N_x\sqrt{N_u}))$ for the SSM components ($A_{\log}, B, C$).
These asymptotic scalings must be anchored at a reference width where $\eta_{\text{base}}$ is tuned; an anchored implementation at $N_u^{\text{ref}} = 512$ is reported in Appendix~\ref{sec:appendix:vankadara_faithful}.
Under AdamW, $g/\sqrt{v}$ already supplies the per-component normalization that the $\Theta(N_u)$ amplification of $\eta_A$ would provide under SGD (Section~\ref{sec:analysis:two_conditions}), so applying both produces double normalization that destabilizes training as $N_u$ grows.

The approximated block-wise $\muP$ used in our experiments keeps the per-component initialization rules ($\sigma_B \in \Theta(\sqrt{N_x/N_u})$, $\sigma_C \in \Theta(1/\sqrt{N_x N_u})$) and the $\eta_C$ scaling of \citet{vankadara2024feature}, and scales $\eta_A$ and $\eta_B$ with the \naive-$\muP$ exponent:
\begin{align}
\eta_{A_{\log}} = \eta_{\text{base}} / N_u, \qquad
\eta_B = \eta_{\text{base}} \sqrt{N_x} / N_u, \qquad
\eta_C = \eta_{\text{base}} / (N_x \sqrt{N_u}),
\end{align}
where $N_u = d_{\text{inner}} = d_{\text{model}} \cdot \texttt{ssm\_expand}$ is the SSM inner dimension.
This drops the $\Theta(N_u)$ amplification of $\eta_A$ (the term redundant with AdamW's $g/\sqrt{v}$) and replaces $\eta_B$'s $\Theta(1/\sqrt{N_u})$ with $\Theta(1/N_u)$, while preserving the per-component structure that distinguishes block-wise $\muP$ from \naive $\muP$ (different scalings per SSM component).

The empirical comparison between the approximated and full block-wise $\muP$, both under AdamW on hybrids, is reported in Appendix~\ref{sec:appendix:vankadara_faithful}: the approximated variant stays within $0.5\%$ of \naive $\muP$ at $w{=}1024$ after 20k steps on \textsc{WikiText-103}, while the full prescription produces $5.6$--$7.7\%$ higher loss on every (architecture, width) cell.

\subsection{Applying $\muP$-SSM Beyond the Pure-SSM SGD Setting}
\label{sec:appendix:vankadara_faithful}

\citet{vankadara2024feature} derived their $\muP$-SSM scaling rule (Table~1) for pure SSMs under plain SGD in the limit $N_u, N_x \!\to\! \infty$ (with $N_u = 16 N_x$ in their reported experiments).
Our work uses a different setting along three axes simultaneously: hybrid Transformer-SSM architectures, AdamW, and a fixed $N_x = 16$ as $N_u$ grows with width.
This appendix records how the Table~1 rule behaves across the gap, in two settings: (i) on pure Mamba under SGD but with our fixed-$N_x$ setup, and (ii) on our headline hybrid grid under AdamW.

We use a properly anchored implementation of Table~1 (anchored at a reference width $N_u^{\text{ref}}$; we use $N_u^{\text{ref}} = 512$, corresponding to our proxy width $w{=}256$):
\begin{align}
\eta_A(N_u) &= \eta_{\text{base}} \cdot \frac{N_u}{N_u^{\text{ref}}} \quad\in\Theta(N_u)\\
\eta_B(N_u) &= \eta_{\text{base}} \cdot \sqrt{\frac{N_u^{\text{ref}}}{N_u}} \quad\in\Theta(1/\sqrt{N_u})\\
\eta_C(N_u) &= \eta_{\text{base}} \cdot \frac{1}{N_x} \sqrt{\frac{N_u^{\text{ref}}}{N_u}} \quad\in\Theta(1/(N_x\sqrt{N_u}))
\end{align}
By construction, all three LRs equal $\eta_{\text{base}}$ (modulo the $1/N_x$ factor for $\eta_C$) at the reference width $N_u^{\text{ref}}$.
At target widths, $\eta_A$ grows linearly with $N_u$ while $\eta_B$ and $\eta_C$ shrink as $1/\sqrt{N_u}$.
At $w{=}1024$ ($N_u{=}2048$) with $\eta_{\text{base}}{=}1.0$: $\eta_A{=}4.0$, $\eta_B{=}0.5$, $\eta_C{=}0.031$.

\paragraph{Setting (i): pure Mamba under SGD with fixed $N_x$.}
We run the Table~1 rule on pure Mamba under plain SGD (no gradient clipping, 20k steps, WikiText-103) across widths $\{128, 1024\}$ and LRs $\{0.3, 1.0, 3.0, 10.0, 30.0, 100.0\}$.
Table~\ref{tab:vankadara_faithful_sanity} reports the results.

\begin{table}[h]
\centering
\small
\caption{Faithful block-wise $\muP$ vs.\ \naive $\muP$ under plain SGD on pure Mamba (20k steps, WikiText-103). Final training loss.}
\label{tab:vankadara_faithful_sanity}
\begin{tabular}{lrrrr}
\toprule
LR & \naive $w{=}128$ & block-wise $w{=}128$ & \naive $w{=}1024$ & block-wise $w{=}1024$ \\
\midrule
0.3   & 10.39 & 8.23  & 8.83 & 8.34 \\
1.0   &  9.52 & 7.01  & 7.92 & 7.43 \\
3.0   &  7.88 & 6.64  & 7.38 & 6.79 \\
10.0  &  6.58 & 6.30  & 6.88 & 6.34 \\
30.0  &  5.96 & 6.05  & 6.39 & 5.95 \\
100.0 &  5.55 & \textbf{DIV} (step 942) & 6.00 & 5.96 \\
\bottomrule
\end{tabular}
\end{table}

The Table~1 rule produces lower loss than \naive $\muP$ at low-to-mid LRs (0.3--3.0) at both widths, consistent with the rule doing the per-component balancing it was designed for.
At the highest LR ($\eta{=}100$), the rule diverges at $w{=}128$ (loss $\to$ $5706$ at step 942) while \naive $\muP$ completes all 20k steps; at $w{=}1024$ both converge to similar loss.
This differs from the Figure~3 pattern of \citet{vankadara2024feature}, which reports a wider stability window for $\muP$-SSM than for the $\muP$ heuristic.

The proximate cause is the $N_u/N_x$ ratio.
\citet{vankadara2024feature}'s Figure~3 setting holds $N_u = 16 N_x$ with both axes scaling together; in our MambaBlock, $N_x = d_{\text{state}} = 16$ is fixed across widths while $N_u = d_{\text{inner}} = 2 d_{\text{model}}$ grows with width.
So $N_u/N_x = 16$ only at $w{=}128$; at $w{=}1024$ we operate at $N_u/N_x = 128$.
The asymptotic scalings $\Theta(N_u)$ and $\Theta(1/\sqrt{N_u})$ behave differently when $N_x$ is held fixed versus scaled with $N_u$, and the growing-$N_x$ limit the theorems use does not match our fixed-$N_x$ setup.
Testing the rule with $N_x$ growing proportionally (i.e., $N_u = 16 N_x$ at every width) closes this regime gap empirically: in a follow-up sweep at $N_u = 16 N_x$ across $w \in \{128, 256, 512, 1024\}$ and the same SGD setup, the Table~1 rule produces lower loss than \naive $\muP$ at low-to-mid LRs ($\eta \in [0.3, 30]$) at every width, with naïve catching up only at $\eta{=}100$, partially recovering Vankadara's reported loss-ordering claim in the setting of their experiments.

\paragraph{Setting (ii): hybrid Transformer-SSM under AdamW.}
This is our actual experimental regime.
The Table~1 rule was derived without modeling AdamW's per-parameter normalization or the block heterogeneity from non-SSM components, and we observe that its behavior here differs from what Setting~(i) suggests.
Because the rule is only one possible point of comparison and the headline of the paper is a finding about \naive $\muP$, we summarize the results here rather than carry full per-cell tables in the main text.

\emph{Transfer at 2k steps (\textsc{WikiText-2}, Table~\ref{tab:width_transfer} grid).}
The Table~1 rule's optimum LR drifts with width on every SSM-bearing architecture we measured (7 of 7: \MM, \MTMT, \MHMH, \HMHM, \HHHH, \HTHT, plus \texttt{JAMBA}): the $w{=}1024,\ \eta{=}3.0$ run diverges on all seven, collapsing the optimum to $\eta^*{=}1.0$, while at smaller widths $\eta^*{=}3.0$ is preferred on 5 of 7 archs (\MM, \MHMH, \HMHM, \HHHH, \texttt{JAMBA}; on \MTMT and \HTHT a single proxy width prefers $\eta^*{=}1.0$).
Only \TT (pure Transformer, where the SSM-specific rules do not fire) preserves $\eta^*{=}1.0$ across widths.
The drift has the same mechanism as in Setting~(i): $\eta_A \!\propto\! N_u$ becomes large at scale relative to the gradient magnitudes our hybrid blocks produce.

\emph{Convergence at 20k steps (\textsc{WikiText-103}).}
At the safely-transferable LR $\eta{=}1.0$, block-wise $\muP$ produces $5.6$--$7.7\%$ higher loss than \naive $\muP$ across HHHH, MHMH, MTMT $\times$ $w \in \{256, 512, 1024\}$ ($n=9$ cells, no exceptions; Table~\ref{tab:vankadara_faithful_hybrid_adamw}).
The approximated variant, used in the main experiments, stays within $0.5\%$ of \naive at $w{=}1024$ (\MHMH, 20k steps, one seed; $0.44\%$ in training and $0.34\%$ in validation loss); under AdamW the explicit per-component LR rule is therefore redundant with $g/\sqrt{v}$, and the redundancy slows training (Section~\ref{sec:analysis:redundant_normalization}).

\begin{table}[h]
\centering
\small
\caption{Faithful block-wise $\muP$ vs.\ \naive $\muP$ on hybrids under AdamW (20k steps, \textsc{WikiText-103}, $\eta_{\text{base}}{=}1.0$, warmup $200$ steps). Final training loss; relative gap is $(\mathcal{L}_{\text{block-wise}} - \mathcal{L}_{\text{\naive}}) / \mathcal{L}_{\text{\naive}}$.}
\label{tab:vankadara_faithful_hybrid_adamw}
\begin{tabular}{l c r r r}
\toprule
Arch & Width & \Naive & Block-wise & Gap \\
\midrule
\HHHH & 256  & 3.514 & 3.712 & $+5.6\%$ \\
\HHHH & 512  & 3.270 & 3.500 & $+7.0\%$ \\
\HHHH & 1024 & 3.191 & 3.414 & $+7.0\%$ \\
\addlinespace
\MHMH & 256  & 3.741 & 3.957 & $+5.8\%$ \\
\MHMH & 512  & 3.317 & 3.520 & $+6.1\%$ \\
\MHMH & 1024 & 3.171 & 3.395 & $+7.1\%$ \\
\addlinespace
\MTMT & 256  & 3.607 & 3.835 & $+6.3\%$ \\
\MTMT & 512  & 3.495 & 3.762 & $+7.7\%$ \\
\MTMT & 1024 & 3.158 & 3.403 & $+7.7\%$ \\
\bottomrule
\end{tabular}
\end{table}

\paragraph{Where this leaves $\muP$-SSM and our work.}
Same-group follow-ups extend the $\muP$ framework along axes other than the SSM rule itself (\citet{haas2024mup2} ($\muP^2$) address sharpness-aware minimization; sparse-architecture extensions appear in \citet{dey2024sparse, malasnicki2025moe}), and a side-by-side study under matched $N_u/N_x$ has not yet appeared in the published literature.
Production hybrids that deploy $\muP$ at scale (Falcon-H1 with 35 per-component multipliers; community Mamba-$\muP$ implementations with zero-initialized convolutions) adopt their own per-component recipes targeting production-training stability, a problem distinct from the LR-transfer question we focus on (Section~\ref{sec:background:gap}).
Under AdamW on hybrids with fixed $N_x$, the block-wise $\muP$ rule fires at the gradient level (Appendix~\ref{sec:appendix:gradient}) but is redundant with Adam's $g/\sqrt{v}$ at training time (Section~\ref{sec:analysis:redundant_normalization}).
Our approximated block-wise variant addresses this redundancy by scaling $\eta_A$ and $\eta_B$ with the \naive-$\muP$ exponent (dropping the $\eta_A \in \Theta(N_u)$ term that AdamW already supplies), preserving the per-component initialization structure while avoiding the double normalization.

\paragraph{Regime gaps from the original derivation.}
We work outside the regime in which \citet{vankadara2024feature}'s rules were derived along four axes: (i) hybrid Transformer-SSM blocks vs the pure SSMs of their experiments; (ii) AdamW vs the SGD step from which their Table~1 is derived; (iii) fixed $N_x{=}16$ vs growing $N_x$ ($N_u/N_x{=}16$ in their reported experiments; proportional scaling of $N_x$ with $N_u$ is impractical at scale because SSM state memory scales as $N_u \!\times\! N_x$, and is therefore not used in any production hybrid we are aware of); and (iv) the discretization $B_d \!\approx\! \Delta t \cdot B$ used in the official Mamba library~\citep{gu2023mamba} vs the true-ZOH form $B_d = (\Delta t \cdot A)^{-1}(\exp(\Delta t \cdot A) - I) \cdot \Delta t \cdot B$ underlying their Table~1 derivation. The simplified-ZOH choice (iv) is the field standard, shared by Mamba-2~\citep{dao2024mamba2}, Jamba~\citep{lieber2024jamba}, Nemotron-H~\citep{nvidia2025nemotron}, and Falcon-H1~\citep{zuo2025falcon}, so it is not a quirk of our setup. We apply Vankadara's initialization formulas and (for the non-approximated variant) their LR exponents anchored at our proxy width $N_u^{\text{ref}}{=}512$; whether the same exponents are theoretically optimal under simplified ZOH is not established. Implementing true-ZOH Mamba and re-evaluating block-wise $\muP$ on its own derived discretization, ideally also under proportional $N_u/N_x$ scaling and at production-scale $N_x$, is left to future work.

\paragraph{Relation to the main text.}
The ``block-wise $\muP$ (approximated)'' results in Section~\ref{sec:results} and Table~\ref{tab:width_transfer} use our approximated variant (Section~\ref{sec:setup}, Appendix~\ref{sec:appendix:a_scaling}) under AdamW on hybrids with a single shared $\eta_{\text{base}}$.
The full block-wise $\muP$ prescription is reported alongside in Table~\ref{tab:width_transfer} (transfer column) and Table~\ref{tab:vankadara_faithful_hybrid_adamw} above (convergence comparison); a gradient-level comparison is in Appendix~\ref{sec:appendix:gradient}.
Our main claim (\naive $\muP$ + AdamW achieves $0.0\%$ transfer gap across all 7 architectures) rests on \naive $\muP$'s own empirical behavior, not on a comparison to block-wise.
In the side-by-side comparisons we do report, \naive achieves lower loss than both block-wise variants under AdamW on hybrids, and the full prescription's optimum additionally drifts with width on every SSM-bearing architecture in the 2k-step transfer grid.
\definecolor{mslr1}{gray}{0.97}
\definecolor{mslr03}{gray}{0.88}
\definecolor{mslr01}{gray}{0.80}
\definecolor{mslr003}{gray}{0.72}
\definecolor{mslr001}{gray}{0.64}
\definecolor{mslr0003}{gray}{0.56}

\section{Unit Interface and Depth Details}
\label{sec:appendix:unit_interface}

This section provides additional details on the unit-interface and depth validation experiments (Section~\ref{sec:results:unit}).

\subsection{Max Stable Learning Rate}
\label{sec:appendix:max_stable_lr}

Table~\ref{tab:max_stable_lr} reports the maximum LR at which training stays bounded for each (interface, scheme, width) cell in the unit-interface sweep.

\begin{table}[H]
\centering
\footnotesize
\caption{Max stable LR at the unit interface across all 9 (T, M, H) ordered pairs (darker = less stable), 1k steps (warmup 50) on \textsc{WikiText-2}.
LR grid: $\{0.001, 0.003, 0.01, 0.03, 0.1, 0.3, 1.0\}$; an entry of $1.0$ indicates stability at the grid maximum (the true max stable LR may be higher).
\Naive $\muP$: stable at $\eta{=}1.0$ across every (interface, width) cell.
SP: max stable LR shrinks with width.
Block-wise $\muP$ (approximated): stable across every cell, but with $\eta_A$ instead set to $\eta_{\text{base}} N_u$ it collapses for serial-SSM interfaces (\texttt{TM}, \texttt{MT}, \texttt{MM}) where the per-component LR correction stacks with Adam's $g/\sqrt{v}$.
Block-wise $\muP$ (faithful, the literal Vankadara prescription): stable across every cell, in contrast to the approximated variant with $\eta_A{=}\eta_{\text{base}} N_u$.}
\label{tab:max_stable_lr}
\begin{tabular}{l l ccccc}
\toprule
 & & \multicolumn{5}{c}{Width ($\dmodel$)} \\
\cmidrule(lr){3-7}
Interface & Scheme & 128 & 256 & 512 & 1024 & 2048 \\
\midrule
\multirow{4}{*}{TT}
 & SP                          & \cellcolor{mslr1}1.0 & \cellcolor{mslr03}0.30 & \cellcolor{mslr03}0.30 & \cellcolor{mslr03}0.30 & \cellcolor{mslr003}0.03 \\
 & \Naive $\muP$               & \cellcolor{mslr1}\textbf{1.0} & \cellcolor{mslr1}\textbf{1.0} & \cellcolor{mslr1}\textbf{1.0} & \cellcolor{mslr1}\textbf{1.0} & \cellcolor{mslr1}\textbf{1.0} \\
 & Block-wise (approx.)        & \cellcolor{mslr1}1.0 & \cellcolor{mslr1}1.0 & \cellcolor{mslr1}1.0 & \cellcolor{mslr1}1.0 & \cellcolor{mslr1}1.0 \\
 & Block-wise (faithful)       & \cellcolor{mslr1}1.0 & \cellcolor{mslr1}1.0 & \cellcolor{mslr1}1.0 & \cellcolor{mslr1}1.0 & \cellcolor{mslr1}1.0 \\
\midrule
\multirow{5}{*}{TM}
 & SP                          & \cellcolor{mslr03}0.30 & \cellcolor{mslr03}0.30 & \cellcolor{mslr03}0.30 & \cellcolor{mslr01}0.10 & \cellcolor{mslr003}0.03 \\
 & \Naive $\muP$               & \cellcolor{mslr1}\textbf{1.0} & \cellcolor{mslr1}\textbf{1.0} & \cellcolor{mslr1}\textbf{1.0} & \cellcolor{mslr1}\textbf{1.0} & \cellcolor{mslr1}\textbf{1.0} \\
 & Block-wise (approx.)        & \cellcolor{mslr1}1.0 & \cellcolor{mslr1}1.0 & \cellcolor{mslr1}1.0 & \cellcolor{mslr1}1.0 & \cellcolor{mslr1}1.0 \\
 & Block-wise (approx.), $\eta_A{=}\eta_{\text{base}} N_u$ & \cellcolor{mslr003}0.03 & \cellcolor{mslr003}0.03 & \cellcolor{mslr001}0.01 & \cellcolor{mslr0003}$\leq$0.003 & \cellcolor{mslr0003}$\leq$0.003 \\
 & Block-wise (faithful)       & \cellcolor{mslr1}1.0 & \cellcolor{mslr1}1.0 & \cellcolor{mslr1}1.0 & \cellcolor{mslr1}1.0 & \cellcolor{mslr1}1.0 \\
\midrule
\multirow{4}{*}{TH}
 & SP                          & \cellcolor{mslr03}0.30 & \cellcolor{mslr03}0.30 & \cellcolor{mslr03}0.30 & \cellcolor{mslr01}0.10 & \cellcolor{mslr003}0.03 \\
 & \Naive $\muP$               & \cellcolor{mslr1}\textbf{1.0} & \cellcolor{mslr1}\textbf{1.0} & \cellcolor{mslr1}\textbf{1.0} & \cellcolor{mslr1}\textbf{1.0} & \cellcolor{mslr1}\textbf{1.0} \\
 & Block-wise (approx.)        & \cellcolor{mslr1}1.0 & \cellcolor{mslr1}1.0 & \cellcolor{mslr1}1.0 & \cellcolor{mslr1}1.0 & \cellcolor{mslr1}1.0 \\
 & Block-wise (faithful)       & \cellcolor{mslr1}1.0 & \cellcolor{mslr1}1.0 & \cellcolor{mslr1}1.0 & \cellcolor{mslr1}1.0 & \cellcolor{mslr1}1.0 \\
\midrule
\multirow{5}{*}{MT}
 & SP                          & \cellcolor{mslr1}1.0 & \cellcolor{mslr03}0.30 & \cellcolor{mslr03}0.30 & \cellcolor{mslr03}0.30 & \cellcolor{mslr01}0.10 \\
 & \Naive $\muP$               & \cellcolor{mslr1}\textbf{1.0} & \cellcolor{mslr1}\textbf{1.0} & \cellcolor{mslr1}\textbf{1.0} & \cellcolor{mslr1}\textbf{1.0} & \cellcolor{mslr1}\textbf{1.0} \\
 & Block-wise (approx.)        & \cellcolor{mslr1}1.0 & \cellcolor{mslr1}1.0 & \cellcolor{mslr1}1.0 & \cellcolor{mslr1}1.0 & \cellcolor{mslr1}1.0 \\
 & Block-wise (approx.), $\eta_A{=}\eta_{\text{base}} N_u$ & \cellcolor{mslr003}0.03 & \cellcolor{mslr001}0.01 & \cellcolor{mslr001}0.01 & \cellcolor{mslr0003}$\leq$0.003 & \cellcolor{mslr001}0.01 \\
 & Block-wise (faithful)       & \cellcolor{mslr1}1.0 & \cellcolor{mslr1}1.0 & \cellcolor{mslr1}1.0 & \cellcolor{mslr1}1.0 & \cellcolor{mslr1}1.0 \\
\midrule
\multirow{5}{*}{MM}
 & SP                          & \cellcolor{mslr1}1.0 & \cellcolor{mslr03}0.30 & \cellcolor{mslr03}0.30 & \cellcolor{mslr03}0.30 & \cellcolor{mslr01}0.10 \\
 & \Naive $\muP$               & \cellcolor{mslr1}\textbf{1.0} & \cellcolor{mslr1}\textbf{1.0} & \cellcolor{mslr1}\textbf{1.0} & \cellcolor{mslr1}\textbf{1.0} & \cellcolor{mslr1}\textbf{1.0} \\
 & Block-wise (approx.)        & \cellcolor{mslr1}1.0 & \cellcolor{mslr1}1.0 & \cellcolor{mslr1}1.0 & \cellcolor{mslr1}1.0 & \cellcolor{mslr1}1.0 \\
 & Block-wise (approx.), $\eta_A{=}\eta_{\text{base}} N_u$ & \cellcolor{mslr003}0.03 & \cellcolor{mslr001}0.01 & \cellcolor{mslr001}0.01 & \cellcolor{mslr0003}$\leq$0.003 & \cellcolor{mslr0003}$\leq$0.003 \\
 & Block-wise (faithful)       & \cellcolor{mslr1}1.0 & \cellcolor{mslr1}1.0 & \cellcolor{mslr1}1.0 & \cellcolor{mslr1}1.0 & \cellcolor{mslr1}1.0 \\
\midrule
\multirow{4}{*}{MH}
 & SP                          & \cellcolor{mslr03}0.30 & \cellcolor{mslr03}0.30 & \cellcolor{mslr03}0.30 & \cellcolor{mslr03}0.30 & \cellcolor{mslr01}0.10 \\
 & \Naive $\muP$               & \cellcolor{mslr1}\textbf{1.0} & \cellcolor{mslr1}\textbf{1.0} & \cellcolor{mslr1}\textbf{1.0} & \cellcolor{mslr1}\textbf{1.0} & \cellcolor{mslr1}\textbf{1.0} \\
 & Block-wise (approx.)        & \cellcolor{mslr1}1.0 & \cellcolor{mslr1}1.0 & \cellcolor{mslr1}1.0 & \cellcolor{mslr1}1.0 & \cellcolor{mslr1}1.0 \\
 & Block-wise (faithful)       & \cellcolor{mslr1}1.0 & \cellcolor{mslr1}1.0 & \cellcolor{mslr1}1.0 & \cellcolor{mslr1}1.0 & \cellcolor{mslr1}1.0 \\
\midrule
\multirow{4}{*}{HT}
 & SP                          & \cellcolor{mslr1}1.0 & \cellcolor{mslr1}1.0 & \cellcolor{mslr03}0.30 & \cellcolor{mslr03}0.30 & \cellcolor{mslr01}0.10 \\
 & \Naive $\muP$               & \cellcolor{mslr1}\textbf{1.0} & \cellcolor{mslr1}\textbf{1.0} & \cellcolor{mslr1}\textbf{1.0} & \cellcolor{mslr1}\textbf{1.0} & \cellcolor{mslr1}\textbf{1.0} \\
 & Block-wise (approx.)        & \cellcolor{mslr1}1.0 & \cellcolor{mslr1}1.0 & \cellcolor{mslr1}1.0 & \cellcolor{mslr1}1.0 & \cellcolor{mslr1}1.0 \\
 & Block-wise (faithful)       & \cellcolor{mslr1}1.0 & \cellcolor{mslr1}1.0 & \cellcolor{mslr1}1.0 & \cellcolor{mslr1}1.0 & \cellcolor{mslr1}1.0 \\
\midrule
\multirow{4}{*}{HM}
 & SP                          & \cellcolor{mslr1}1.0 & \cellcolor{mslr03}0.30 & \cellcolor{mslr03}0.30 & \cellcolor{mslr03}0.30 & \cellcolor{mslr003}0.03 \\
 & \Naive $\muP$               & \cellcolor{mslr1}\textbf{1.0} & \cellcolor{mslr1}\textbf{1.0} & \cellcolor{mslr1}\textbf{1.0} & \cellcolor{mslr1}\textbf{1.0} & \cellcolor{mslr1}\textbf{1.0} \\
 & Block-wise (approx.)        & \cellcolor{mslr1}1.0 & \cellcolor{mslr1}1.0 & \cellcolor{mslr1}1.0 & \cellcolor{mslr1}1.0 & \cellcolor{mslr1}1.0 \\
 & Block-wise (faithful)       & \cellcolor{mslr1}1.0 & \cellcolor{mslr1}1.0 & \cellcolor{mslr1}1.0 & \cellcolor{mslr1}1.0 & \cellcolor{mslr1}1.0 \\
\midrule
\multirow{4}{*}{HH}
 & SP                          & \cellcolor{mslr1}1.0 & \cellcolor{mslr03}0.30 & \cellcolor{mslr03}0.30 & \cellcolor{mslr03}0.30 & \cellcolor{mslr01}0.10 \\
 & \Naive $\muP$               & \cellcolor{mslr1}\textbf{1.0} & \cellcolor{mslr1}\textbf{1.0} & \cellcolor{mslr1}\textbf{1.0} & \cellcolor{mslr1}\textbf{1.0} & \cellcolor{mslr1}\textbf{1.0} \\
 & Block-wise (approx.)        & \cellcolor{mslr1}1.0 & \cellcolor{mslr1}1.0 & \cellcolor{mslr1}1.0 & \cellcolor{mslr1}1.0 & \cellcolor{mslr1}1.0 \\
 & Block-wise (faithful)       & \cellcolor{mslr1}1.0 & \cellcolor{mslr1}1.0 & \cellcolor{mslr1}1.0 & \cellcolor{mslr1}1.0 & \cellcolor{mslr1}1.0 \\
\bottomrule
\end{tabular}
\end{table}

\subsection{Depth Scaling}
\label{sec:appendix:depth_scale}

\begin{figure}[H]
  \centering
  \includegraphics[width=\textwidth]{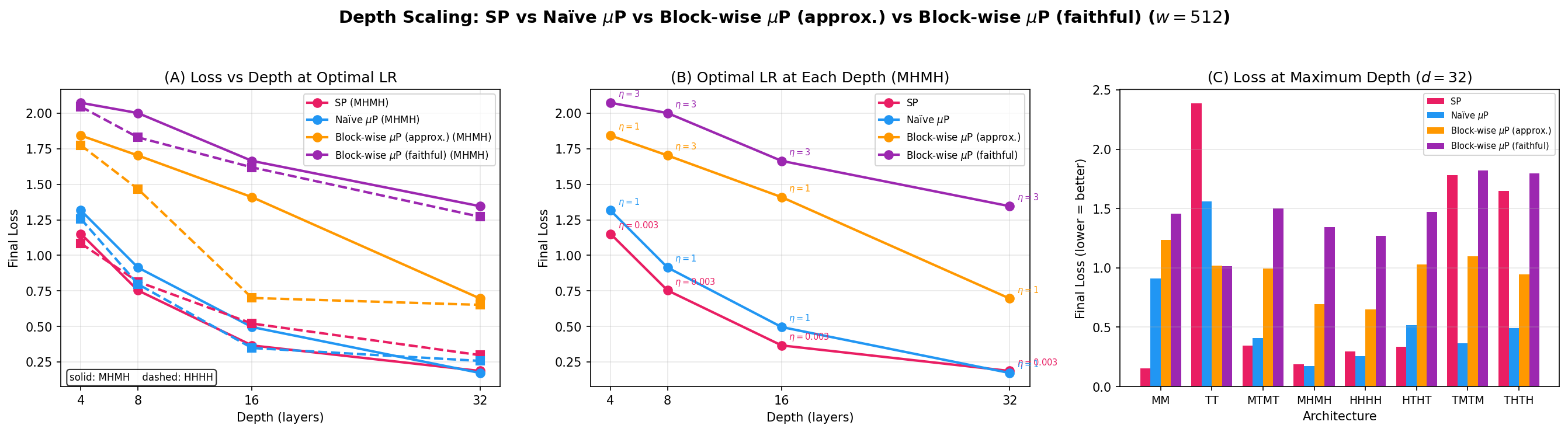}
  \caption{Depth scaling results comparing all four schemes (SP, \naive $\muP$, block-wise $\muP$ approximated, block-wise $\muP$ faithful) at $w{=}512$, 2k steps on \textsc{WikiText-2}.
  (a) Loss vs.\ depth at each scheme's optimal LR for \MHMH and \HHHH.
  (b) Loss vs.\ depth for \MHMH with the optimal LR labeled at each point.
  (c) Loss at maximum depth ($d{=}32$) for all 8 architectures.
  All schemes show decreasing loss with depth; the two block-wise variants have higher loss at the 2k-step horizon on \textsc{WikiText-2}; the approximated variant's 20k-step loss on \textsc{WikiText-103} stays within $0.5\%$ of \naive $\muP$ at $w{=}1024$ (Appendix~\ref{sec:appendix:vankadara_faithful}).
  The faithful variant retains a $5$--$8\%$ loss gap relative to \naive $\muP$ even at convergence (Table~\ref{tab:vankadara_faithful_hybrid_adamw}).}
  \label{fig:phase2_depth_scaling}
\end{figure}

Figure~\ref{fig:phase2_depth_scaling} shows loss as a function of depth. For all four schemes, loss decreases with depth on both \MHMH and \HHHH (panels a and b).

\section{Width Transfer Details}
\label{sec:appendix:width_transfer}

Each full LR-sweep table in this section shows final training loss unless its caption states validation loss (lower is better).
Bold values indicate the best LR at each (architecture, width) combination.
\textit{div} indicates divergence (loss $>$ 20).
All experiments use depth$=16$ unless otherwise noted.

\subsection{Joint Depth-Width Transfer}
\label{sec:appendix:depth_width}

\begin{figure}[H]
  \centering
  \includegraphics[width=0.8\textwidth]{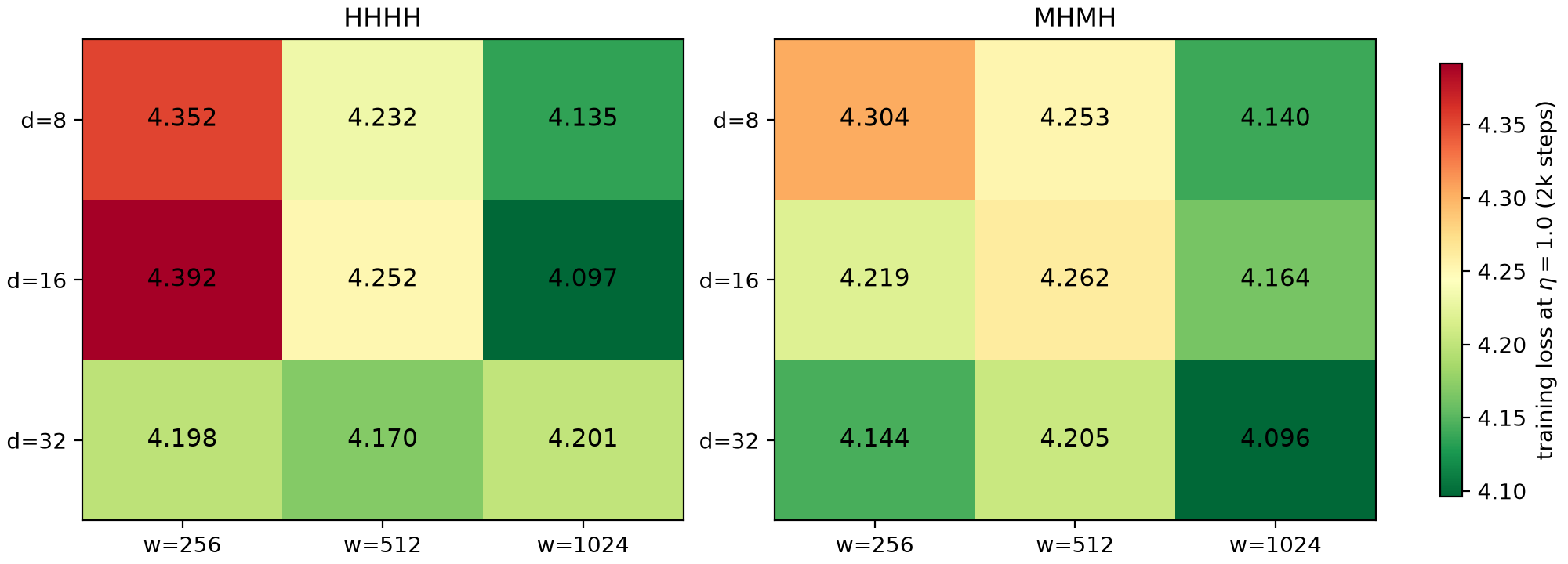}
  \caption{Loss at $\eta{=}1.0$ across the width$\times$depth grid (training loss, 2k steps; values from Table~\ref{tab:full_depthwidth}).
  Best: \MHMH $(w{=}1024, d{=}32)$ at 4.096.}
  \label{fig:depth_width}
\end{figure}

$\eta{=}1.0$ is optimal at the full target ($w{=}1024$, $d{=}32$) for both \HHHH and \MHMH, with $0.0\%$ gap from the proxy model ($w{=}256$, $d{=}8$, ${\sim}11$--$14\times$ smaller in parameters).
At the proxy horizon, loss at $\eta{=}1.0$ is not monotonic in depth or width across the grid (Figure~\ref{fig:depth_width}).
The full LR sweep across the depth$\times$width grid is in Table~\ref{tab:full_depthwidth}.

\begin{table}[H]
\centering\small
\caption{Depth-width joint transfer for \MHMH and \HHHH (\textsc{WikiText-103}, 2k steps, \naive $\muP$). Gap = relative loss penalty of using the proxy-transferred LR ($\eta{=}1$) instead of each row's best-of-grid value, $(\mathcal{L}(\eta{=}1) - \mathcal{L}^*)/\mathcal{L}^*$. The 3-point grid does not sample the basin between $\eta{=}1$ and $\eta{=}3$ (Tab~\ref{tab:full_finegrained} shows the true basin lies near $\eta\!\approx\!1.5$--$2$); cells where $\eta{=}3$ wins by less than 1\% should be read as basin-edge sampling artifacts, not as preferring high LR.}
\label{tab:full_depthwidth}
\begin{tabular}{llcccr}
\toprule
Arch & Depth $\times$ Width & $\eta$=0.3 & $\eta$=1.0 & $\eta$=3.0 & Gap @ $\eta{=}1$ \\
\midrule
\multirow{9}{*}{MHMH} & $d$=8,  $w$=256  & 4.508 & \textbf{4.304} & 4.318          & 0.00\% \\
                     & $d$=8,  $w$=512  & 4.488 & 4.253          & \textbf{4.230} & 0.55\% \\
                     & $d$=8,  $w$=1024 & 4.403 & \textbf{4.140} & 4.204          & 0.00\% \\
                     & $d$=16, $w$=256  & 4.411 & \textbf{4.219} & 4.241          & 0.00\% \\
                     & $d$=16, $w$=512  & 4.495 & \textbf{4.262} & 4.843          & 0.00\% \\
                     & $d$=16, $w$=1024 & 4.453 & \textbf{4.164} & 4.246          & 0.00\% \\
                     & $d$=32, $w$=256  & 4.329 & \textbf{4.144} & 4.819          & 0.00\% \\
                     & $d$=32, $w$=512  & 4.437 & \textbf{4.205} & 4.305          & 0.00\% \\
                     & $d$=32, $w$=1024 & 4.388 & \textbf{4.096} & 4.206          & 0.00\% \\
\midrule
\multirow{9}{*}{HHHH} & $d$=8,  $w$=256  & 4.527 & \textbf{4.352} & 4.363          & 0.00\% \\
                     & $d$=8,  $w$=512  & 4.458 & \textbf{4.232} & 4.258          & 0.00\% \\
                     & $d$=8,  $w$=1024 & 4.413 & \textbf{4.135} & 4.156          & 0.00\% \\
                     & $d$=16, $w$=256  & 4.562 & \textbf{4.392} & 4.465          & 0.00\% \\
                     & $d$=16, $w$=512  & 4.504 & \textbf{4.252} & 4.359          & 0.00\% \\
                     & $d$=16, $w$=1024 & 4.384 & \textbf{4.097} & 4.161          & 0.00\% \\
                     & $d$=32, $w$=256  & 4.366 & \textbf{4.198} & 4.293          & 0.00\% \\
                     & $d$=32, $w$=512  & 4.364 & \textbf{4.170} & 4.226          & 0.00\% \\
                     & $d$=32, $w$=1024 & 4.476 & \textbf{4.201} & 4.288          & 0.00\% \\
\bottomrule
\end{tabular}
\end{table}

\subsection{FineWeb-Edu Width Transfer}
\label{sec:appendix:fineweb_detail}

\begin{figure}[H]
  \centering
  \includegraphics[width=1.0\textwidth]{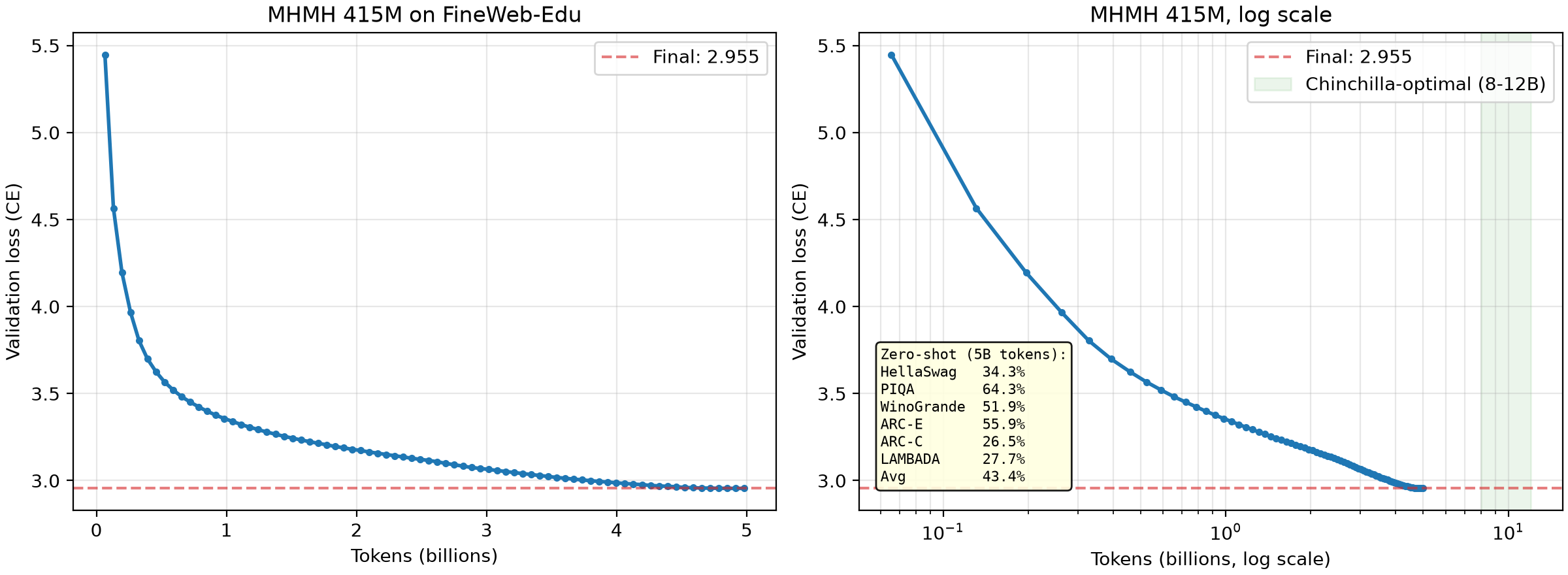}
  \caption{Validation loss curve for \MHMH 415M on \textsc{FineWeb-Edu} (5.0B tokens, 38k steps).
  Smooth convergence from 5.45 (step 500) to 2.96; $\eta{=}1.0$ transferred from proxy ($w{=}256$, 65M params) without retuning.}
  \label{fig:fineweb}
\end{figure}

We additionally validate width transfer on \textsc{FineWeb-Edu} (164M training tokens from a 500M-token subset, 20k steps) for \HHHH, \MHMH, and \MTMT under both SP and \naive $\muP$ (Table~\ref{tab:fineweb}); the corresponding validation-loss trajectory at the 5B-token run of Section~\ref{sec:results:width} is shown in Figure~\ref{fig:fineweb}.

\begin{table}[H]
\centering
\small
\caption{FineWeb-Edu width transfer (20k steps, 164M training tokens from a 500M-token subset).
SP's optimal LR shifts $3\times$ ($0.003 \!\to\! 0.001$); transferring the proxy optimal LR incurs $1.1$--$8.7\%$ penalty (largest on \MHMH and \MTMT, smallest on \HHHH).}
\label{tab:fineweb}
\begin{tabular}{llccccc}
\toprule
 & & $w{=}256$ & \multicolumn{3}{c}{$w{=}1024$} \\
\cmidrule(lr){3-3} \cmidrule(lr){4-6}
Arch & Scheme & Loss (opt.\ LR) & Opt.\ LR & Transferred & Retuned \\
\midrule
HHHH & \Naive $\muP$ & 4.467 ($\eta{=}1.0$) & 1.0 & \textbf{4.115} & --- \\
HHHH & SP & 4.452 ($\eta{=}0.003$) & 0.001 & 4.194 & 4.149 \\
\midrule
MHMH & \Naive $\muP$ & 4.313 ($\eta{=}1.0$) & 1.0 & \textbf{3.993} & --- \\
MHMH & SP & 4.315 ($\eta{=}0.003$) & 0.001 & 4.339 & 3.998 \\
\midrule
MTMT & \Naive $\muP$ & 4.451 ($\eta{=}1.0$) & 1.0 & \textbf{4.033} & --- \\
MTMT & SP & 4.460 ($\eta{=}0.003$) & 0.001 & 4.403 & 4.052 \\
\bottomrule
\end{tabular}
\end{table}

\subsection{Width 2048 and Billion Scale}
\label{sec:results:upto_w2048}
\label{sec:appendix:scale}

Extending the transfer range from $4\times$ ($256{\to}1024$) to $8\times$ ($256{\to}2048$), we train \MHMH (1.24B params) and \HHHH (1.66B params) at $w{=}2048$.
\Naive $\muP$'s optimal LR at the 2k-step horizon is $\eta{=}1.5$ for both architectures, consistent with smaller widths (Table~\ref{tab:full_finegrained}); at convergence (20k, \textsc{WikiText-103}) the optimum shifts back to $\eta{=}1.0$ (\MHMH, Table~\ref{tab:w2048_multiseed}).
SP at $w{=}2048$ only survives at $\eta \leq 0.003$ ($\eta{=}0.01$ near-diverges to loss $>10$); \naive $\muP$ beats SP by $0.7\%$ (\MHMH) and $2.1\%$ (\HHHH) at their respective optimal LRs.
Width 2048 for \MHMH and \HHHH (\textsc{WikiText-103}, 2k steps). \Naive $\muP$ sweeps $\eta \in [0.3, 3.0]$ (Table~\ref{tab:full_w2048_naive}); SP sweeps $\eta \in \{0.001, 0.003, 0.01\}$ (Table~\ref{tab:full_w2048_sp}). SP's basin is narrower and shifts to $\eta \approx 0.001$ at this scale, with $\eta{=}0.01$ near-diverging.

\begin{table}[H]
\centering\small
\caption{Width 2048, \naive $\muP$ sweep.}
\label{tab:full_w2048_naive}
\begin{tabular}{lccccccc r}
\toprule
Arch & $\eta{=}0.3$ & $\eta{=}0.5$ & $\eta{=}0.7$ & $\eta{=}1.0$ & $\eta{=}1.5$ & $\eta{=}2.0$ & $\eta{=}3.0$ & Gap @ $\eta{=}1$ \\
\midrule
MHMH & 4.234 & 4.070 & 3.989 & 3.934 & \textbf{3.907} & 3.984 & 4.044 & 0.70\% \\
HHHH & 4.480 & 4.296 & 4.209 & 4.161 & \textbf{4.121} & 4.173 & 4.270 & 0.97\% \\
\bottomrule
\end{tabular}
\end{table}

\begin{table}[H]
\centering\small
\caption{Width 2048, SP sweep.}
\label{tab:full_w2048_sp}
\begin{tabular}{lccc}
\toprule
Arch & $\eta{=}0.001$ & $\eta{=}0.003$ & $\eta{=}0.01$ \\
\midrule
MHMH & \textbf{3.935} & 4.102 & \textit{11.3} \\
HHHH & \textbf{4.211} & 4.341 & \textit{10.8} \\
\bottomrule
\end{tabular}
\end{table}

\paragraph{Billion scale, horizon, and seeds.}
The 2k sweep above uses one seed. The runs below extend \MHMH{} along three axes, each separately: three seeds at $w{=}2048$ (1.24B parameters), the full training horizon at the same size, and depth up to 1.88B parameters. Billion scale, the full training horizon, and multiple seeds are thus validated individually rather than jointly, and no single run combines them with a full LR sweep.
All three report validation loss.

\emph{Three seeds at 20k and 60k steps.}
Table~\ref{tab:w2048_multiseed} repeats the sweep with seeds 42, 123, and 456 at 20k steps (164M tokens) and at 60k steps (492M tokens) on \textsc{WikiText-103}.
At both horizons $\eta{=}1.0$ is the argmin at every seed, and no run diverges (0/24).
At 20k this is also the $w{=}256$ proxy's optimum (Table~\ref{tab:convergence}), so the transfer gap is $0.0\%$; across seeds the loss at $\eta{=}1.0$ varies by a coefficient of variation of about $0.1\%$, and $\eta{=}1.5$ is $0.289$ nats worse on the seed mean.
At 60k, we observe that the training loss favors $\eta{\leq}0.7$ while the validation loss still selects $\eta{=}1.0$; we conjecture that this reflects these runs passing over \textsc{WikiText-103} about four times, and read the optimum from validation loss.

\begin{table}[H]
\centering\small
\caption{\MHMH{} at $w{=}2048$ (1.24B parameters) on \textsc{WikiText-103}, three seeds: validation loss at 20k and 60k steps. Bold marks the argmin of each column; -- marks an LR not run at that horizon.}
\label{tab:w2048_multiseed}
\begin{tabular}{lcccccccc}
\toprule
 & \multicolumn{4}{c}{20k steps (164M tokens)} & \multicolumn{4}{c}{60k steps (492M tokens)} \\
\cmidrule(lr){2-5} \cmidrule(lr){6-9}
$\eta$ & seed 42 & seed 123 & seed 456 & mean & seed 42 & seed 123 & seed 456 & mean \\
\midrule
0.5 & -- & -- & -- & -- & 3.290 & 3.284 & 3.275 & 3.283 \\
0.7 & 3.234 & 3.237 & 3.238 & 3.236 & 3.260 & 3.257 & 3.256 & 3.258 \\
1.0 & \textbf{3.223} & \textbf{3.231} & \textbf{3.229} & \textbf{3.228} & \textbf{3.215} & \textbf{3.227} & \textbf{3.220} & \textbf{3.220} \\
1.5 & 3.636 & 3.415 & 3.500 & 3.517 & 3.315 & 3.301 & 3.254 & 3.290 \\
2.0 & 3.661 & 3.563 & 3.481 & 3.568 & -- & -- & -- & -- \\
\bottomrule
\end{tabular}
\end{table}

\emph{Chinchilla-horizon run.}
Without retuning, the same model is trained at $\eta{=}1.0$ on \textsc{FineWeb-Edu} for 190k steps at sequence length 2048 and effective batch 64, which is 24.9B tokens or about 20 tokens per parameter.
It converges without divergence to a validation loss of $2.577$, with a zero-shot average of $0.50$ over HellaSwag, PIQA, WinoGrande, ARC-Easy, ARC-Challenge, and LAMBADA (normalized accuracy where reported, accuracy otherwise).
This is one LR and one seed, so it shows that the transferred LR trains to the full horizon; it does not measure a transfer gap there.

\emph{Depth to 1.88B parameters.}
At $w{=}1024$, \MHMH{} at depth 48, 64, 96, and 128 (834M to 1.88B parameters) is swept over the fine-grained 7-point grid for 2k steps with effective batch 64 and one seed (Table~\ref{tab:depth_1p88b}).
$\eta{=}1.0$ is the argmin at depth 48 and 64 and stays within $0.043$ nats of the best grid point at depth 96 and 128.
With one seed and a short horizon, this bounds the cost of reusing $\eta{=}1.0$ rather than establishing a $0\%$ gap.

\begin{table}[H]
\centering\small
\caption{\MHMH{} at $w{=}1024$ and depth 48--128 (\textsc{WikiText-103}, 2k steps, effective batch 64, one seed, validation loss). $\eta^*$ is the best of $\{0.3, 0.5, 0.7, 1.0, 1.5, 2.0, 3.0\}$; $\Delta$ is the loss of $\eta{=}1.0$ above it.}
\label{tab:depth_1p88b}
\begin{tabular}{lcccc}
\toprule
Depth & Parameters & $\eta^*$ & $\Delta$ (nats) & $\Delta$ (\%) \\
\midrule
48 & 834M & 1.0 & 0.000 & 0.00\% \\
64 & 1.04B & 1.0 & 0.000 & 0.00\% \\
96 & 1.46B & 0.7 & 0.043 & 1.06\% \\
128 & 1.88B & 1.5 & 0.004 & 0.09\% \\
\bottomrule
\end{tabular}
\end{table}

\subsection{Fine-Grained LR Sweep}
\label{sec:results:finegrained_LR}

To rule out discrete-grid artifacts in our $0.0\%$ gap claim, we fill in intermediate LR points ($\eta \in \{0.5, 0.7, 1.5, 2.0\}$) to produce 7-point loss-vs-LR curves at each width.
The curves superimpose across widths for both \MHMH and \HHHH. At this 2k-step horizon the optimum is $\eta{=}1.5$ at every width, and $\eta{=}1.0$ costs at most $0.6\%$ (Table~\ref{tab:full_finegrained}).
The transfer gap from proxy ($w{=}256$) to target ($w{=}1024$) is $0.0\%$ for both architectures at the grid optimum; not a grid artifact.
At 20k steps (\HHHH; the intermediate points use a 1000-step warmup), the optimum is $\eta{=}1.0$ at all three widths with a basin spanning $\eta \in [0.5, 1.5]$ (per-cell spread ${\leq}0.7\%$), and at $\eta{=}2.0$ the $w{=}1024$ run becomes unstable (229 spikes, final loss $3.80$), localizing the upper edge of the basin to $\eta \!\approx\! 1.5$ at the largest width.
The full 7-point sweep at all 3 widths for both architectures is in Table~\ref{tab:full_finegrained}.

\begin{table}[H]
\centering\small
\caption{7-point loss vs. LR for \MHMH and \HHHH (\textsc{WikiText-103}, 2k steps, \naive $\muP$). Gap = $(\mathcal{L}(\eta{=}1) - \mathcal{L}^*)/\mathcal{L}^*$ in \%, the relative penalty of using the proxy-transferred LR ($\eta{=}1$) instead of the row's best-of-grid value.}
\label{tab:full_finegrained}
\begin{tabular}{llcccccccr}
\toprule
Arch & Width & $\eta$=0.3 & $\eta$=0.5 & $\eta$=0.7 & $\eta$=1.0 & $\eta$=1.5 & $\eta$=2.0 & $\eta$=3.0 & Gap @ $\eta{=}1$ \\
\midrule
\multirow{3}{*}{MHMH} & 256  & 4.411 & 4.290 & 4.262 & 4.218 & \textbf{4.202} & 4.209 & 5.523          & 0.38\% \\
                     & 512  & 4.495 & 4.354 & 4.301 & 4.265 & \textbf{4.240} & 4.249 & 4.255          & 0.59\% \\
                     & 1024 & 4.453 & 4.285 & 4.221 & 4.170 & \textbf{4.145} & \textbf{4.145} & 4.264 & 0.60\% \\
\midrule
\multirow{3}{*}{HHHH} & 256  & 4.562 & 4.468 & 4.421 & 4.403 & \textbf{4.387} & 4.399 & 4.443          & 0.36\% \\
                     & 512  & 4.504 & 4.368 & 4.294 & 4.268 & \textbf{4.255} & 4.286 & 4.349          & 0.31\% \\
                     & 1024 & 4.386 & 4.217 & 4.142 & 4.107 & \textbf{4.083} & 4.126 & 4.174          & 0.59\% \\
\bottomrule
\end{tabular}
\end{table}

\subsection{Architecture Variants}
\label{sec:appendix:full_arch}
\label{sec:appendix:arch_variants}

\subsubsection{Mamba-2 Compatibility}

Mamba-2 blocks substituted into the \MHMH layout (\textsc{WikiText-103}, 2k steps). \Naive $\muP$ sweeps the high-LR basin ($\eta \in [0.5, 5.0]$, Table~\ref{tab:full_mamba2_naive}); SP sweeps the low-LR basin ($\eta \in [0.0003, 0.03]$, Table~\ref{tab:full_mamba2_sp}). The \naive $\muP$ basin minimum is at $\eta{=}2.0$ at both widths, with $\eta{=}3.0$ tied at $w{=}256$; $\eta{=}5.0$ is past the basin. SP optimal LR shifts $10\times$ from $w{=}256$ ($\eta{=}0.01$) to $w{=}1024$ ($\eta{=}0.001$), a larger shift than Mamba-1's $3\times$ (Table~\ref{tab:full_sp_fair}); $\eta{=}0.03$ near-diverges at $w{=}1024$.

\begin{table}[H]
\centering\small
\caption{Mamba-2 \MHMH, \naive $\muP$ sweep.}
\label{tab:full_mamba2_naive}
\begin{tabular}{lccccccc r}
\toprule
W & $\eta{=}0.5$ & $\eta{=}0.7$ & $\eta{=}1.0$ & $\eta{=}1.5$ & $\eta{=}2.0$ & $\eta{=}3.0$ & $\eta{=}5.0$ & Gap @ $\eta{=}1$ \\
\midrule
256  & 4.190 & 4.142 & 4.113 & 4.117 & \textbf{4.093} & 4.096 & 4.318 & 0.48\% \\
1024 & 4.360 & 4.299 & 4.253 & 4.235 & \textbf{4.232} & 4.332 & 4.480 & 0.51\% \\
\bottomrule
\end{tabular}
\end{table}

\begin{table}[H]
\centering\small
\caption{Mamba-2 \MHMH, SP sweep.}
\label{tab:full_mamba2_sp}
\begin{tabular}{lccccc}
\toprule
W & $\eta{=}0.0003$ & $\eta{=}0.001$ & $\eta{=}0.003$ & $\eta{=}0.01$ & $\eta{=}0.03$ \\
\midrule
256  & 5.163 & 4.541 & 4.247 & \textbf{4.198} & 4.825 \\
1024 & 4.624 & \textbf{4.342} & 4.384 & 4.580 & \textit{11.3} \\
\bottomrule
\end{tabular}
\end{table}

\subsubsection{SP Fair Baseline}

\begin{table}[H]
\centering\small
\caption{SP fair baseline (\textsc{WikiText-103}, 20k steps).}
\label{tab:full_sp_fair}
\begin{tabular}{lccccc}
\toprule
 & $\eta$=0.0003 & $\eta$=0.001 & $\eta$=0.003 & $\eta$=0.01 & $\eta$=0.03 \\
\midrule
MHMH $w$=256 & 4.181 & 3.826 & \textbf{3.774} & 3.858 & 4.505 \\
MHMH $w$=1024 & 3.267 & \textbf{3.210} & 3.635 & 3.853 & \textit{11.5} \\
MTMT $w$=256 & 4.004 & 3.685 & \textbf{3.615} & 3.671 & 4.462 \\
MTMT $w$=1024 & 3.226 & \textbf{3.186} & 3.559 & \textit{11.5} & \textit{11.5} \\
\bottomrule
\end{tabular}
\end{table}

\subsubsection{Gated Linear Attention}
\label{sec:appendix:gla}

To test a linear-recurrent mixer outside the Mamba family, we replace the Mamba block with a Gated Linear Attention (GLA) layer~\citep{yang2024gla} and interleave it with Transformer blocks (\texttt{GTGT}, depth 16).
Every GLA weight is a real-valued, fan-in-scaled projection, so \naive $\muP$ applies to it as to the Transformer blocks.
Under \naive $\muP$ on \textsc{WikiText-103} (20k steps, one seed), $\eta{=}1.0$ is the argmin at every width on both training and validation loss, so the transfer gap across this fourfold width range is $0.0\%$ (Table~\ref{tab:gla}); no run diverges.

\begin{table}[H]
\centering\small
\caption{\texttt{GTGT} (GLA interleaved with Transformer blocks, depth 16) under \naive $\muP$ on \textsc{WikiText-103} (20k steps, one seed): validation loss. Bold marks the argmin of each row.}
\label{tab:gla}
\begin{tabular}{lccccc}
\toprule
Width & Parameters & $\eta{=}0.3$ & $\eta{=}1.0$ & $\eta{=}3.0$ & $\eta^*$ \\
\midrule
256 & 60M & 3.732 & \textbf{3.708} & 3.785 & 1.0 \\
512 & 137M & 3.575 & \textbf{3.556} & 3.587 & 1.0 \\
1024 & 342M & 3.488 & \textbf{3.450} & 3.501 & 1.0 \\
\bottomrule
\end{tabular}
\end{table}

\subsection{Nemotron-H Block Composition}
\label{sec:appendix:nemotron_blocks}

Section~\ref{sec:results:nemotron} reports an LR-transfer experiment on a Nemotron-H~\citep{nvidia2025nemotron} reproduction, but the headline architecture set used in the rest of the paper (T, M, H blocks) does not include the block types Nemotron-H is built from.
For reproducibility, we record the block-type letters our implementation uses and the layer-pattern they produce.

\paragraph{Block types.}
Nemotron-H uses three block types, which we expose via single-letter codes consistent with the production paper:
\begin{itemize}
    \item \textbf{N}: Mamba-2 layer~\citep{dao2024mamba2} (SSD-style selective scan; requires the \texttt{mamba\_ssm} CUDA kernel).
    \item \textbf{F}: FFN layer (squared-ReLU MLP only, no attention or SSM in this layer).
    \item \textbf{A}: attention-only layer that replaces an N in an N-F pair when sparse attention is inserted.
\end{itemize}
This is distinct from our headline T (Transformer block: GQA + SwiGLU FFN, fused) and M (Mamba-1) blocks: Nemotron-H separates the FFN into its own F layer instead of fusing it with attention or SSM, and inserts a small fraction of attention-only A layers rather than full Transformer blocks.

\paragraph{Layer pattern.}
The base pattern alternates N and F at every depth (Mamba-2 followed by MLP). A small fraction of attention layers (default $\alpha{=}0.08$) is dispersed evenly by replacing an N with an A in some N-F pairs; the first layer is always N and the last is always F. At the depths we test:
\begin{align*}
\text{depth }16&: \texttt{NFNFNFNFAFNFNFNF} \quad \text{(7~N + 8~F + 1~A; }6.2\%\text{ attention)} \\
\text{depth }32&: \texttt{NFNFNFNFNFNFNFNFAFNFNFNFNFNFNFNF} \quad \text{(15~N + 16~F + 1~A; }3.1\%\text{ attention)}
\end{align*}
The attention fraction we observe ($3$--$6\%$) is slightly below the $\sim$$8\%$ target because rounding to integer attention positions takes precedence over the exact ratio at small depth.

\paragraph{Deviations from the production specification.}
Our Nemotron-H reproduction matches the published model on the load-bearing components: Mamba-2 SSD blocks, squared-ReLU FFN, no positional encoding in attention, and an $\sim$$8\%$ attention-layer density (rounded to integer positions per depth as above).
We tested two attention implementations to bracket the production model:
\begin{itemize}
    \item \textbf{Multi-head attention (MHA) variant.} Our initial reproduction used full multi-head attention ($n_{\text{kv}} \!=\! n_{\text{heads}}$) without rotary position embeddings. This is simpler than the production architecture but probes the same N/F/A composition.
    \item \textbf{Precise GQA variant.} A second reproduction uses grouped-query attention at ratio~4 ($n_{\text{kv}} \!=\! n_{\text{heads}}/4$) with no rotary embeddings, matching the published Nemotron-H attention configuration. This is the implementation reported in Section~\ref{sec:results:nemotron} and Table~\ref{tab:nemotron}.
    \item \textbf{Dense attention pattern (no sliding window or other sparsification).} The A-layers are full dense self-attention in both variants. Nemotron-H reports sparse attention at production scale; the relevant $\muP$ analysis for sparse architectures is in \citet{dey2024sparse}, who show that sparsity-aware $\muP$ rules generalize the standard prescription. Our finding ($0.0\%$ LR transfer gap on Nemotron-H under \naive $\muP$) is therefore a result on the dense-attention reproduction and does not yet probe the sparse-pattern regime; we expect transfer to extend, but flag this as untested.
    \item \textbf{Smaller SSM state dimension.} Our reproduction uses $N_x{=}16$ across all widths, while the production Nemotron-H 8B model uses $N_x{=}128$ in its Mamba-2 SSD blocks. Our small-scale validation tests the LR-transfer mechanism rather than reproducing the production model exactly; the field-standard $N_x{=}16$ choice (also used in Mamba-1, Jamba, and Falcon-H1) keeps our setup comparable across architectures; the released-configuration runs below use $N_x{=}128$.
\end{itemize}

\paragraph{Released-configuration dimensions.}
Against the released Nemotron-H-8B configuration, our reproduction matches 10 of 14 structural properties (MLP activation, Mamba-2 head dimension and expansion, convolution kernel, GQA ratio, positional encoding, embedding tying, first- and last-layer types, and attention-layer fraction, within rounding as above) and differs in four fixed dimensions: the MLP intermediate ratio ($2.75$ vs $5.25$), the SSM state ($16$ vs $128$), the number of Mamba-2 groups ($1$ vs $8$), and the attention head dimension ($32$ vs $128$). None of them scales with width.
With these four set to the released values, and width and depth kept at our scale, \naive $\muP$ selects $\eta^*{=}1.0$ at widths 512, 1024 and 2048 on both training and validation loss, and no run diverges (Table~\ref{tab:nemotron_production}).

\begin{table}[h]
\centering
\small
\caption{Nemotron-H with the released Nemotron-H-8B configuration's MLP ratio, SSM state, Mamba-2 groups and attention head dimension (\naive $\muP$, depth 16, \textsc{WikiText-103}, 20k steps, warmup=1000). Final training loss; validation loss selects the same $\eta^*$ at every width, and the transfer gap from $w{=}512$ is $0.0\%$.}
\label{tab:nemotron_production}
\begin{tabular}{l c c c c c}
\toprule
Width (params) & $\eta{=}0.3$ & $1.0$ & $3.0$ & $5.0$ & $10.0$ \\
\midrule
512 (144M)  & 3.529 & \textbf{3.467} & 3.498 & 3.533 & 3.614 \\
1024 (355M) & 3.271 & \textbf{3.209} & 3.223 & 3.259 & 3.302 \\
2048 (980M) & 3.113 & \textbf{3.027} & 3.052 & 3.061 & 3.086 \\
\bottomrule
\end{tabular}
\end{table}

\paragraph{Longer training.}
At the transferred $\eta{=}1.0$, we trained the depth-16 reproduction of Table~\ref{tab:nemotron} on \textsc{FineWeb-Edu} at about 20 tokens per parameter, with batch 64 and sequence length 128: 57M parameters on 1.15B tokens ($w{=}256$, 140k steps) and 299M parameters on 6.0B tokens ($w{=}1024$, 733k steps). Neither run diverges; the final validation losses are $4.330$ and $3.742$.

\paragraph{MHA vs GQA: transfer property holds in both.}
Both attention implementations show $\sim$$0\%$ LR transfer gap under \naive $\muP$, but the optimum LR and absolute loss differ.
Table~\ref{tab:nemotron_mha_vs_gqa} compares the two on the same (depth, width) grid.

\begin{table}[h]
\centering
\small
\caption{Nemotron-H attention-implementation comparison (\naive $\muP$, \textsc{WikiText-103}; MHA rows 2k steps with warmup=100, GQA rows 20k steps with warmup=1000). Both variants achieve clean LR transfer; absolute losses and the optimum LR are not comparable across the two horizons.}
\label{tab:nemotron_mha_vs_gqa}
\begin{tabular}{l c c c c c c}
\toprule
Variant & Depth & Best LR & $w{=}256$ & $w{=}512$ & $w{=}1024$ & Gap \\
\midrule
MHA, no RoPE     & 16 & $3.0$ & $4.219$ & $4.120$ & $\mathbf{4.151}$ & $\mathbf{0.07\%}$ \\
MHA, no RoPE     & 32 & $3.0$ & $4.453$ & $4.113$ & $\mathbf{4.067}$ & $\mathbf{0.00\%}$ \\
\addlinespace
GQA, no RoPE     & 16 & $1.0$ & $3.760$ & $3.344$ & $\mathbf{3.369}$ & $\mathbf{0.00\%}$ \\
GQA, no RoPE     & 32 & $1.0$ & $3.527$ & $3.313$ & $\mathbf{3.138}$ & $\mathbf{0.00\%}$ \\
\bottomrule
\end{tabular}
\end{table}

The GQA rows are 20k-step runs and the MHA rows 2k-step runs, so their absolute losses and optimum LRs are not directly comparable. Both implementations preserve the headline transfer property: \naive $\muP$ with AdamW achieves $\sim$$0\%$ transfer gap independent of the attention configuration.

\paragraph{Why this is the right test for the paper's claim.}
Nemotron-H sits outside our headline grid (different block types, different per-layer composition, different attention placement strategy), so a successful $0.0\%$ LR transfer gap on either reproduction is evidence that \naive $\muP$ with AdamW transfers across genuinely independent hybrid designs, not just within the T/M/H block family our main experiments enumerate.
\section{Robustness Checks}
\label{sec:appendix:robustness}

\subsection{Convergence at 20k Steps}
\label{sec:appendix:convergence}

\begin{figure}[H]
  \centering
  \includegraphics[width=1.0\textwidth]{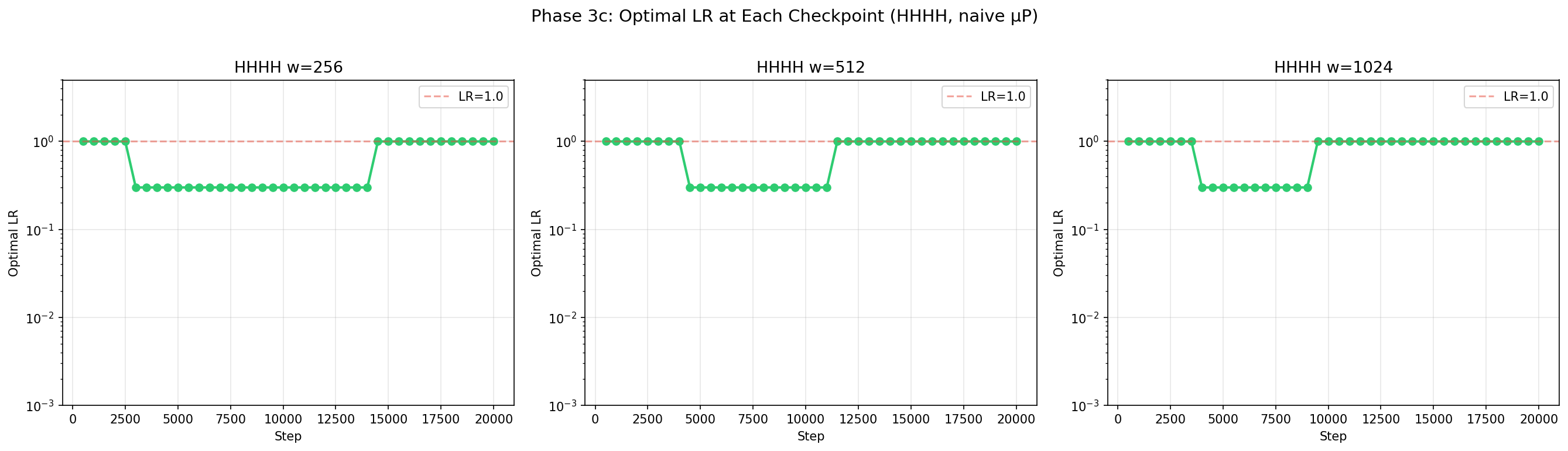}
  \caption{Optimal LR at each training checkpoint (20k steps, WikiText-103).
  At $w{=}256$, $\eta{=}0.3$ briefly leads during mid-training (steps 3k--14k) before $\eta{=}1.0$ recovers.
  By convergence, $\eta{=}1.0$ is universally optimal.}
  \label{fig:lr_trajectory}
\end{figure}

The 0.0\% transfer gap holds at convergence (20k steps on WikiText-103, Table~\ref{tab:convergence}).
Figure~\ref{fig:lr_trajectory} shows a mid-training transient (steps 3k--14k for width 256, shorter for wider architectures) where $\eta{=}0.3$ briefly leads, which is self-correcting.
SP divergence rates \emph{increase} with training duration, reaching 100\% on all three measured architectures (\HHHH, \MHMH, \MTMT) at 20k steps on the 3-LR$\times$3-width grid.

\begin{table}[H]
\centering
\small
\caption{Convergence validation (20k steps, \textsc{WikiText-103}, \naive $\muP$). Optimal LR is $\eta{=}1.0$ at all 9 (arch, width) conditions.
SP divergence rate is on the 3-LR $\times$ 3-width grid ($\eta \in \{0.3, 1.0, 3.0\}$, $w \in \{256, 512, 1024\}$, $9$ runs per arch); rates rise with training duration.}
\label{tab:convergence}
\begin{tabular}{lcccccc}
\toprule
 & \multicolumn{3}{c}{Loss at $\eta{=}1.0$} & \multicolumn{2}{c}{SP divergence rate} \\
\cmidrule(lr){2-4} \cmidrule(lr){5-6}
Arch & $w{=}256$ & $w{=}512$ & $w{=}1024$ & @2k & @20k \\
\midrule
HHHH & 3.514 & 3.270 & 3.191 & 67\% & \textbf{100\%} \\
MHMH & 3.741 & 3.317 & 3.171 & 67\% & \textbf{100\%} \\
MTMT & 3.607 & 3.495 & 3.158 & 78\% & \textbf{100\%} \\
\bottomrule
\end{tabular}
\end{table}

\subsection{Multi-Seed Validation}
\label{sec:appendix:multiseed}

\begin{figure}[H]
  \centering
  \includegraphics[width=\textwidth]{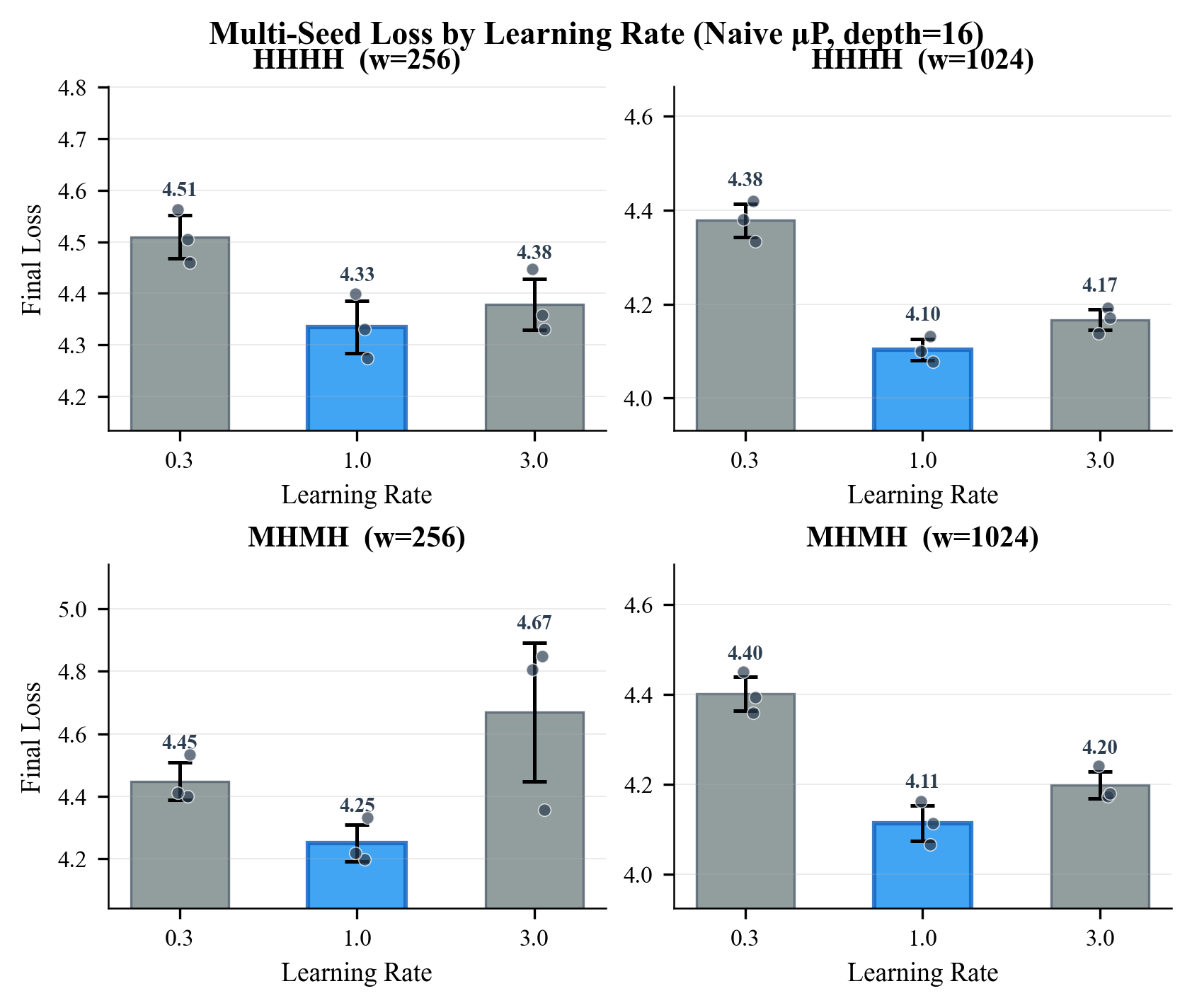}  %
  \caption{Final training loss (mean $\pm$ std) across 3 seeds (2k steps, \textsc{WikiText-103}).
  $\eta{=}1.0$ (blue) is optimal in all 12 per-seed conditions.
  Cross-seed CV $\leq$ 1.7\% at the optimum, confirming high reproducibility.
  MHMH $\eta{=}3.0$ at $w{=}256$ shows anomalously high variance (CV $\approx$ 5.8\%).}
  \label{fig:multiseed}
\end{figure}

To show that the 0.0\% gap is not a lucky draw, we repeat the width transfer experiment across 3 seeds for both HHHH and MHMH (36 configs: 2 archs $\times$ 2 widths $\times$ 3 LRs $\times$ 3 seeds). LR=1.0 is optimal in 12/12 per-seed conditions (seeds 42, 123, 456).
Transfer gap was 0.0\% at every seed.
Cross-seed CV $\leq$ 1.7\% at the optimum (mean $1.2\%$ across the four (arch, width) cells), showing that the results are highly reproducible (Figure~\ref{fig:multiseed}).

\subsection{Batch Size Robustness}
\label{sec:appendix:batchsize}

We wanted to check if the LR transfer depends on batch size or not on the hybrid architectures. 
We sweep 7 LRs at batch sizes $\{8, 16, 32, 64\}$ for both widths (256, 1024) and both architectures (Table~\ref{tab:batchsize}).
Smaller batches shift the optimal LR downward (from $\eta{\approx}1.5$ at $b{=}64$ to $\eta{\approx}0.5$--$0.7$ at $b{=}8$), but this shift is \emph{width-independent}; learning rate transfer is still preserved at all batch sizes.

\begin{table}[H]
\centering
\small
\caption{Width transfer gap at each batch size (\naive $\muP$, 2k steps, \textsc{WikiText-103}, training loss). Gap: loss at w=256 optimal LR applied to w=1024 vs.\ best at w=1024.}
\begin{tabular}{lcccc}
\toprule
Batch size & MHMH gap & HHHH gap & MHMH optimal LR & HHHH optimal LR \\
\midrule
8  & 0.5\% & 0.0\% & 0.7--1.0 & 0.5 \\
16 & 0.0\% & 1.1\% & 1.0      & 0.7--1.0 \\
32 & 0.3\% & 0.0\% & 1.5--2.0 & 1.0 \\
64 & 0.0\% & 0.0\% & 1.5--2.0 & 1.5 \\
\bottomrule
\end{tabular}
\label{tab:batchsize}
\end{table}

All gaps are $\leq$1.1\%.
The practical implication: batch size shifts the optimal LR uniformly across widths, so practitioners can change batch size and re-tune at proxy scale; the transferred LR remains valid at target scale.

\subsection{Sequence Length Scaling}
\label{sec:appendix:seqlen}

72 configs: HHHH/MHMH/TT $\times$ $w \in \{256, 1024\}$ $\times$ 3 LRs $\times$ 4 seq\_lens (128--1024) under \naive $\muP$.
Batch size adjusted to keep tokens/step constant at 8192.
$\eta{=}1.0$ is optimal across all sequence lengths 128--1024, all 3 architectures, and both widths (24/24 conditions); zero divergence across 72 configs.
Extending the sweep to seq$=$2048, the length of the 5.0B- and 24.9B-token runs, adds 18 runs and gives the same result: $\eta{=}1.0$ is optimal in all 6 (architecture, width) cells on both training and validation loss, the width-transfer gap is $0.0\%$, and none of the 90 runs diverges.
Final loss at the optimum is reported in Table~\ref{tab:seqlen}; MHMH leads at seq$=$1024, consistent with SSM's long-context advantage.

\begin{table}[H]
\centering
\small
\caption{Sequence length scaling at $\eta{=}1.0$ (\naive $\muP$, 2k steps, \textsc{WikiText-103}). Final loss at each (arch, $w$, seq\_len). $\eta{=}1.0$ is optimal at every cell; LR transfer is sequence-length invariant for tested lengths 128--1024.}
\label{tab:seqlen}
\begin{tabular}{llcccc}
\toprule
Arch & Width & seq=128 & seq=256 & seq=512 & seq=1024 \\
\midrule
\multirow{2}{*}{HHHH} & $w{=}256$  & 4.399 & 4.100 & \textbf{4.074} & 4.477 \\
                     & $w{=}1024$ & 4.102 & 4.262 & 4.215 & 4.509 \\
\midrule
\multirow{2}{*}{MHMH} & $w{=}256$  & 4.213 & 4.388 & 4.580 & \textbf{4.084} \\
                     & $w{=}1024$ & 4.161 & 4.323 & 4.300 & 4.338 \\
\midrule
\multirow{2}{*}{TT}   & $w{=}256$  & 4.258 & 4.235 & 4.454 & 4.267 \\
                     & $w{=}1024$ & 4.255 & 4.459 & 4.207 & 4.547 \\
\bottomrule
\end{tabular}
\end{table}

LR sensitivity across sequence length follows the same architecture ranking we observe at fixed seq\_len: HHHH stays within $0.01$--$0.10$ between $\eta{=}1.0$ and $\eta{=}3.0$, MHMH within $0.02$--$1.4$ (one outlier at $w{=}1024$, seq=256), and TT spans $0.41$--$0.70$, $\sim 10\times$ more sensitive than HHHH; hybrid architectures degrade more gracefully under LR misspecification.

\subsection{T:M Ratio Transfer}
\label{sec:appendix:tm_ratio}

We test whether \naive $\muP$ LR transfer holds across different Transformer-to-SSM parameter ratios within H blocks.
The \texttt{ssm\_expand} parameter controls $\dinner = \dmodel \times \texttt{expand}$, which determines the Mamba sub-block size relative to the Transformer sub-block; the resulting per-block parameter breakdown is shown in Table~\ref{tab:tm_params}.

\begin{table}[H]
\centering
\caption{H-block parameter breakdown by \texttt{ssm\_expand} ($\dmodel=512$).}
\label{tab:tm_params}
\begin{tabular}{ccrrrrrr}
\toprule
\texttt{expand} & $\dinner$ & Attention & SwiGLU FFN & T total & M total & T:M Ratio & H block \\
\midrule
1 & 512  & 1.05M & 2.16M & 3.21M & 0.85M & 3.8:1 & 4.06M \\
2 & 1024 & 1.05M & 2.16M & 3.21M & 1.69M & 1.9:1 & 4.91M \\
4 & 2048 & 1.05M & 2.16M & 3.21M & 3.39M & 0.9:1 & 6.60M \\
\bottomrule
\end{tabular}
\end{table}

\begin{figure}[H]
  \centering
  \includegraphics[width=\textwidth]{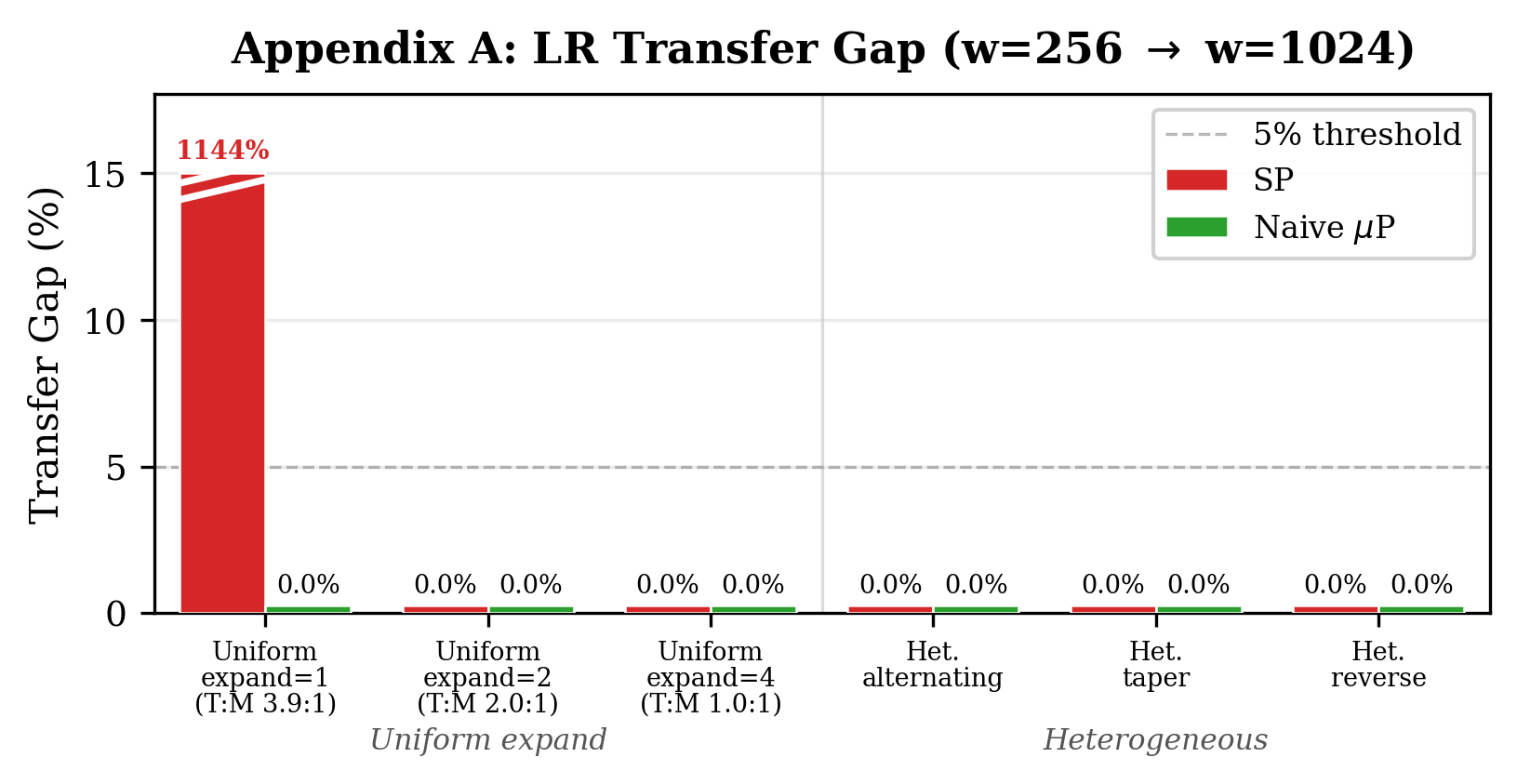}
  \caption{Transfer gap across T:M ratios (\naive $\muP$ and SP, \textsc{WikiText-2}, 2k steps, warmup 100; training loss).
  \Naive $\muP$: 0.0\% gap for all 6 configurations.
  SP: 0.0\% for 5/6 but catastrophic (1144\%) for \texttt{expand1} where the optimal LR shifts with width.
  LR=1.0 is invariant to the T:M ratio; no re-tuning needed when changing \texttt{ssm\_expand}.}
  \label{fig:tm_ratio}
\end{figure}

We test 3 uniform ratios plus 3 heterogeneous per-layer patterns (\texttt{het\_alt}, \texttt{het\_taper}, \texttt{het\_reverse}).
252 configs total: 6 patterns $\times$ 3 widths $\times$ 2 schemes $\times$ 7 LRs.
\Naive $\muP$ achieves 0.0\% transfer gap for all 6 configurations (Figure~\ref{fig:tm_ratio}).

\noindent
\textbf{Note on SP transferring at $5/6$ ratios.}
SP appears to transfer cleanly on five of the six configurations because over the tested width range its optimum happens to coincide at the proxy and target widths for those particular layouts; this is a feature of the limited grid here rather than a contradiction of our broader finding.
The architecture-spanning width-transfer sweep (Table~\ref{tab:width_transfer}) confirms SP's optimum drifts $3$--$10\times$ on every arch we tested.
The \texttt{expand=1} cell is the critical case for this appendix: the SP optimum genuinely shifts with width, producing the $1144\%$ catastrophic gap, while \naive $\muP$ stays at $\eta{=}1.0$.

\texttt{het\_taper} (T-dominant early $\to$ SSM-heavy late) achieves the best loss ($0.114$), beating uniform \texttt{expand=2} ($0.130$) by $\sim$$12\%$.
This suggests more SSM capacity in late layers is architecturally beneficial.

\subsection{Serial T:M Ratio Sweep}
\label{sec:appendix:jamba_imbalanced}

Section~\ref{sec:appendix:tm_ratio} covers parallel imbalance inside an H block; this subsection covers \emph{serial} imbalance, where Transformer and Mamba blocks are interleaved at varying ratios along the depth axis.
We sweep six fixed-depth ($d{=}16$) ratios spanning the full T--M spectrum:
\texttt{TT} (0\% M), \texttt{M1T3} ($\texttt{MTTTMTTTMTTTMTTT}$, 25\% M), \texttt{MTMT} (50\% M), \texttt{M3T1} ($\texttt{MMMTMMMTMMMTMMMT}$, 75\% M), Jamba (\texttt{MMMMMMMTMMMMMMMT}, 87.5\% M), and \texttt{MM} (100\% M).
Each is run at $\eta \in \{0.003, 0.01, 0.03, 0.1, 0.3, 1.0, 3.0\}$ across $w \in \{256, 512, 1024\}$ under SP, \naive $\muP$, approximated block-wise $\muP$, and the full block-wise prescription (2k steps on \textsc{WikiText-2}).

\begin{table}[H]
\centering
\small
\caption{Serial T:M ratio transfer at depth $16$. Optimal $\eta^*$ at each (scheme, pattern, width); shading mirrors Table~\ref{tab:width_transfer}, where uniform color within a row indicates clean LR transfer. \Naive $\muP$ holds $\eta^*{=}1.0$ across the full ratio range; approximated block-wise $\muP$ drifts on the SSM-bearing layouts; SP drifts on every layout where the SSM density is non-trivial. The full block-wise prescription shifts on every SSM-bearing layout, with $\eta{=}3.0$ diverging at $w{=}1024$ on each.}
\label{tab:serial_ratio}
\resizebox{\columnwidth}{!}{%
\begin{tabular}{l c ccc ccc ccc ccc}
\toprule
 & & \multicolumn{3}{c}{\Naive $\muP$} & \multicolumn{3}{c}{Block-wise $\muP$ (approx.)} & \multicolumn{3}{c}{Block-wise $\muP$} & \multicolumn{3}{c}{SP} \\
\cmidrule(lr){3-5} \cmidrule(lr){6-8} \cmidrule(lr){9-11} \cmidrule(lr){12-14}
Pattern & \%M & $w{=}256$ & $w{=}512$ & $w{=}1024$ & $w{=}256$ & $w{=}512$ & $w{=}1024$ & $w{=}256$ & $w{=}512$ & $w{=}1024$ & $w{=}256$ & $w{=}512$ & $w{=}1024$ \\
\midrule
\texttt{TT}          & 0\%    & \cellcolor{lr1}\textbf{1.0} & \cellcolor{lr1}\textbf{1.0} & \cellcolor{lr1}\textbf{1.0} & \cellcolor{lr1}1.0 & \cellcolor{lr1}1.0 & \cellcolor{lr1}1.0 & \cellcolor{lr1}1.0 & \cellcolor{lr1}1.0 & \cellcolor{lr1}1.0 & \cellcolor{lr003}0.003 & \cellcolor{lr003}0.003 & \cellcolor{lr003}0.003 \\
\texttt{M1T3}        & 25\%   & \cellcolor{lr1}\textbf{1.0} & \cellcolor{lr1}\textbf{1.0} & \cellcolor{lr1}\textbf{1.0} & \cellcolor{lr1}1.0 & \cellcolor{lr3}3.0 & \cellcolor{lr3}3.0 & \cellcolor{lr1}1.0 & \cellcolor{lr3}3.0 & \cellcolor{lr1}1.0 & \cellcolor{lr003}0.003 & \cellcolor{lr003}0.003 & \cellcolor{lr003}0.003 \\
\texttt{MTMT}        & 50\%   & \cellcolor{lr1}\textbf{1.0} & \cellcolor{lr1}\textbf{1.0} & \cellcolor{lr1}\textbf{1.0} & \cellcolor{lr1}1.0 & \cellcolor{lr1}1.0 & \cellcolor{lr1}1.0 & \cellcolor{lr1}1.0 & \cellcolor{lr3}3.0 & \cellcolor{lr1}1.0 & \cellcolor{lr01}0.010 & \cellcolor{lr003}0.003 & \cellcolor{lr003}0.003 \\
\texttt{M3T1}        & 75\%   & \cellcolor{lr1}\textbf{1.0} & \cellcolor{lr1}\textbf{1.0} & \cellcolor{lr1}\textbf{1.0} & \cellcolor{lr1}1.0 & \cellcolor{lr1}1.0 & \cellcolor{lr3}3.0 & \cellcolor{lr1}1.0 & \cellcolor{lr3}3.0 & \cellcolor{lr1}1.0 & \cellcolor{lr01}0.010 & \cellcolor{lr003}0.003 & \cellcolor{lr003}0.003 \\
\texttt{JAMBA} (1:7) & 87.5\% & \cellcolor{lr1}\textbf{1.0} & \cellcolor{lr1}\textbf{1.0} & \cellcolor{lr1}\textbf{1.0} & \cellcolor{lr1}1.0 & \cellcolor{lr3}3.0 & \cellcolor{lr3}3.0 & \cellcolor{lr3}3.0 & \cellcolor{lr3}3.0 & \cellcolor{lr1}1.0 & \cellcolor{lr01}0.010 & \cellcolor{lr003}0.003 & \cellcolor{lr003}0.003 \\
\texttt{MM}          & 100\%  & \cellcolor{lr1}\textbf{1.0} & \cellcolor{lr1}\textbf{1.0} & \cellcolor{lr1}\textbf{1.0} & \cellcolor{lr1}1.0 & \cellcolor{lr3}3.0 & \cellcolor{lr3}3.0 & \cellcolor{lr3}3.0 & \cellcolor{lr3}3.0 & \cellcolor{lr1}1.0 & \cellcolor{lr01}0.010 & \cellcolor{lr003}0.003 & \cellcolor{lr003}0.003 \\
\midrule
\multicolumn{2}{l}{Transfers?} & \multicolumn{3}{c}{\textbf{Yes} (0/6 shift)} & \multicolumn{3}{c}{Partially (4/6 shift)} & \multicolumn{3}{c}{Drifts (5/6 shift)} & \multicolumn{3}{c}{No (4/6 shift)} \\
\bottomrule
\end{tabular}%
}
\end{table}

\noindent
\textit{Faithful block-wise $\muP$ on this grid:} the literal Vankadara prescription shifts on every SSM-bearing layout, as in the width-transfer table (Table~\ref{tab:width_transfer}).
At $w{=}1024$, $\eta{=}3.0$ diverges on each of them, so the optimum falls to $\eta{=}1.0$; at smaller widths it is $\eta{=}3.0$ (Jamba and \MM{} at $w{\leq}512$, \texttt{M1T3}, \MTMT{}, and \texttt{M3T1} at $w{=}512$).

\noindent
\textbf{Approximated block-wise $\muP$ across the ratio range.}
At the proxy width $w{=}256$, $\eta^*{=}1.0$ for every ratio, matching \naive.
The drift only emerges at larger widths: \texttt{TT} and \texttt{MTMT} stay at $\eta^*{=}1.0$ throughout; \texttt{M3T1} shifts only at $w{=}1024$; \texttt{M1T3}, Jamba, and \texttt{MM} shift starting at $w{=}512$.
Because the proxy is at $\eta{=}1.0$, transferring it to the target width still trains, but the residual loss penalty grows with target SSM density.
Table~\ref{tab:serial_ratio_loss} reports the absolute loss at the transferred $\eta{=}1.0$ versus the per-cell optimum for the SSM-bearing patterns at $w{=}1024$: penalties stay near $0\%$ for \texttt{TT} and \texttt{MTMT}, climb to $4.2\%$ on \texttt{M1T3}, and reach $24$--$36\%$ on the heavily SSM-dominated \texttt{M3T1}, Jamba, and \texttt{MM} layouts.

\begin{table}[H]
\centering
\small
\caption{Approximated block-wise $\muP$ residual loss penalty at $w{=}1024$ (\textsc{WikiText-2}, 2k steps). Penalty is the fractional loss increase incurred by transferring the proxy optimum $\eta{=}1.0$ instead of using each row's per-cell optimum at $w{=}1024$. \Naive $\muP$ has $0\%$ penalty on every row.}
\label{tab:serial_ratio_loss}
\begin{tabular}{lcccc}
\toprule
Pattern & \%M & Loss at $\eta{=}1.0$ & Best loss (best $\eta$) & Penalty \\
\midrule
\texttt{TT}    & 0\%    & 1.048 & 1.048 ($\eta{=}1.0$) & $0.0\%$ \\
\texttt{M1T3}  & 25\%   & 0.755 & 0.725 ($\eta{=}3.0$) & $4.2\%$ \\
\texttt{MTMT}  & 50\%   & 0.734 & 0.734 ($\eta{=}1.0$) & $0.0\%$ \\
\texttt{M3T1}  & 75\%   & 0.818 & 0.600 ($\eta{=}3.0$) & $36.4\%$ \\
\texttt{JAMBA} & 87.5\% & 0.852 & 0.630 ($\eta{=}3.0$) & $35.2\%$ \\
\texttt{MM}    & 100\%  & 1.144 & 0.916 ($\eta{=}3.0$) & $24.8\%$ \\
\bottomrule
\end{tabular}
\end{table}

\noindent
\textbf{Faithful block-wise $\muP$ on Jamba.}
The faithful Vankadara prescription~\citep{vankadara2024feature} drifts on Jamba at 2k steps, mirroring the pattern seen on every other SSM-bearing architecture in Section~\ref{sec:appendix:width_transfer}: $\eta^*{=}3.0$ at $w \!\leq\! 512$ shifts to $\eta^*{=}1.0$ at $w{=}1024$ as $\eta_A \!\propto\! N_u$ destabilizes training (the $w{=}1024,\ \eta{=}3$ run diverges).
Compared to the approximated variant whose drift onset is partial (T-dominant ratios still transfer cleanly), the full prescription shifts on \emph{every} ratio with non-trivial SSM density, consistent with the observation in Table~\ref{tab:width_transfer} that block-wise $\muP$'s $\eta^*$ does not transfer cleanly on any SSM-bearing layout.

\noindent
\textbf{SP at any imbalance.}
SP's optimum drifts on $4$ of the $6$ ratios (\texttt{MTMT}, \texttt{M3T1}, Jamba, \texttt{MM}); on the pure-T and 25\%-M layouts (\texttt{TT}, \texttt{M1T3}) the SP optimum happens to coincide between proxy and target widths over our grid, but this is a feature of the local optimum rather than a transfer property (Table~\ref{tab:width_transfer} shows SP drifts on every architecture once the LR grid extends below $\eta{=}0.003$).
For Jamba specifically, the proxy LR ($\eta{=}0.01$ at $w{=}256$) produces catastrophic loss at $w{=}1024$ where $\eta^*{=}0.003$.

Combined with the parallel-imbalance evidence in Section~\ref{sec:appendix:tm_ratio}, this covers every serial T:M ratio used by the production hybrids of Section~\ref{sec:background:gap}.

\subsection{Gradient Clipping Threshold}
\label{sec:appendix:grad_clip}

We use $\text{grad\_clip}{=}1.0$ throughout the main experiments and verify that the LR transfer claim does not depend on this specific threshold.
We sweep $\text{grad\_clip} \in \{\text{off}, 0.5, 1.0, 5.0\}$ on \MHMH at $w \in \{256, 1024\}$ and $\eta \in \{0.5, 1.0, 1.5\}$ for 2k steps on \textsc{WikiText-103} (24 configs).
At the $\muP$ optimum $\eta{=}1.0$, the final loss varies by $\leq 0.020$ nats across clip values at both widths, and the optimal LR is unchanged.
Disabling clipping entirely does not destabilize training (max-grad-norm trajectories stay below $\sim$5 throughout), confirming that gradient clipping is not load-bearing for the LR-transfer property and that the headline claim is robust to this hyperparameter.

\subsection{Where \Naive $\muP$ Transfer Does Not Extend}
\label{sec:appendix:limits}

The headline claim of $0.0\%$ LR transfer gap holds for \naive $\muP$ + AdamW across the architectures, widths, and depths in our main grid.
This subsection documents four boundary conditions where the transfer property weakens or does not apply.
Each case fits the two-condition decomposition of Section~\ref{sec:analysis:two_conditions}; together they sharpen, not contradict, the headline claim.

\textbf{(i) Standard parameterization.}
SP is the baseline against which $\muP$ was originally designed; without per-component re-scaling, the optimum LR shifts $3$--$10\times$ with width across the architectures we tested (Appendix~\ref{sec:appendix:width_transfer}).
On the imbalanced Jamba layout (Section~\ref{sec:appendix:jamba_imbalanced}), the SP optimum drifts from $\eta{=}0.01$ at $w{=}256$ to $\eta{=}0.003$ at $w{=}1024$, with the proxy LR producing catastrophic loss at the target width.
This is the regime $\muP$ was designed to fix; our headline claim is measured against it.

\textbf{(ii) Approximated block-wise $\muP$ on SSM-heavy serial layouts.}
On the serial T:M sweep (Section~\ref{sec:appendix:jamba_imbalanced}, Table~\ref{tab:serial_ratio}), the approximated optimum drifts by one grid step at $w \!\geq\! 512$ for SSM-bearing layouts (\texttt{M1T3}, Jamba, \texttt{MM}; \texttt{M3T1} only at $w{=}1024$), staying at $\eta^*{=}1.0$ on \texttt{TT} and \texttt{MTMT}.
Reusing the proxy $\eta{=}1.0$ at $w{=}1024$ incurs a residual loss penalty of $4$--$36\%$ on the SSM-dominated layouts (Table~\ref{tab:serial_ratio_loss}); the same parameterization transfers cleanly with $0\%$ gap on the balanced \MTMT{} layout in the main results.

\textbf{(iii) Faithful block-wise $\muP$ under AdamW at scale.}
The SSM-specific rules of \citet{vankadara2024feature} were derived and validated for pure SSMs trained with SGD in the infinite-width limit.
On hybrids at 20k steps on \textsc{WikiText-103}, the faithful prescription produces uniformly higher loss than \naive $\muP$ across the architectures and widths we measured: $+5.6$ to $+7.7\%$ on HHHH, MHMH, and MTMT at $w \in \{256, 512, 1024\}$ (full table in Appendix~\ref{sec:appendix:vankadara_faithful}).
At the 2k-step transfer horizon, its optimum LR also drifts with width on every SSM-bearing architecture ($\eta^*{=}1$ at $w{=}1024$ across all $6$, with $\eta{=}3$ diverging at that width on each); only pure-Transformer (\TT) transfers cleanly.
The approximated variant used in the main experiments stays within $0.5\%$ of \naive $\muP$ at $w{=}1024$ by dropping the $\eta_A \in \Theta(N_u)$ term that AdamW's $g/\sqrt{v}$ already supplies (Section~\ref{sec:analysis:redundant_normalization}, Appendix~\ref{sec:appendix:a_scaling}).

\textbf{(iv) Optimizers without per-parameter normalization.}
Plain SGD does not exhibit the same transfer behavior even when $\muP$ is applied (full optimizer ablation in Section~\ref{sec:appendix:optimizer}).
This localizes the LR-transfer property of \naive $\muP$ + AdamW to the per-parameter-normalization mechanism ($g/\sqrt{v}$), which matches the two-condition decomposition: $\muP$ alone does not deliver transfer in our hybrid setting without an optimizer that supplies Condition~2.

\Naive $\muP$ + AdamW transfers cleanly because $\muP$ supplies the global condition and AdamW supplies the local one. Adding component-specific scaling on top does not help at the widths and budgets we test.

\section{Analysis Details}
\label{sec:appendix:analysis}

\subsection{Coordinate Check}
\label{sec:appendix:coord_check}

Standard $\muP$ validation: train for 10 steps at 5 widths (128, 256, 512, 1024, 2048) with 5 seeds and record mean absolute activation per layer.
Under correct $\muP$, lines should overlap; under SP, they diverge.

\begin{figure}[H]
  \centering
  \includegraphics[width=\textwidth]{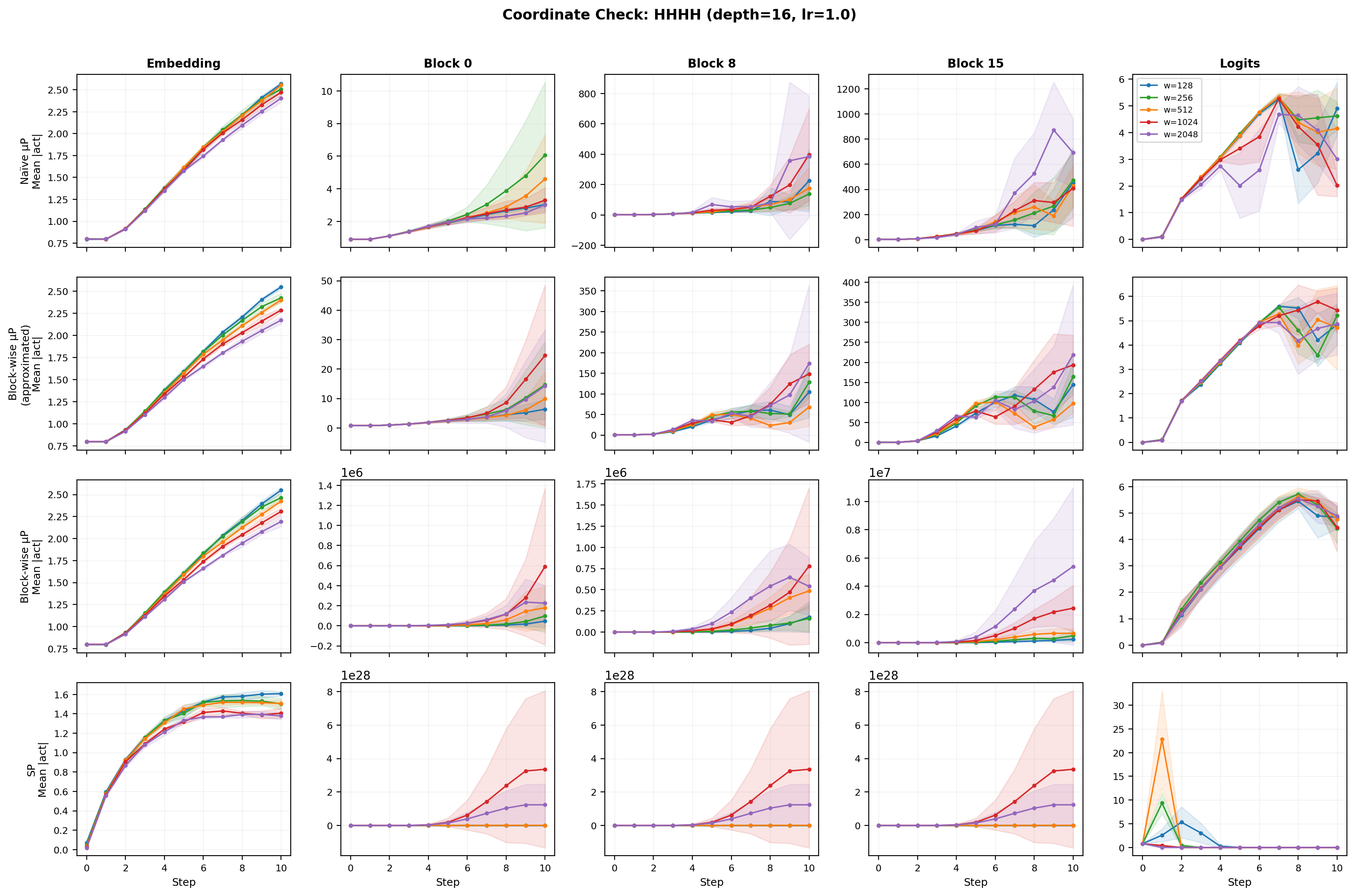}
  \caption{Coordinate check for HHHH ($d{=}16$, widths 128--2048).
  Row 1: \naive $\muP$, hidden activations vary with width (median block width-spread $2.78\times$, Table~\ref{tab:coord_check_spread}; violates textbook $\muP$).
  Row 2: block-wise $\muP$ (approximated), similar pattern.
  Row 3: block-wise $\muP$ (literal Vankadara Table~1), hidden activations predominantly $10^3$--$10^4\times$ larger in absolute magnitude than \naive at deeper blocks, with the largest width-spread.
  Row 4: SP, activations explode.
  Despite the hidden-state divergence, logit-level magnitudes (rightmost column) are comparable across all $\muP$ variants ($2.0$--$5.4$ at step 10, with width-spread $\leq$$2.5\times$ within each variant), suggesting layer-wise normalization absorbs much of the width-dependence; \naive $\muP$ achieves $0.0\%$ LR transfer under Adam.
  Each row has its own y-axis.}
  \label{fig:coord_check_appendix_hhhh}
  \label{fig:coord_check_appendix}
\end{figure}

\begin{figure}[H]
  \centering
  \includegraphics[width=\textwidth]{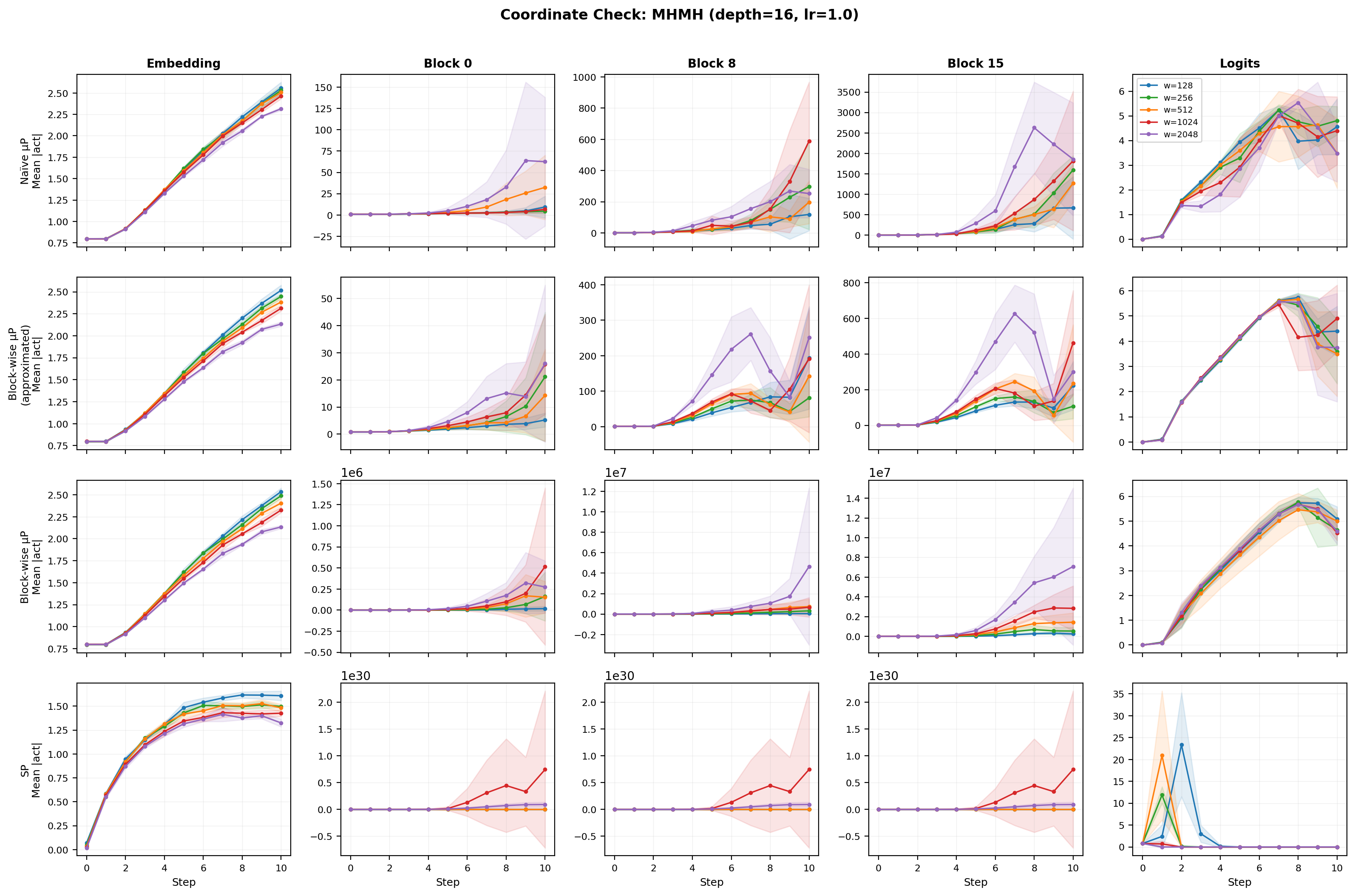}
  \caption{Coordinate check for MHMH ($d{=}16$, widths 128--2048). Same pattern as HHHH. Each row has its own y-axis.}
  \label{fig:coord_check_appendix_mhmh}
\end{figure}

Key observation: the activation magnitude is monotonic with depth (block\_0 $<$ block\_8 $<$ block\_15), suggesting cumulative residual stream amplification.

\begin{figure}[H]
  \centering
  \includegraphics[width=\textwidth]{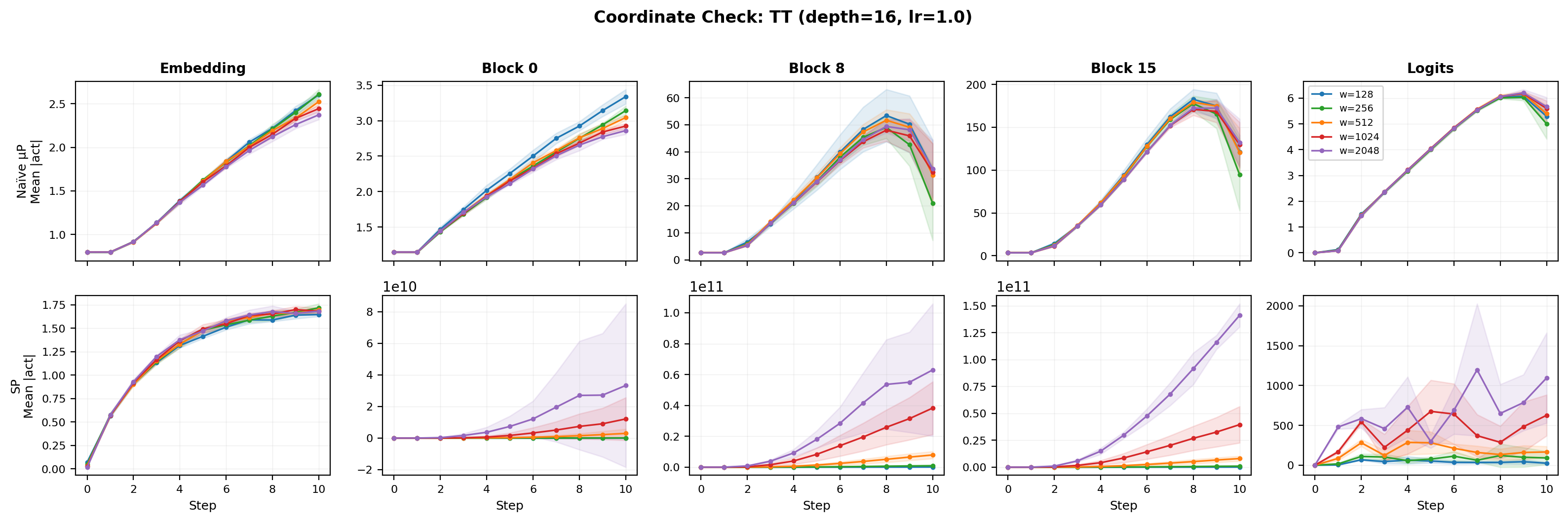}
  \caption{Coordinate check for \texttt{TT} (pure Transformer, $d{=}16$, widths 128--2048) with the same instrument.
  Row 1: \naive $\muP$, width lines overlap (median block width-spread $1.52\times$, Table~\ref{tab:coord_check_spread}).
  Row 2: SP, activations diverge with width.}
  \label{fig:coord_check_appendix_tt}
\end{figure}

\paragraph{Width-spread summary.}
Table~\ref{tab:coord_check_spread} summarizes each coordinate check by one number.
For each hidden block, we average the mean absolute activation over the five seeds at step 10 and take the ratio of its largest to its smallest value across the five widths; the table reports the median of this ratio over the 16 blocks.
A width-invariant block gives 1.
The pure Transformer (\texttt{TT}, Figure~\ref{fig:coord_check_appendix_tt}) and the GLA--Transformer hybrid (\texttt{GTGT}, Appendix~\ref{sec:appendix:gla}) are included for comparison.

\begin{table}[H]
\centering\small
\caption{Median width-spread of the coordinate check (step 10, widths 128--2048, five seeds, $d{=}16$): for each hidden block, the largest over the smallest seed-averaged mean absolute activation across widths, and the median over the 16 blocks. A value of 1 means width-invariant.}
\label{tab:coord_check_spread}
\begin{tabular}{llc}
\toprule
Arch & Scheme & Median width-spread \\
\midrule
\texttt{TT} & \Naive $\muP$ & 1.52 \\
 & SP & 819 \\
\texttt{GTGT} & \Naive $\muP$ & 1.23 \\
 & SP & $7.3{\times}10^{3}$ \\
\texttt{HHHH} & \Naive $\muP$ & 2.78 \\
 & Block-wise (approx.) & 2.45 \\
 & Block-wise (faithful) & 5.71 \\
 & SP & $6.0{\times}10^{7}$ \\
\texttt{MHMH} & \Naive $\muP$ & 4.23 \\
 & Block-wise (approx.) & 3.33 \\
 & Block-wise (faithful) & 65.67 \\
 & SP & $1.4{\times}10^{9}$ \\
\bottomrule
\end{tabular}
\end{table}

\subsection{Gradient Measurements}
\label{sec:appendix:gradient}

On pure Transformers, \citet{haas2025surprising} identify a large-LR controlled-divergence regime where even SP achieves non-trivial feature learning at infinite width; our gradient measurements test whether AdamW's per-parameter normalization plays an analogous role across the SSM boundary in the heterogeneous hybrid setting.
We measure per-component gradient RMS at initialization across widths 128--1024 to quantify the per-component gradient imbalance.

\begin{figure}[H]
  \centering
  \includegraphics[width=\textwidth]{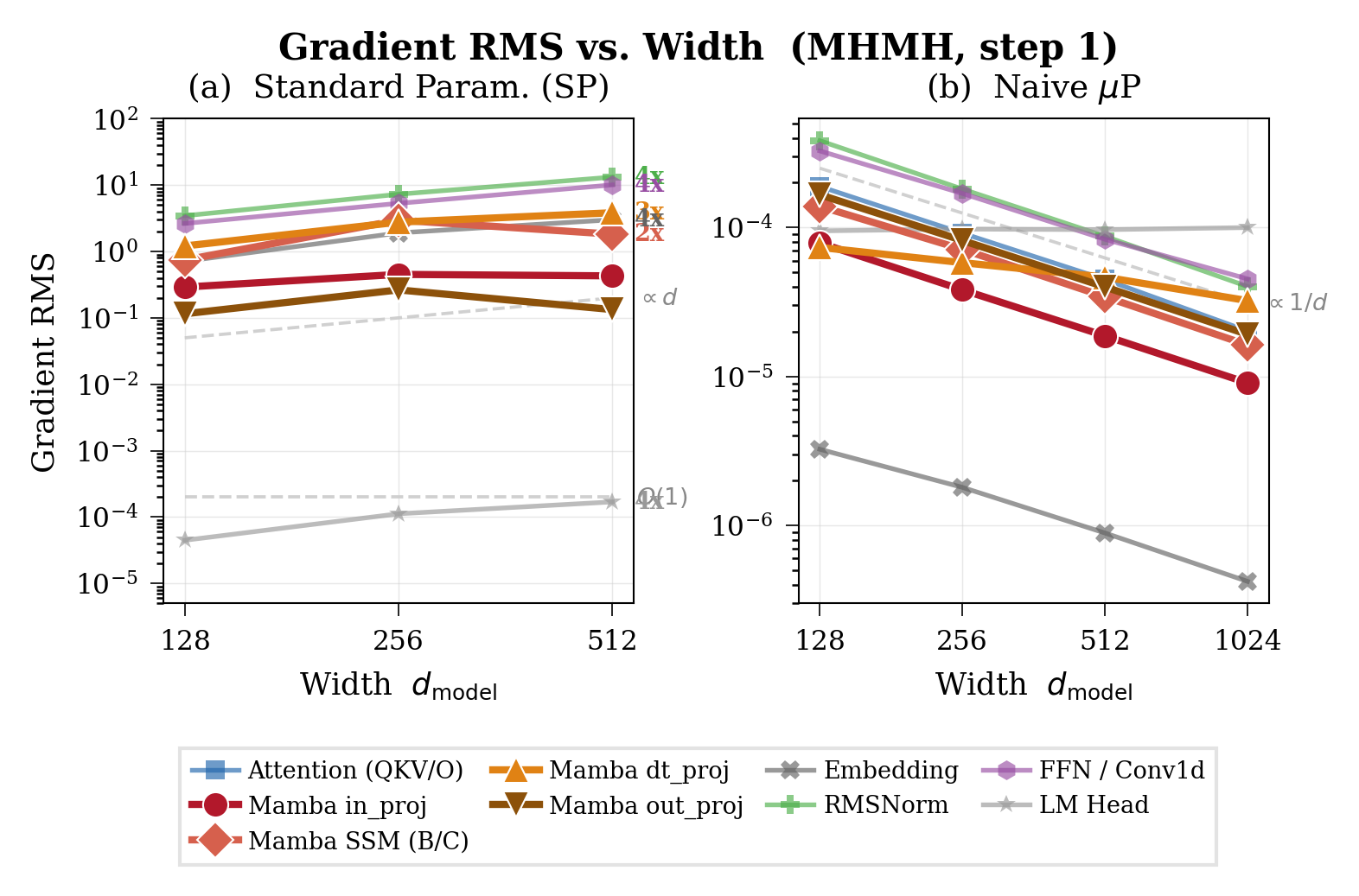}
  \caption{Gradient RMS per component vs.\ width for \MHMH{} at step 1, under SP (widths 128--512) and \naive $\muP$ (128--1024).
  Components differ widely in gradient scale.
  Under Adam, the $1/\sqrt{v}$ denominator normalizes these disparate scales.}
  \label{fig:appendix_gradient_rms}
\end{figure}

These measurements support the two-condition decomposition (Section~\ref{sec:analysis}): \naive $\muP$ satisfies Condition~1 (global update-to-weight ratio) but not per-component gradient balance.
Adam's adaptive normalization restores Condition~2, enabling LR transfer despite the gradient imbalance.

\paragraph{Total gradient norm at initialization.}
At the aggregate level, the total gradient norm at initialization (single forward-backward pass on the MT interface, no optimizer step) shows the same SP-vs-$\muP$ separation seen at the per-component level (Table~\ref{tab:gradient_norm_init}).
SP produces gradient norms that grow from $0.6$ to $2.5$ across $w{=}128$--$2048$, while both $\muP$ variants stay between $0.1$ and $0.5$ at the same widths, $5\times$ lower at $w{=}2048$.

\begin{table}[H]
\centering
\small
\caption{Gradient norm at initialization across widths and schemes (MT interface).}
\label{tab:gradient_norm_init}
\begin{tabular}{lccccc}
\toprule
 & \multicolumn{5}{c}{Width} \\
\cmidrule(lr){2-6}
Scheme & 128 & 256 & 512 & 1024 & 2048 \\
\midrule
SP & 0.6 & 0.9 & 1.2 & 1.7 & 2.5 \\
\Naive $\muP$ & 0.1 & 0.2 & 0.3 & 0.4 & 0.5 \\
Block-wise $\muP$ & 0.1 & 0.2 & 0.3 & 0.4 & 0.5 \\
\bottomrule
\end{tabular}
\end{table}

SP's gradient norm grows ${\sim}4\times$ from $w{=}128$ to $w{=}2048$ ($0.60 \!\to\! 2.48$); both $\muP$ variants grow by a similar factor (${\sim}4\times$) but from a much lower base ($0.12 \!\to\! 0.50$).
Notably, \naive and block-wise $\muP$ (both variants) produce nearly identical gradient norms at initialization; the divergence between them emerges only during training under Adam (Section~\ref{sec:analysis}).

\paragraph{Joint $(N_u, N_x, d)$ gradient-scaling sweep.}
The 1D measurement above varies only $d_{\text{model}}$ at fixed $\texttt{ssm\_expand}=2$, so $N_u = 2d$ is perfectly collinear with $d$ and the 1D fit against $N_u$ alone cannot separate a dependence on $d$ from a dependence on $N_u$.
To break this degeneracy we run a second, larger sweep varying the three axes independently: $d_{\text{model}} \in \{128, 256, 512\}$, $\texttt{ssm\_expand} \in \{1, 2, 4\}$ (so $N_u = \texttt{expand}\cdot d$ spans $[128, 2048]$), $\texttt{ssm\_d\_state} \in \{8, 16, 32, 64\}$ (so $N_x$ varies independently), each for 3 seeds and 200 training steps.

\paragraph{Mapping to Vankadara's $A$, $B$, $C$.}
The Mamba block parameters in our measurement correspond to the components of \citet{vankadara2024feature}'s framework as follows: $\texttt{mamba\_A\_log}$ is the SSM state-transition matrix $A$; the $B$ and $C$ matrices in Vankadara's prescription are slices of the input-projection $\texttt{x\_proj}$ (the rows producing the per-step $B$ and $C$ at each position).
We use $\texttt{ssm\_split\_projections}{=}\text{True}$ for this sweep so that $B$ and $C$ are separate parameter groups, allowing Vankadara's per-component rules ($\eta_B \!\propto\! 1/\sqrt{N_u}$, $\eta_C \!\propto\! 1/(N_x \sqrt{N_u})$) to act at the granularity their derivation assumes; this matches the configuration used in the main width-transfer experiments (Table~\ref{tab:width_transfer}).
The labels $\texttt{mamba\_in\_proj}$ ($d_\text{model}\!\to\!2 d_\text{inner}$ input gate) and $\texttt{mamba\_out\_proj}$ ($d_\text{inner}\!\to\!d_\text{model}$) are \emph{not} $B$ or $C$; they are the input/output projections that surround the SSM core.

\paragraph{Effective update at step 1: the direct $\muP$ test under SGD.}
$\muP$ theory constrains the \emph{update} ($\eta_{\text{eff}}(W, \text{scheme}) \cdot \text{grad}(W)$), not the raw gradient.
Vankadara's prescription predicts that the effective update is width-invariant ($\alpha(N_u)\!=\!0$) under SGD when the per-component rules are properly applied.
We compute the step-1 effective update RMS for $A_{\log}$, $B$, and $C$ under both \naive $\muP$ and the faithful Vankadara prescription, and fit $\log u \sim \alpha \log N_u + \gamma \log d + \log C$ over the joint $(d, \texttt{expand}, \texttt{d\_state})$ grid (3 seeds, $n\!=\!108$ per row; the headline result is reported in the main paper as Table~\ref{tab:vankadara_step1_split}).

\paragraph{Vankadara's prediction holds at init for $B$ and $C$ but fails for $A_{\log}$.}
At step 1 the faithful prescription produces near-zero $N_u$ dependence for $B$ ($\alpha\!=\!+0.006$) and $C$ ($\alpha\!=\!+0.029$): $\eta_B$ and $\eta_C$ cancel the underlying gradient growth and deliver the predicted width-invariant update.
For $A_{\log}$, however, the faithful update has $\alpha\!=\!+1.00$, growing approximately linearly with $N_u$.
This is not a finite-width residual: under \naive $\muP$ the $A_{\log}$ update already has $\alpha\!=\!+0.86$ at step 1, so $\eta_A\!\propto\!N_u$ is not canceling an underlying $1/N_u$ decay (no such decay is present in our gradients), but \emph{amplifying} the existing $N_u$-dependence.
This is consistent with the simplified-ZOH discretization gap (Appendix~\ref{sec:appendix:vankadara_faithful}): Vankadara's derivation assumes true ZOH, where $A_{\log}$'s gradient may decay as predicted, but our implementation uses the field-standard simplified form (Mamba library, Mamba-2~\citep{dao2024mamba2}, Jamba~\citep{lieber2024jamba}, Nemotron-H~\citep{nvidia2025nemotron}, Falcon-H1~\citep{zuo2025falcon}).

\paragraph{The init-level invariance does not survive training.}
Across the stable window (steps 100--199), the faithful $\alpha(N_u)$ for $B$ drifts from $+0.006$ to $+1.35$ ($\Delta\alpha\!=\!+1.34$) and for $C$ from $+0.029$ to $+0.71$ ($\Delta\alpha\!=\!+0.68$); for $A_{\log}$, faithful drifts from $+1.00$ to $+3.42$ ($\Delta\alpha\!=\!+2.42$).
The explicit per-component LR rules thus achieve their predicted invariance only at step~1 for $B$ and $C$; during training the structure is destroyed at a rate \emph{larger} for the faithful prescription than for \naive ($\Delta\alpha$ for \naive stays in $[0.33, 0.68]$ across all three components).
The init-level Vankadara replication is therefore a $1$-step phenomenon, not a property the framework continues to enforce as the optimizer trajectory leaves the infinite-width regime; the LR-operating-point caveat below sharpens this reading.

\paragraph{Stable window: LR-specific operating points, not scaling laws.}
The stable-window exponents in Table~\ref{tab:vankadara_step1_split} are at base LR $\eta\!=\!1$ under SGD and should be read as LR-specific equilibrium operating points rather than fundamental scaling laws.
Cross-checking \naive $\muP$ SGD stable-window fits at base LR $\eta\!=\!1$ and $\eta\!=\!10$, the coefficients inflate $4$--$18\times$ while residuals remain near-Gaussian in both cases; alternative functional forms (semi-log, $\log\!+\!$linear envelope) produce the same coefficient inflation, so LR-dependence is not a log-log artifact.
The residual-shape distinction is robust across LRs and describes a structural difference between optimizers: AdamW's $g/\sqrt{v}$ introduces coordinate-wise coupling that disrupts log-linear structure (heavy-tailed residuals, excess kurtosis $+5$ to $+19$), while SGD's unit update reflects pure weight-trajectory dynamics (near-Gaussian residuals).
The Table~\ref{tab:vankadara_step1_split} stable-window column should be read as confirming that the init-level invariance for $B$ and $C$ does not persist along the trajectory, not as a precise width-scaling exponent for any particular LR.

\begin{figure}[H]
\centering
\includegraphics[width=\linewidth]{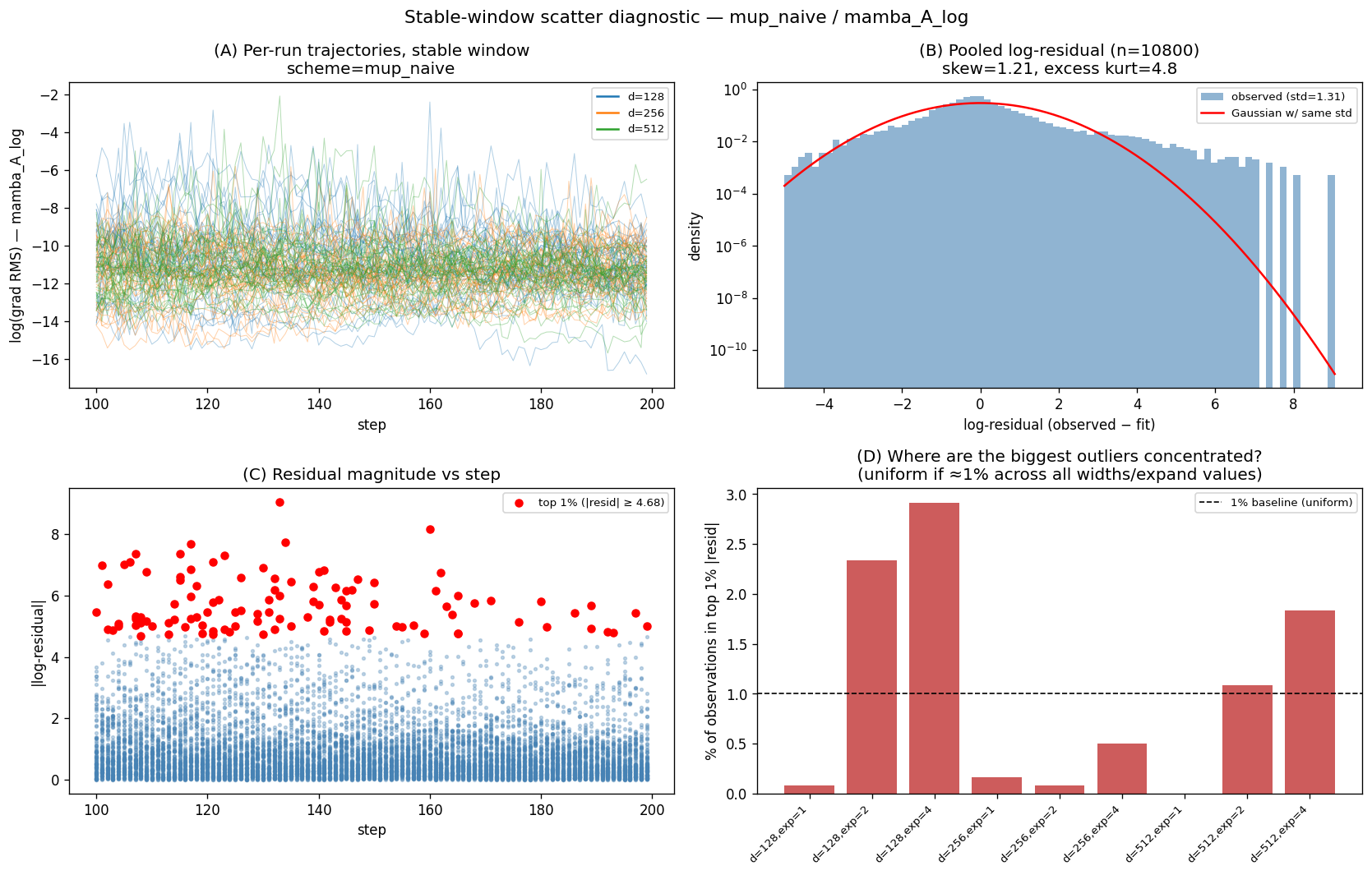}
\caption{Stable-window scatter diagnostic for MHMH \naive $\muP$, $A_{\log}$.
(A) Per-run log-gradient trajectories over steps 100--199 show isolated single-step collapses (downward spikes) rather than smooth noise.
(B) Pooled log-residual histogram against a Gaussian with matching standard deviation. The residuals are strongly heavy-tailed (excess kurtosis $+4.8$).
(C) Top-1\% $|$residual$|$ values scattered throughout the stable window; no concentration at early or late steps.
(D) Outlier concentration by $(d, \texttt{expand})$ cell: non-uniform, concentrated in $\texttt{expand}=2, 4$ cells; $\texttt{expand}=1$ quiet.
The outliers are transient update-instability events structured by configuration, not uniform measurement noise.}
\label{fig:stable_scatter_appendix}
\end{figure}

\paragraph{Outlier structure (AdamW, \naive $\muP$).}
Pooled log-residuals under AdamW are heavy-tailed for every scheme (excess kurtosis $+5$ to $+19$, skewness $+1$ to $+3$).
Top-$1\%$ $|$residual$|$ events cluster in specific $(d, \texttt{expand})$ cells rather than uniformly (Fig.~\ref{fig:stable_scatter_appendix}):\ for \naive $\muP$, $\texttt{expand}=1$ is uniformly quiet ($\sim\!0\%$), while $\texttt{expand}\in\{2,4\}$ concentrate outliers at the extreme $d$ values.
Per-run trajectories show these as isolated single-step gradient collapses: transient update-instability events, not Gaussian measurement noise.
This clarifies that the low $R^2$ of AdamW stable-window fits reflects structured outliers (configuration-dependent transient spikes) rather than purely random scatter.

\paragraph{The cleanliness--performance inversion under AdamW.}
The same effective-update analysis run on AdamW data (not shown for length) inverts the ordering: at the stable-window operating point under AdamW, the faithful prescription's $A_{\log}$ update has the cleanest log-log structure (highest $R^2$), but block-wise underperforms \naive on hybrid training (Appendix~\ref{sec:appendix:vankadara_faithful}, Appendix~\ref{sec:appendix:optimizer}).
This is consistent with the redundancy story (Section~\ref{sec:analysis:redundant_normalization}): under SGD, $\eta_A \!\propto\! N_u$ is doing useful but in our setting overcorrected work for $A_{\log}$, while under AdamW the $g/\sqrt{v}$ normalization already supplies width-stable per-component balance, so adding the explicit rule slows convergence without changing the convergent optimum.
Clean stable-window gradient structure under AdamW is therefore a \emph{diagnostic} that a per-parameter LR rule is firing, not a predictor of training success.

\subsection{Optimizer Ablations}
\label{sec:appendix:optimizer}

We ablate Adam's components one at a time while holding the parameterization fixed.
Adam differs from SGD in three ways: adaptive LR ($g/\sqrt{v}$), momentum ($\beta_1$), and weight decay ($\lambda$).

\subsubsection{SGD vs AdamW at 2k Steps}

56 configs: HHHH/MHMH $\times$ $w \in \{256, 1024\}$ $\times$ SGD/AdamW $\times$ 7 LRs, 2k steps on WikiText-103.

SGD barely trains at 2k steps (loss $\sim$7--9 vs random baseline 11.5), while AdamW achieves loss $\sim$4.1--4.4 at $\eta{=}1.0$.
Inconclusive for transfer comparison at this short training horizon.

\subsubsection{SGD at 20k Steps with Extended LR Range}

To ensure SGD trains meaningfully, we extend to 20k steps and LR up to 100.0.
48/48 configs complete (MHMH + HHHH, both widths, AdamW + SGD; Table~\ref{tab:full_sgd_vs_adamw}).

\begin{table}[H]
\centering\small
\caption{SGD vs AdamW for \MHMH and \HHHH (\naive $\muP$, \textsc{WikiText-103}, 20k steps, warmup 1000).}
\label{tab:full_sgd_vs_adamw}
\begin{tabular}{llcccccc}
\toprule
Arch & Optimizer $\times$ Width & $\eta$=0.3 & $\eta$=1.0 & $\eta$=3.0 & $\eta$=10.0 & $\eta$=30.0 & $\eta$=100.0 \\
\midrule
\multirow{4}{*}{MHMH} & AdamW $w$=256  & 3.566 & \textbf{3.518} & 3.616 & 4.018 & 9.909 & 5.445 \\
                     & AdamW $w$=1024 & 3.226 & \textbf{3.129} & 3.567 & 3.667 & 3.838 & \textit{10.5} \\
                     & SGD   $w$=256  & 6.927 & 6.503 & 6.176 & 5.668 & 5.238 & \textbf{5.121} \\
                     & SGD   $w$=1024 & 7.296 & 6.736 & 6.329 & 6.147 & 5.687 & \textbf{5.176} \\
\midrule
\multirow{4}{*}{HHHH} & AdamW $w$=256  & 3.505 & \textbf{3.487} & 3.581 & \textit{div} & 4.542 & \textit{10.9} \\
                     & AdamW $w$=1024 & 3.160 & \textbf{3.082} & 3.476 & 3.691 & 3.829 & \textit{10.9} \\
                     & SGD   $w$=256  & 6.870 & 6.374 & 5.986 & 5.506 & 5.233 & \textbf{5.122} \\
                     & SGD   $w$=1024 & 7.389 & 6.856 & 6.399 & 5.841 & 5.403 & \textbf{5.174} \\
\bottomrule
\end{tabular}
\end{table}

SGD requires at least $100\times$ higher LR than Adam and is $46\%$ worse at its best tested LR (MHMH $w{=}256$: SGD $5.121$ vs AdamW $3.518$).
Critically, SGD shows \emph{reversed width scaling}: $w{=}1024$ performs worse than $w{=}256$ at every LR up to $\eta{=}100$ (delta $+1.1\%$ to $+8.6\%$).
Under AdamW, wider models perform better as expected, and $\eta{=}1.0$ remains optimal across widths (0.0\% gap).

\paragraph{Above $\eta{=}100$.}
\label{sec:appendix:sgd_extended}
Every SGD row of Table~\ref{tab:full_sgd_vs_adamw} has its best loss at $\eta{=}100$, the top of the grid.
We extend \HHHH{} in the same configuration (the \naive-$\muP$ parameter groups of Section~\ref{sec:setup:schemes}, 20k steps, one seed) to $\eta \in \{300, 1000, 3000, 7000\}$ (Table~\ref{tab:sgd_extended}).
The argmin is $\eta{=}1000$ at $w{=}256$ and $\eta{=}3000$ at $w{=}1024$, and the lowest LR that diverges is $3000$ at $w{=}256$ and $7000$ at $w{=}1024$; in this configuration both the argmin and the stability edge move with width.
The extension is limited to \HHHH{}: \MHMH{} above $\eta{=}100$ was run only at $w{=}256$.

\begin{table}[H]
\centering\small
\caption{\HHHH{} under SGD above $\eta{=}100$, in the configuration of Table~\ref{tab:full_sgd_vs_adamw} (\naive-$\muP$ parameter groups, \textsc{WikiText-103}, 20k steps, one seed): final training loss. The $\eta{=}100$ column repeats Table~\ref{tab:full_sgd_vs_adamw}. Bold marks the argmin of each row; \textit{div}: diverged; --: not run.}
\label{tab:sgd_extended}
\begin{tabular}{lccccc}
\toprule
Optimizer $\times$ Width & $\eta$=100 & $\eta$=300 & $\eta$=1000 & $\eta$=3000 & $\eta$=7000 \\
\midrule
SGD $w$=256 & 5.122 & 5.077 & \textbf{5.067} & \textit{div} & -- \\
SGD $w$=1024 & 5.174 & 5.033 & 5.010 & \textbf{5.004} & \textit{div} \\
\bottomrule
\end{tabular}
\end{table}

\subsubsection{Approximated Block-wise $\muP$ with SGD}

24 additional configs: HHHH/MHMH $\times$ $w \in \{256, 1024\}$ $\times$ 6 LRs, approximated block-wise $\muP$ + SGD (momentum=0.9), 20k steps.
This completes the $2{\times}2$ matrix of $\{\text{naive}, \text{approximated block-wise}\} \times \{\text{AdamW}, \text{SGD}\}$.

\begin{table}[H]
\centering
\small
\caption{$2{\times}2$ matrix: parameterization $\times$ optimizer (MHMH, best LR at each width).}
\label{tab:2x2_matrix}
\setlength{\tabcolsep}{4pt}
\begin{tabular}{llcccc}
\toprule
Parameterization & Optimizer & Best LR @256 & Loss @256 & Best LR @1024 & Loss @1024 \\
\midrule
\Naive $\muP$ & AdamW & 1.0 & 3.518 & 1.0 & \textbf{3.129} \\
\Naive $\muP$ & SGD & 100.0 & 5.121 & 100.0 & 5.176 \\
Approximated block-wise $\muP$ & SGD & 100.0 & 5.122 & 100.0 & 5.408 \\
\bottomrule
\end{tabular}
\end{table}

Approximated block-wise $\muP$ does not fix SGD's reversed width scaling.
Both SGD variants show $w{=}1024$ performing worse than $w{=}256$, with the gap \emph{larger} under approximated block-wise $\muP$ for MHMH ($+5.6\%$ vs $+1.1\%$).

\begin{figure}[H]
\centering
\includegraphics[width=\linewidth]{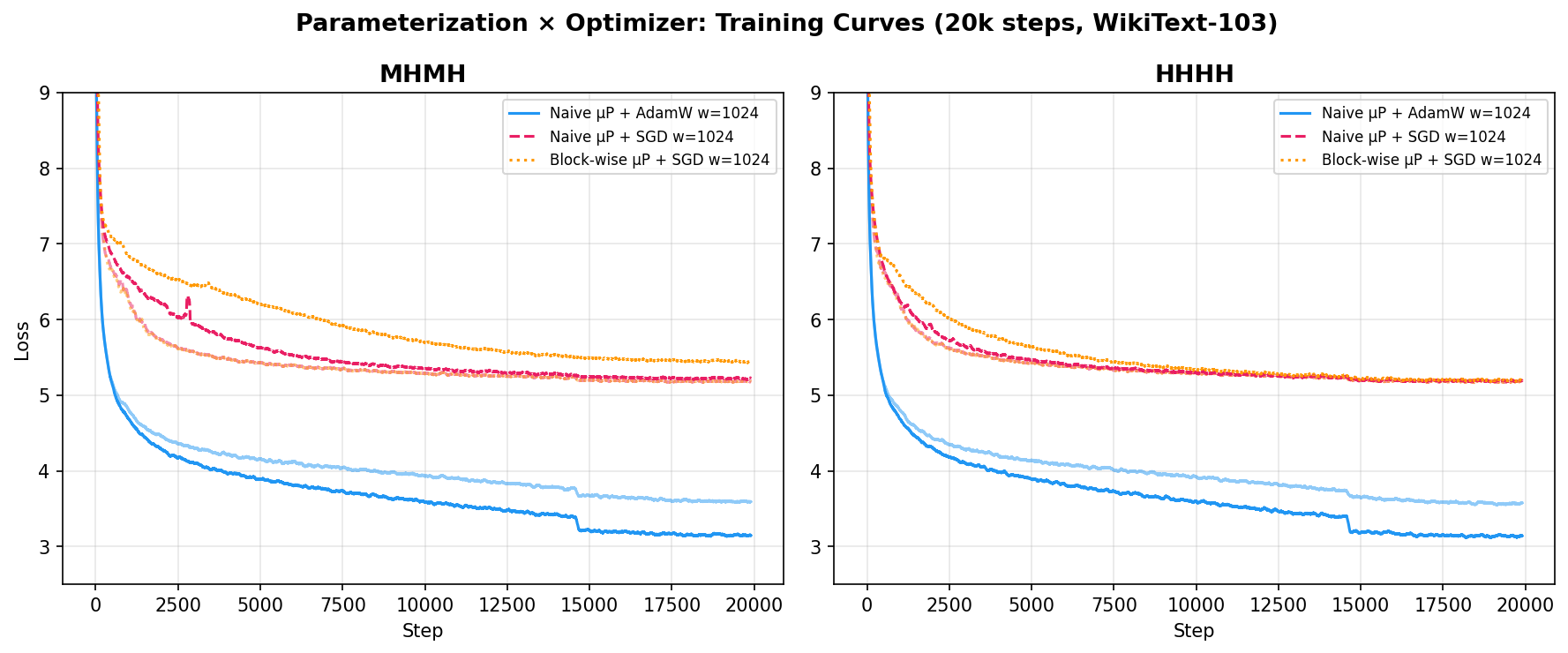}
\caption{Training curves for three parameterization--optimizer combinations.
AdamW (blue) separates by width (wider = better); SGD variants (pink, orange) overlap or show reversed scaling.
Approximated block-wise $\muP$ does not improve SGD over \naive $\muP$.}
\label{fig:blockwise_sgd_curves}
\end{figure}

\begin{figure}[H]
\centering
\includegraphics[width=\linewidth]{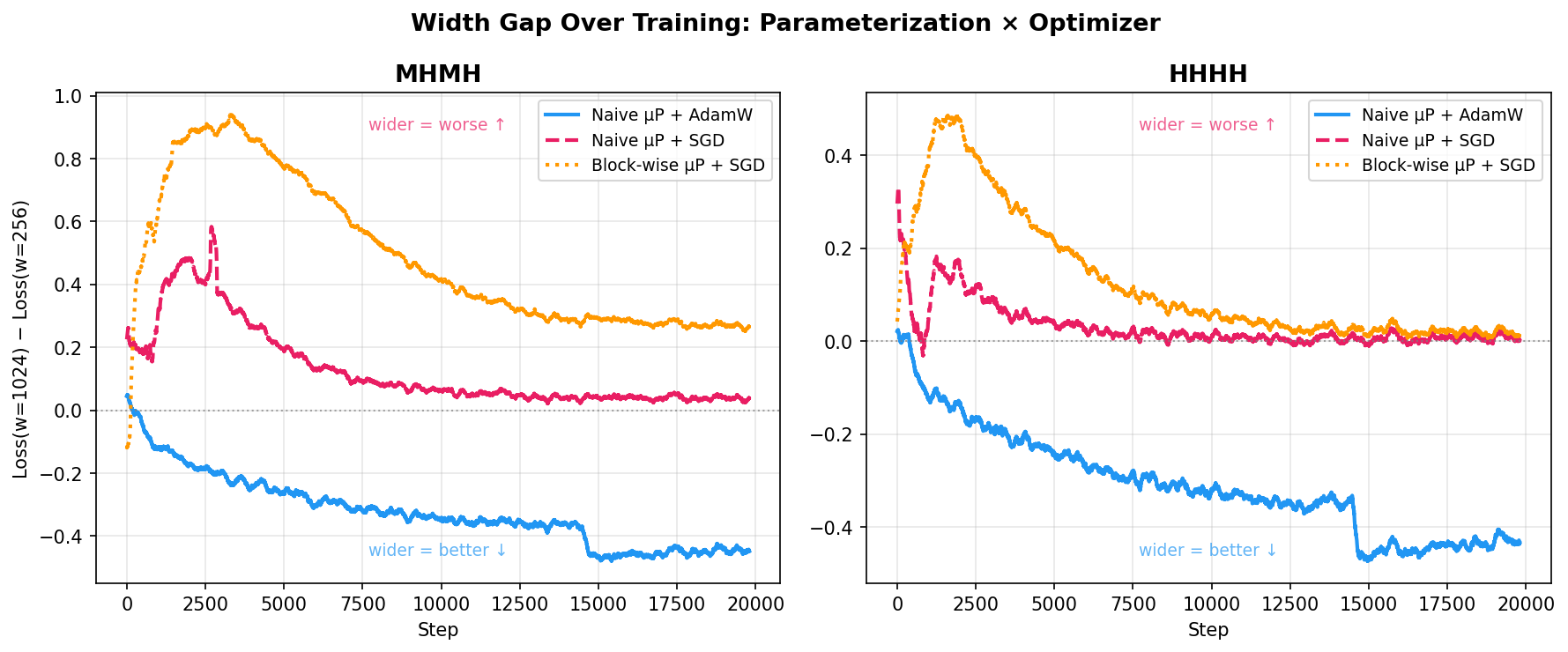}
\caption{Width gap (loss at $w{=}1024$ minus loss at $w{=}256$) over training.
Positive values indicate reversed scaling (wider = worse).
AdamW (blue) achieves healthy scaling; both SGD variants (pink, orange) show persistent reversed scaling.}
\label{fig:blockwise_sgd_width_gap}
\end{figure}

\subsubsection{Adam without Weight Decay}

36 configs: MM/HHHH/MHMH $\times$ $w \in \{256, 1024\}$ $\times$ WD $\in \{0.0, 0.1\}$ $\times$ 3 LRs (Table~\ref{tab:full_adam_nowd}).
$\eta{=}1.0$ transfers for all 3 architectures with WD=0.0; weight decay shifts the loss by at most $0.4\%$ and leaves the optimal LR unchanged.\footnote{At $w{=}1024$, \MM{} with weight decay reaches a $0.1\%$ lower loss at $\eta{=}3.0$ than at $\eta{=}1.0$, within the flat basin around the optimum. The ablation runs for 2k steps, shorter than the weight-decay timescale at $w{=}1024$.}
Weight decay is therefore not load-bearing for the LR-transfer property in our setting.

\begin{table}[H]
\centering\small
\caption{Adam no-WD ablation (\textsc{WikiText-103}, 2k steps, \naive $\muP$).}
\label{tab:full_adam_nowd}
\begin{tabular}{lccc}
\toprule
 & $\eta$=0.3 & $\eta$=1.0 & $\eta$=3.0 \\
\midrule
HHHH $w$=256 WD=0 & 4.573 & \textbf{4.411} & 4.517 \\
HHHH $w$=256 WD=0.1 & 4.561 & \textbf{4.404} & 5.013 \\
HHHH $w$=1024 WD=0 & 4.392 & \textbf{4.090} & 4.471 \\
HHHH $w$=1024 WD=0.1 & 4.381 & \textbf{4.102} & 4.161 \\
MHMH $w$=256 WD=0 & 4.411 & \textbf{4.234} & 4.296 \\
MHMH $w$=256 WD=0.1 & 4.411 & \textbf{4.217} & 4.234 \\
MHMH $w$=1024 WD=0 & 4.455 & \textbf{4.165} & 4.300 \\
MHMH $w$=1024 WD=0.1 & 4.451 & \textbf{4.170} & 4.254 \\
MM $w$=256 WD=0 & 4.557 & \textbf{4.355} & 4.479 \\
MM $w$=256 WD=0.1 & 4.549 & \textbf{4.361} & 4.447 \\
MM $w$=1024 WD=0 & 4.599 & \textbf{4.397} & 4.452 \\
MM $w$=1024 WD=0.1 & 4.593 & 4.402 & \textbf{4.397} \\
\bottomrule
\end{tabular}
\end{table}

\subsubsection{Momentum Ablation}

56 configs: HHHH/MHMH $\times$ $w \in \{256, 1024\}$ $\times$ 7 LRs $\times$ $\{$Adam ($\beta_1{=}0$), SGD (no momentum)$\}$ (Table~\ref{tab:full_optim_ablation}).

\begin{table}[H]
\centering\footnotesize
\caption{Optimizer ablation for \MHMH and \HHHH (2k steps, \textsc{WikiText-103}, \naive $\muP$). Final loss at each LR; bold marks the optimum within each row.}
\label{tab:full_optim_ablation}
\begin{tabular}{llccccccc}
\toprule
Arch & Optim $\times$ Width & $\eta$=0.001 & $\eta$=0.003 & $\eta$=0.01 & $\eta$=0.03 & $\eta$=0.1 & $\eta$=0.3 & $\eta$=1.0 \\
\midrule
\multirow{8}{*}{MHMH} & AdamW       $w$=256  & 9.968 & 8.215 & 6.638 & 5.625 & 4.898 & 4.412 & \textbf{4.219} \\
                     & AdamW       $w$=1024 & 9.303 & 7.950 & 6.665 & 5.706 & 5.020 & 4.450 & \textbf{4.160} \\
                     & Adam ($\beta_1{=}0$) $w$=256  & 9.951 & 8.194 & 6.726 & 5.653 & 4.954 & \textbf{4.448} & 4.511 \\
                     & Adam ($\beta_1{=}0$) $w$=1024 & 9.289 & 7.935 & 6.718 & 5.737 & 5.059 & 4.608 & \textbf{4.418} \\
                     & SGD         $w$=256  & \textit{11.5} & \textit{11.5} & \textit{11.5} & \textit{11.4} & 9.999 & 8.462 & \textbf{7.626} \\
                     & SGD         $w$=1024 & \textit{11.5} & \textit{11.5} & \textit{11.5} & \textit{11.4} & \textit{10.9} & 8.821 & \textbf{7.911} \\
                     & SGD (no momentum) $w$=256  & \textit{11.5} & \textit{11.5} & \textit{11.5} & \textit{11.5} & \textit{11.5} & \textit{11.3} & \textbf{9.820} \\
                     & SGD (no momentum) $w$=1024 & \textit{11.5} & \textit{11.5} & \textit{11.5} & \textit{11.5} & \textit{11.5} & \textit{11.4} & \textit{10.9} \\
\midrule
\multirow{8}{*}{HHHH} & AdamW       $w$=256  & 9.974 & 8.249 & 6.729 & 5.742 & 5.051 & 4.561 & \textbf{4.392} \\
                     & AdamW       $w$=1024 & 9.277 & 7.923 & 6.719 & 5.694 & 4.961 & 4.386 & \textbf{4.099} \\
                     & Adam ($\beta_1{=}0$) $w$=256  & 9.957 & 8.230 & 6.778 & 5.767 & 5.104 & 4.615 & \textbf{4.470} \\
                     & Adam ($\beta_1{=}0$) $w$=1024 & 9.263 & 7.909 & 6.719 & 5.720 & 5.020 & 4.492 & \textbf{4.465} \\
                     & SGD         $w$=256  & \textit{11.5} & \textit{11.5} & \textit{11.5} & \textit{11.3} & 9.298 & 8.415 & \textbf{7.732} \\
                     & SGD         $w$=1024 & \textit{11.5} & \textit{11.5} & \textit{11.5} & \textit{11.4} & \textit{10.9} & 8.753 & \textbf{7.955} \\
                     & SGD (no momentum) $w$=256  & \textit{11.5} & \textit{11.5} & \textit{11.5} & \textit{11.5} & \textit{11.5} & \textit{11.3} & \textbf{9.398} \\
                     & SGD (no momentum) $w$=1024 & \textit{11.5} & \textit{11.5} & \textit{11.5} & \textit{11.5} & \textit{11.5} & \textit{11.4} & \textit{10.8} \\
\bottomrule
\end{tabular}
\end{table}

Removing Adam's momentum costs $2$--$9\%$ in loss (e.g., MHMH $w{=}1024$: $4.160 \!\to\! 4.418$, $+6.2\%$; HHHH $w{=}1024$: $4.099 \!\to\! 4.465$, $+8.9\%$) but leaves transfer intact: $3/4$ (arch, width) cells keep $\eta^*{=}1.0$, with one borderline shift to $\eta{=}0.3$ (Adam $\beta_1{=}0$ on MHMH $w{=}256$).
Removing SGD's momentum is catastrophic: loss climbs to $9.4$--$10.9$, barely below the random baseline of $11.5$, an additional $22$--$38\%$ degradation that wipes out training entirely.
Momentum therefore aids optimization speed but is not the transfer mechanism; only Adam's $g/\sqrt{v}$ rescaling is.

\paragraph{Note on the LR grid boundary.}
The grid in Table~\ref{tab:full_optim_ablation} stops at $\eta{=}1.0$, so any AdamW or Adam ($\beta_1{=}0$) optimum reported at $\eta{=}1.0$ is at the upper boundary of \emph{this} grid.
The width-transfer sweep (Table~\ref{tab:width_transfer}) extends to $\eta{=}3.0$ on the same (arch, width) cells under \naive $\muP$ + AdamW and confirms $\eta^*{=}1.0$ remains optimal there, so the AdamW optimum is genuinely $\eta{=}1.0$ rather than an artifact of the local boundary.
The SGD rows here also peg the boundary, but the wider sweep in Table~\ref{tab:full_sgd_vs_adamw} extends to $\eta{=}100$ and, with Table~\ref{tab:sgd_extended}, shows SGD's best loss stays near $5$ at both widths, never approaching AdamW's $\sim 3.1$; the SGD-vs-AdamW comparison is therefore qualitative and unchanged by the boundary.

\subsubsection{Adam Epsilon Sensitivity}
\label{sec:appendix:adam_eps}

We test whether the LR transfer claim is sensitive to AdamW's $\epsilon$.
We sweep $\epsilon \in \{10^{-10}, 10^{-8}, 10^{-6}, 10^{-4}\}$ on \MHMH at $w \in \{256, 1024\}$ and $\eta \in \{0.5, 1.0, 1.5\}$ for 2k steps on \textsc{WikiText-103} (24 configs, Table~\ref{tab:adam_eps}).

\begin{table}[H]
\centering
\small
\caption{AdamW $\epsilon$ sensitivity for \MHMH (\textsc{WikiText-103}, 2k steps, \naive $\muP$). Each $\eta$ block has paired rows for $w{=}256$ and $w{=}1024$ so width-invariance can be read directly across the row pair; bold marks the optimum across the four $\epsilon$ values within each row.}
\label{tab:adam_eps}
\begin{tabular}{cccccc}
\toprule
$\eta$ & $w$ & $\epsilon{=}10^{-10}$ & $\epsilon{=}10^{-8}$ & $\epsilon{=}10^{-6}$ & $\epsilon{=}10^{-4}$ \\
\midrule
\multirow{2}{*}{$0.5$} & $256$  & 4.306 & 4.293 & \textbf{4.260} & 4.557 \\
                      & $1024$ & 4.287 & 4.286 & \textbf{4.281} & 4.852 \\
\midrule
\multirow{2}{*}{$1.0$} & $256$  & 4.236 & 4.209 & \textbf{4.177} & 4.316 \\
                      & $1024$ & 4.176 & 4.167 & \textbf{4.150} & 4.447 \\
\midrule
\multirow{2}{*}{$1.5$} & $256$  & 4.219 & 4.211 & \textbf{4.163} & 4.243 \\
                      & $1024$ & 4.156 & \textbf{4.148} & 4.209 & 4.302 \\
\bottomrule
\end{tabular}
\end{table}

At the $\muP$ optimum $\eta{=}1.0$, varying $\epsilon$ over $\{10^{-10}, 10^{-8}, 10^{-6}\}$ shifts the loss by less than $0.07$ nats at both widths, and $\eta{=}1.0$ stays within $0.5\%$ of the best LR in every cell; $\epsilon{=}10^{-6}$ slightly outperforms the default $10^{-8}$ ($\sim 0.03$ nats better at both widths).
At $\epsilon{=}10^{-4}$ the loss degrades sharply ($+0.10$ nats at $w{=}256$, $+0.30$ nats at $w{=}1024$): a large $\epsilon$ suppresses Adam's per-parameter normalization.
This is consistent with the two-condition decomposition (Section~\ref{sec:analysis:two_conditions}): when $\epsilon$ is small relative to typical $\sqrt{v_W}$, the $g/\sqrt{v}$ normalization operates as designed and Condition~2 holds; when $\epsilon$ becomes comparable to $\sqrt{v_W}$, AdamW degenerates toward SGD-like behavior and Condition~2 fails, producing the same width-dependent degradation we observe under SGD itself.

\subsubsection{Sophia and Muon}
\label{sec:appendix:sophia_muon}

We run two further optimizers on \HHHH{} and \MHMH{} with \naive $\muP$ at $w \in \{256, 512, 1024\}$, 20k steps on \textsc{WikiText-103}, three seeds each, and report validation loss.
Sophia~\citep{liu2024sophia} (SophiaG, $\beta{=}(0.965, 0.99)$, $\rho{=}0.04$, Hessian estimate every 10 steps) uses the \naive-$\muP$ parameter groups.
Muon~\citep{jordan2024muon} (momentum $0.95$) orthogonalizes the updates of the hidden weight matrices; that group uses Muon's own update scaling at $\eta_{\text{base}}$, without the $\muP$ multiplier, and the remaining parameters (embeddings, output head, norms, SSM parameters) use AdamW with the \naive-$\muP$ groups.
Sophia keeps its optimum at every width, while Muon's optimum moves at $w{=}1024$ (Table~\ref{tab:sophia_muon}).
Sophia's grid has three points spaced about $3\times$ apart, so its $0.00\%$ gap holds at that resolution.
A first run of the same grids without a held-out evaluator selected the same $\eta^*$ as the rerun at $w{=}256$ and $w{=}1024$ for every seed, with Muon training-loss gaps of 3.3--11.3\%; Table~\ref{tab:sophia_muon} reports the rerun that adds the evaluator.
Sophia and Muon differ in prescription as well as in normalizer, so the comparison does not isolate per-element normalization.

\begin{table}[H]
\centering\small
\caption{Sophia and Muon on \textsc{WikiText-103} (20k steps, three seeds, validation loss). $\eta^*$ is the argmin of the seed-mean loss; the gap is the transfer gap from $w{=}256$ to $w{=}1024$, for the seed mean and per seed. LR grids: Sophia $\{0.03, 0.1, 0.3\}$, Muon $\{0.002, 0.005, 0.01, 0.02, 0.05, 0.1\}$.}
\label{tab:sophia_muon}
\begin{tabular}{llccccc}
\toprule
Optimizer & Arch & $\eta^*$ $w{=}256$ & $w{=}512$ & $w{=}1024$ & Gap (mean) & Gap (per seed) \\
\midrule
Sophia & \HHHH & 0.1 & 0.1 & 0.1 & 0.00\% & 0.00\% (3/3) \\
Sophia & \MHMH & 0.1 & 0.1 & 0.1 & 0.00\% & 0.00\% (3/3) \\
Muon & \HHHH & 0.01 & 0.01 & 0.005 & 3.58\% & 2.29--5.74\% \\
Muon & \MHMH & 0.01 & 0.01 & 0.005 & 7.83\% & 5.62--10.38\% \\
\bottomrule
\end{tabular}
\end{table}

\subsubsection{Confound Resolution Summary}

\begin{table}[H]
\centering
\small
\caption{Adam-component contribution to LR transfer: which feature is required vs.\ incidental.}
\begin{tabular}{llll}
\toprule
Adam component & Role for LR transfer & Evidence \\
\midrule
Adaptive LR ($g/\sqrt{v}$) & \textbf{Load-bearing} & Only Adam variants transfer \\
Momentum ($\beta_1$) & Not load-bearing & 2--9\% degradation, transfer preserved \\
Weight decay ($\lambda$) & Not load-bearing & $\leq$0.4\% effect \\
\bottomrule
\end{tabular}
\end{table}

\end{document}